\documentclass[twoside,11pt]{article}

\usepackage[hyperref]{jmlr2e}
\usepackage{algorithm}
\usepackage{algorithmic}
\usepackage{amsmath}
\usepackage{mathrsfs}
\usepackage{mathtools}
\usepackage{array}
\usepackage{subfigure}
\usepackage{multirow}
\usepackage{caption}
\usepackage{booktabs}
\usepackage{rotating}
\usepackage{setspace}
\usepackage{bm}
\usepackage{multirow}
\usepackage{xcolor}
\usepackage{dcolumn}
\usepackage{setspace}
\usepackage{enumitem}
\usepackage{appendix}
\usepackage{pgfplots}
\usepackage{clipboard}
\newclipboard{myclipboard}

\usepgfplotslibrary{fillbetween}
\usetikzlibrary{pgfplots.groupplots}
\pgfplotsset{compat=1.14}
\pgfplotscreateplotcyclelist{mylist}{
	{blue,mark options={fill=blue!80!black},mark=*},
	{red,mark options={fill=red!80!black},mark=square*},
	{brown!60!black,mark options={fill=brown!80!black,scale=1.2},mark=diamond*},
	{densely dashed,blue,mark options={solid},mark=o},
	{densely dashed,red,mark options={solid},mark=square},
	{densely dashed,brown!60!black,mark options={solid,scale=1.2},mark=diamond}
}

\DeclareMathOperator*{\argmax}{arg\,max}
\DeclareMathOperator*{\argmin}{arg\,min}
\DeclareMathOperator{\sign}{\overrightarrow{\rm{sign}}}

\DeclareMathOperator\supp{supp}
\newcommand*{\dif}{\mathop{}\!d}
\newcommand{\PR}{\mathbb P}
\newcommand{\EX}{\mathbb E}
\newcommand{\YY}{\mathcal Y}
\newcommand{\XX}{\mathcal X}
\newcommand{\Ind}[1]{\mathbb{I}(#1)}

\newcommand{\RV}{} 

\newtheorem{assumption}{Assumption}
\newtheorem{remark}{Remark}
\newenvironment{lastassumption}[1] {\renewcommand{\theassumption}{\ref{#1}'}
	\addtocounter{assumption}{-1}
	\begin{assumption}}
	{\end{assumption}}

\ShortHeadings{Nonparametric sparse contextual learning}{Li, Chen and Hong}
\firstpageno{1}
\begin{document}

\title{Dimension Reduction in Contextual Online Learning via Nonparametric Variable Selection}

\author{\name Wenhao Li \email wwenhao.li@mail.utoronto.ca \\
	\addr Rotman School of Management\\
		  University of Toronto\\
	 	  Toronto, Ontario M5S 3E6, Cananda
 	  	  \AND
   	\name Ningyuan Chen \email ningyuan.chen@utoronto.ca \\  \addr Rotman School of Management\\
   	  	  University of Toronto\\
     	  Toronto, Ontario M5S 3E6, Cananda
       	\AND
    \name L. Jeff Hong \email hong\_liu@fudan.edu.cn \\
	\addr School of Management and School of Data Science \\
		  Fudan University\\
	  	  Shanghai, China}

\editor{}

\maketitle
\begin{abstract}
	We consider a contextual online learning (multi-armed bandit) problem with high-dimensional covariate $\bm{x}$ and decision $\bm{y}$. The reward function to learn, $f(\bm{x},\bm{y})$, does not have a particular parametric form. The literature has shown that the optimal regret is $\tilde{O}(T^{(d_x\!+\!d_y\!+\!1)/(d_x\!+\!d_y\!+\!2)})$, where $d_x$ and $d_y$ are the dimensions of $\bm x$ and $\bm y$, and thus it suffers from the curse of dimensionality. In many applications, only a small subset of variables in the covariate affect the value of $f$, which is referred to as \textit{sparsity} in statistics. To take advantage of the sparsity structure of the covariate, we propose a variable selection algorithm called \textit{BV-LASSO}, which incorporates novel ideas such as binning and voting to apply LASSO to nonparametric settings.
	Using it as a subroutine, we can achieve the regret $\tilde{O}(T^{(d_x^*\!+\!d_y\!+\!1)/(d_x^*\!+\!d_y\!+\!2)})$, where $d_x^*$ is the effective covariate dimension.
	The regret matches the optimal regret when the covariate is $d^*_x$-dimensional and thus cannot be improved. Our algorithm may serve as a general recipe to achieve dimension reduction via variable selection in nonparametric settings.
\end{abstract}



\section{Introduction}\label{sec:intro}
Online learning is a popular paradigm to study dynamic decision making when new information can be collected actively to improve the quality of decisions simultaneously.
It has seen numerous applications in the past decades in advertising, retailing, health care and so on.
To accommodate the increasingly complex nature of many modern applications, the online learning framework has been extended in various directions,
including
\begin{itemize}
	\item A large (sometimes infinite) set of possible decisions. For instance, in dynamic pricing, a firm sets prices dynamically for a number of products over time, in order to learn the substitution patterns as well as the demand elasticity, to maximize revenues in the long run.
	The candidate decisions are the prices charged for various products, which are virtually infinite and high-dimensional.
	The discrete set of decisions used in MAB cannot properly capture the nature of dynamic pricing, and researchers have designed algorithms for continuous and high-dimensional decision variables.
	\item Contextual information or covariates.
	Covariates refer to the contextual information that is available for the decision maker to assess the current situation and make better decisions.
	In the example of dynamic pricing, when setting prices for a particular consumer, the personal information such as age, gender, and address can be used to infer the shopping habit of the consumer.
	It allows the firm to extract more revenues from consumers by price discrimination, but at the same time calls for more sophisticated decision rules to incorporate the covariates when learning the demand.
	The existence of covariates is ubiquitous in practice.
	\item Modeling the reward function.
	Learning and maximizing the reward function is the central goal of online learning.
	However, when little information is available, it is sometimes risky to even impose a model of \textit{what to learn}.
	In dynamic pricing, it is tempting to assume that the demand is linear in the prices, and simplify the problem by learning only the linear coefficients.
	If the actual demand-price relationship is not linear, i.e., the model is misspecified, then the decision maker has little hope to find the optimal decision in the long run.
\end{itemize}
\label{page:intro-nonpar}
{\RV \Copy{cop:intro-nonpar}{Next we informally describe the framework of \textit{nonparametric contextual bandits} \citep{lu2009showing,slivkins2014contextual} to incorporate those extensions.}}
A formal formulation is introduced in Section~\ref{sec:problem}.
Consider a reward function $f(\bm x,\bm y)$, where $\bm x$ represents the covariate and $\bm y$ represents the decision.
Both $\bm x$ and $\bm y$ can be vectors.
The function is \textit{nonparametric} and does not have a specific form except for a few general structures such as continuity and smoothness.
In period $t$, a covariate $\bm{X}_t$ is generated and observed; the decision maker makes a decision $\bm y_t$ based on $\bm{X}_t$ as well as the historical information to maximize $f(\bm{X}_t,\bm y_t)$.
The goal is to learn the optimal decision $\bm y(\bm X_t)=\argmax_{\bm y} f(\bm{X}_t,\bm y)$.

Unfortunately, it has been shown that the problem suffers from the \textit{curse of dimensionality}.
In particular, the optimal regret of the problem, a common metric in online learning, is\footnote{We use $\tilde{O}$ to indicate asymptotic approximation neglecting logarithmic terms.} $\tilde{O}(T^{(d_x+d_y+1)/(d_x+d_y+2)})$ (see, e.g., \citealt{kleinberg2008multi,slivkins2014contextual}), where $d_x$ and $d_y$ are the dimensions of $\bm x$ and $\bm y$, respectively, and $T$ is the length of the learning horizon.
In other words, the difficulty to learn the unknown reward function scales rapidly with $d_x$ and $d_y$.
No decision makers are able to break the fundamental limit without imposing additional assumptions on the reward function $f$.

On the other hand, in many applications, the information in the covariate $\bm x$ is likely to contain a great deal of redundancy.
That is, out of $d_x$ variables in $\bm x$, many may not affect the value of $f$ at all.
This is referred to as \textit{sparsity} in statistics.
In the example of dynamic pricing, for instance,
the firm may have collected a rich set of personal information of a consumer (large $d_x$), while only a few key
variables such as the income level actually affect the purchasing behavior.
If we use $d_x^*$ to denote the effective covariate dimension, or the number of \textit{relevant} variables,
then the question is, without knowing how many and which variables are redundant/relevant, can the decision maker achieve the regret $\tilde{O}(T^{(d_x^*+d_y+1)/(d_x^*+d_y+2)})$?

This paper provides an affirmative answer to the above question.
Although such dimension reduction or variable selection has been one of the central topics in statistics for a few decades and has been well studied, the problem we consider is still very challenging because of the nonparametric nature of the reward function.
In particular, statistical tools that are commonly used in variable selection such as LASSO \citep{hastie2015statistical} are designed for certain parametric (linear) models.
Applied to our nonparametric setting where any parametric family may be misspecified, it is unclear if they would work at all.
Our paper addresses this challenge and contributes to the literature in the following aspects:
\begin{itemize}
	\item Through the lens of online learning, we provide a nonparametric variable selection algorithm based on which the online learning can achieve regret $\tilde{O}(T^{(d_x^*+d_y+1)/(d_x^*+d_y+2)})$.
	In other words, the algorithm facilitates the learning of the decision maker as if s/he is informed of the sparsity structure of the covariate, i.e., how many and which variables are relevant, in advance.
	The regret matches the optimal regret when the covariate is only $d^*_x$-dimensional and thus cannot be improved.
	Therefore, we answer the fundamental question raised previously: when the covariate is sparse, we are able to identify the relevant variables and effectively lift the curse of dimensionality in online learning, even if the reward function is nonparametric.
	\item Our algorithm has two recipes that contribute to the successful variable selection in the nonparametric setting. Both may be of independent interest. The first one is \textit{localized LASSO} (see Section~\ref{sec:monotone_algorithm}).
	We partition the covariate space into small bins. Within each bin, we apply LASSO to the observations.
	Although LASSO only works for linear functions, we are able to show that the misspecification error incurred by approximating an arbitrary function $f$ by linear functions can be controlled in a localized bin.
	That is, with properly chosen bin size and parameters, LASSO is able to identify relevant variables with high probability using the observations inside the bin despite the misspecification.
	This serves as the building block of our algorithm.
	\item Localized LASSO doesn't completely address the curse of dimensionality.
	To contain the approximation error of linear functions, the bin size needs to be small.
	As a result, the number of bins in a $d_x$-dimensional space grows exponentially in $d_x$ and there are few observations in each bin.
	We resolve this issue by our second recipe, \textit{weighted voting} (see Section~\ref{sec:monotone_algorithm}).
	We aggregate the outcomes of variable selection in each bin and obtain a global set of selected variables.
	Each bin has a ``vote'' for whether a variable is relevant or not, and the weights of their votes depend on their ``predictive power'', which is calculated by our algorithm.
	For example, the localized LASSO applied to bin $A$ predicts that $x_1$ is redundant, while bin $B$ predicts the opposite.
	If $A$'s vote carries more weight by our algorithm, possibly because it has more observations than $B$, then the algorithm makes a judgement that $x_1$ tends to be redundant.
	In this way, all the data in the covariate space are effectively utilized.
	The efficient use of data is reflected in our theoretical guarantee: the converence rate depends on the number of all observations as if the covariate space hadn't been partitioned.
\end{itemize}

We point out that the nonparametric variable selection algorithm is designed as a subroutine to select variables before applying the existing online learning algorithms. The algorithm may serve as a general recipe for variable selection in nonparametric settings, and therefore can be applied to other problems such as supervised learning.
Next we review the related literature in the domain.

\section{Related Literature}\label{sec:literature}
Our work is related to the literature studying nonparametric variable selection, contextual bandits and dynamic pricing with demand learning.
We review the three streams below.

\subsection{Nonparametric Variable Selection}\label{sec:literature-variable-selection}
In machine learning and statistics, the variable selection problem has been studied extensively. 
Suppose samples of $(Y, X_1,\dots,X_{d_x})$ can be observed.
Variable selection is concerned with the identification of relevant $X_i$s that matter to the value of $Y$.
Among the various methods proposed, LASSO is probably the most well-known.
It combines computational efficiency and analytical tractability and is widely used in practice
(see \citealt{buhlmann2011statistics, hastie2015statistical} for a complete bibliography).
However, LASSO assumes that $Y$ depends on $(X_1,\dots,X_{d_x})$ linearly.
In general, variable selection is notoriously difficult in the nonparametric setting \citep{xu2016faithful}, when the dependence of $Y$ on $(X_1,\dots,X_{d_x})$ can be arbitrary.
The difficulty lies in the potentially ``local'' behavior of a nonparametric function.
Some variables may be irrelevant in some regions and affect the value of $Y$ significantly elsewhere.
One idea is to focus on the neighborhood of a given point and select relevant variables locally.
For instance, \citet{lafferty2008rodeo} propose a RODEO (regularization of derivative expectation operator) algorithm which identifies the relevant variables by adjusting the bandwidth of a local linear regression. A recent work \citep{giordano2020grid} improves RODEO by further distinguishing the linear dependent variables from the nonlinear ones.
\citet{bertin2008selection} apply LASSO to the observations locally near the given point.
They provide consistency and finite sample bound when selecting variables in this way.\footnote{Part of our algorithm is motivated by this work, but we improve their theoretical performance, see Remark \ref{rmk:Compare-Bertin08}.}
\citet{miller2010local} discuss several local variable selection methods.
It is not clear how to obtain a global sparsity structure from these methods, since locally the set of relevant variables may differ from region to region.
The local methods also suffer from high dimensionality, as the observations in a neighborhood in a high-dimensional space are rather scarce.
Although our algorithm builds on this idea, we provide an approach to aggregate the local predictions and create a global variable selector, which has a much better performance in high dimensions.



In this literature, the setup in \citet{comminges2011tight} is closest to this study.
They develop a procedure focusing on the Fourier coefficients of the function and show that the relevant variables can be selected with high probability. Our study differs from theirs in the assumptions, algorithms, and also the theoretical performances. Most importantly, the goal of our study is to provide an algorithm with theoretical guarantees which can be implemented (See Section \ref{sec:numerical} for numerical experiments). For their work, it's acknowledged in \citep{giordano2020grid} that ``the procedure is only of theoretical interest and no implementation is given''.

Other papers use the Reproducing Kernel Hilbert Space (RKHS) to represent nonparametric functions and conduct variable selection \citep{rosasco2013nonparametric, ye2012learning, yang2016model, he2018scalable}.
The choice of the kernel crucially determines the class of the functions.
In a recent paper \citet{xu2016faithful} study the problem assuming the reward function $\EX[Y]=f(X_1,\dots,X_{d_x})$ is convex and sparse.
Different from these approaches, we do not impose kernel structures or shape constraints, and only assume more general structures such as continuity and smoothness.

Compared to the literature, the objective and method in this study are different.
First, we do not allow $d_x$ to scale with the number of observations, which is the focus of many studies in statistics.
Moreover, besides selecting relevant variables, we do not want to recover the functional form $f(X_1,\dots,X_{d_x})$, which is the goal of sparse regression.
They allow us to derive a strong theoretical guarantee and achieve near-optimal regret for online learning.
Second, we provide a method called weighted voting, which effectively aggregates the information of local variable selections.
It improves the localized methods in the literature and may be of independent interest.

\subsection{Contextual Bandits}\label{sec:literature-bandit}
The literature on contextual bandits studies adaptive data collection and sequential decision-making (see \citealt{bubeck2012regret} for a complete bibliography).
Many papers in this area consider linear reward in the covariates (see, e.g., \citealt{li2010contextual}).
Among them, the sparsity structure of the contextual/covariate space has been studied by \citet{carpentier2012bandit,deshpande2012linear,abbasi2012online,gilton2017sparse}.
To our knowledge, \citet{bastani2020online} are the first to use the LASSO estimator to identify the sparsity.
Under so-called margin conditions, they propose the ``LASSO bandit'' algorithm, which obtains the regret $O((d^*_x)^2 (\log T +\log d_x)^2)$ almost only dependent on the effective dimension $d_x^{*}$, compared with the regret bound $O(d_x^3 \log T)$ of linear bandits without sparsity \citep{goldenshluger2013linear}. So the performance improves significantly if $d_x^*\ll d_x$. After that, \citet{wang2018minimax} improves the regret to $O((d^*_x)^2 (\log d_x +d^*_x) \log T)$ by adopting minimax concave penalized technique. Additionally, when no margin condition exists, \citet{kim2019doubly,ren2020dynamic} develop LASSO estimator based algorithms achieving the regret $\tilde{O}(d^*_x \sqrt{ T})$ and $\tilde{O}(\sqrt{d^*_x T})$.
Recently, \cite{oh2020sparsity} propose an algorithm solving the issue that the sparsity index $d_x^*$ is not available in practice.
However, these methods are not applicable to the nonparametric setting that we consider in this paper.
On one hand, there is no variable selection algorithm that is as powerful as LASSO in nonparametric settings. On the other hand, variable selection is particularly important for nonparametric online learning because the regret grows exponentially in the covariates dimension $d_x$.
As a result, efficient nonparametric variable selection is both challenging and important.
In this paper, we design new algorithms with a nonparametric setup and theoretically prove that the dependence of the regret on $d_x$ can be reduced to $d_x^*$ for online learning.

There are studies on nonparametric contextual bandits with finite arms and continuous reward functions \citep{yang2002randomized, rigollet2010nonparametric,perchet2013multi,qian2016kernel}.
A similar stream of literature studies the continuum-armed bandits, where the arm/decision space is continuous just like the contextual space \citep{agrawal1995continuum,kleinberg2005nearly,auer2007improved,kleinberg2008multi,kleinberg2010sharp,bubeck2010x,magureanu2014lipschitz}.
A common result in the literature is that for continuous reward functions\footnote{For reward functions with a higher order of smoothness, the regret may be lower. See \citet{hu2019smooth,gur2019smoothness}.}, the regret depends exponentially on $d_x$.
For example, \citet{lu2009showing,slivkins2014contextual}
present a uniformly partition and a zooming algorithm for reward functions that are Lipschitz continuous in both the decision and covariate.
Both algorithms attain near-optimal regret $\tilde{O}(T^{1-1/(d_x+d_y+2)})$, where $d_x$, $d_y$ are the dimensions of the covariate and decision space. Recently, \citet{reeve2018k,guan2018nonparametric} develop $k$-Nearest Neighbour ($k$-NN) based algorithms to address the dimensionality issue.
Their algorithms automatically take advantage of the situations where the covariates are supported on a metric space of a lower effective dimension, such as a low-dimensional manifold embedded in a high dimensional space.
However, they cannot be used to identify the sparsity structure.
Our study attempts to lift the curse of dimensionality in the regret, particularly the exponential dependence on $d_x$.
To the best of our knowledge, this is the first work to address the dimensionality issue in nonparametric contextual online learning by taking advantage of the sparsity structure.
\label{page:literature-continuum}
{\RV \Copy{cop:literature-continuum}{  Note that we formulate the problem for continuum-armed bandits since the assumption (Lipschitz continuous reward) is relatively simple and general. Our approach can also be extended to discrete arms, if some technical conditions, such as the margin condition in \citep{perchet2013multi}, are satisfied.}}
Our work contributes to the contextual bandits literature by providing a general recipe to mitigate the curse of dimensionality for online learning.

\label{page:literature-decision}
{\RV \Copy{Cop:literature-decision}{While we focus on the sparsity in the covariate space, there are recent studies that focus on the dimension reduction of the decision/arm space \citep{djolonga2013high,tyagi2013continuum, kwon2017sparse,kwon2016gains}.
It turns out that if the reward function $f$ is concave in $\bm y$, then algorithms can be developed to achieve regret $\tilde{O}(d_yT^{(d_x+1)/(d_x+2)})$ \citep{li2019dimension,cesa2017algorithmic} that scales linearly instead of exponentially with $d_y$.
Although the approaches are different, our paper can complement this stream of literature:
applying the variable selection algorithm in our paper as a subroutine, their algorithms can also achieve even smaller regret $\tilde{O}(T^{(d_x^*+1)/(d_x^*+2)})$ under covariate sparsity.}}


\subsection{Dynamic Pricing with Demand Learning}
Our paper is also related to the literature on personalized dynamic pricing with demand learning  \citep{besbes2009dynamic,keskin2014dynamic,den2014simultaneously,den2015dynamic}. In this stream of literature, demand functions are typically assumed to be linear in prices and consumer features (covariates).
\citet{qiang2016dynamic} show a myopic pricing policy can exhibit near-optimal revenue performance with regret $O(d_x \log T)$. \citet{cohen2020feature} find a multi-dimensional binary search algorithm for adversarial features, which has regret $O(d_x^2 \log(T/d_x))$.
\citet{javanmard2019dynamic} consider the sparsity structure of features and propose a pricing policy achieving regret $O(d^*_x \log d_x \log T)$.
\citet{ban2020personalized} take into account the feature-dependent price sensitivity and 
show a minimax regret $O(d_x^* \sqrt{T}(\log d_x+\log T))$.
In the studies above, the dependence of regret on $d_x$ or $d_x^*$ is not exponential as the demand is assumed to have a parametric (linear) form.

Going beyond the parametric setting, 
\citet{chen2021nonparametric} propose a nonparametric pricing policy achieving a near-optimal regret $O((\log T)^2 T^{(2+d_x)/(4+d_x)})$, which indeed depends on $d_x$ exponentially.
A similar dependence is found in network revenue management \citep{besbes2012blind} in which the dimension of the decision space $d_y$ appears in the regret $O(T^{(2+d_y)/(3+d_y)})$.
Therefore, the dimension of the covariate significantly complicates the learning problem in the nonparametric formulation.
Our work proposes a dimension reduction method that significantly mitigates the dimensionality problem.
Although we formulate the problem for online learning in general, our approach is applicable to dynamic pricing with consumer features.

\section{Problem Formulation}\label{sec:problem}
We now formulate the online learning problem.
We define the decision and covariate space as $\XX\coloneqq [0,1]^{d_x}$ and $\YY\coloneqq [0,1]^{d_y}$.
Let $\mathcal{T}=\{1,2,\dots,T\}$ denote the sequence of decision periods faced by the decision maker.
At the beginning of each period $t \in \mathcal{T}$, the covariate $\bm{X}_t \in \XX$, drawn independently from some unknown distribution\footnote{\citet{slivkins2014contextual} assumes that the covariate arrivals $x_t$ are fixed before the first round.
	We follow \citet{perchet2013multi} and assume that $\bm{X}_t$s are i.i.d.}, is revealed to the decision maker.
Then the decision maker chooses a decision $\bm{Y}_t$ in $\YY$.
The reward in period $t$ is a random variable $Z_t$:
\begin{equation*}
Z_t=f(\bm{X}_t,\bm{Y}_t)+\epsilon_t,
\end{equation*}
where $f(\bm{X}_t,\bm{Y}_t)$ is the mean reward function which is unknown. The noises $\epsilon_t$ satisfy the following standard assumption.
\begin{assumption}[Sub-Gaussian]\label{ass:sub_gau}
	The noises $\{\epsilon_t\}_{t=1}^T$ are independent $\sigma$ sub-Gaussian, i.e., for any $\xi \geq 0$,
	\begin{equation*}
	\PR (\epsilon_t \geq \xi) \leq \exp\left(-\frac{\xi^2}{2 \sigma^2}\right).
	\end{equation*}
\end{assumption}

Assumption \ref{ass:sub_gau} is widely used in statistics and many classical distributions are sub-Gaussian, such as any bounded and centered distribution or the normal distribution.

Now we formally define \textit{policy} and \textit{regret} which are critical concepts in designing online learning algorithms.

\textbf{Policy.}
Before making decisions in period $t$, the information revealed to the decision-maker includes observed covariates $\{\bm{X}_s\}_{s=1}^{t}$, the adopted decisions $\{\bm{Y}_s\}_{s=1}^{t-1}$ and the realized rewards $\{Z_s\}_{s=1}^{t-1}$. 
A policy $\pi_t$ is defined as a function mapping the past history to the decision space:
\begin{equation*}
\bm{Y}_t=\pi_t\left(\bm{X}_t,Z_{t-1},\bm{Y}_{t-1},\bm{X}_{t-1},Z_{t-2}, \bm{Y}_{t-2}, \bm{X}_{t-2},\ldots,Z_1,\bm{Y}_1,\bm{X}_1\right).
\end{equation*}

\textbf{Regret.}
If the reward function is known, then the optimal decision and reward given covariate $\bm x$ are
\begin{equation*}
\bm{y}^{\ast}(\bm{x}) \coloneqq \argmax_{\bm{y}\in \mathcal Y} f(\bm{x},\bm{y}),\quad f^{\ast}(\bm{x}) \coloneqq \max_{\bm{y}\in\mathcal Y} f(\bm{x},\bm{y}),
\end{equation*}
Since the decision maker does not have access to the unknown reward function, the total expected reward of any policy $\pi$ is always lower than $\sum_{t=1}^T \mathbb{E}[f^*(\bm{X}_t)]$.
A standard performance measure of a policy is defined as the expected gap between the reward with known $f$ and the reward under policy $\pi$, aggregated over the entire time horizon, i.e.,
\begin{equation*}
R_{\pi} (T) \coloneqq \sum_{t=1}^T \mathbb{E} \left[f^{\ast}(\bm{X}_t)-f(\bm{X}_t,\pi_t)\right].
\end{equation*}
For the decision maker, the objective is thus to design a policy that achieves small regret for a class of functions $f$.

\begin{remark}[Motivating Problem]\label{rmk:dynamic-pricing}
	To motivate the formulation, consider the following example of personalized dynamic pricing.
	An online retailer sets personalized prices for an assortment of products to consumers with observable features such as education backgrounds, incomes, occupations, etc.
	The demand for the products depends not only on the prices, but also on the personal information.
	The retailer observes the information of each arriving customer ($\bm{X}_t$), decides personalized prices ($\bm{Y}_t$) accordingly, and observes the revenue ($Z_t$). {\RV \Copy{cop:example-pricing}{The revenue ($Z_t$) is the product of demand and prices.}}
	If the relationship ($f$) between customers' information, prices and revenue is unknown to the retailer,
	then it has to be learned from historical observations and the goal is to maximize the long-run revenue.
\end{remark}

A standard assumption in online learning of nonparametric functions is that $f(\bm x,\bm y)$ is continuous, as it is virtually impossible to learn $f$ if it can be arbitrarily discontinuous.
Therefore, we assume that
\begin{assumption}[Continuously Differentiable]\label{ass:continuous}
	The function $f(\bm x,\bm y)$ is continuously\\ differentiable.
\end{assumption}

Under a slightly weaker assumption that $f(\bm{x},\bm{y})$ is Lipschitz continuous in both $\bm{x}$ and $\bm{y}$, the optimal rate of regret is (see, e.g., \citealt{slivkins2014contextual})
\begin{equation} \label{eq:formu_lb}
\min_{\pi} \sup_{f} R_{\pi}(T) \geq  \Omega(T^{1-1/(2+d_x+d_y)}).
\end{equation}
The lower bound here reflects the curse of dimensionality in nonparametric online learning.
The regret grows almost linearly in $T$ for large $d_x$ and $d_y$. For example, if $d_x=d_y=5$, then $R_{\pi}(T) \geq \Omega(T^{\frac{11}{12}})$, which is much worse than $\Omega(\sqrt{T})$, the typical lower bound in the parametric setting.
Since the regret in \eqref{eq:formu_lb} cannot be further improved under the assumption that $f$ is Lipschitz continuous, the dependence on dimensionality looks dire.
We next introduce a sparsity structure on the covariate space that may remedy the high dimension $d_x$.
In this paper, as we focus on the dimension reduction in the covariate space, we set $d_y=1$ in the rest of the paper for the ease of exposition.
All the results can be generalized to the cases where $d_y>1$.

\subsection{Assumptions on the Sparsity Structure}

In many practical cases, not all the variables in the covariate have an impact on the value of $f$.
In other words, out of $d_x$ variables in the covariate, many are redundant.
Such sparsity has been one of the central topics in statistics.
More precisely, we consider
\begin{assumption}[Sparse Reward Function] \label{ass:spa_cov}
	There exists $d_x^*\le d_x$, a subset $J=\{i_1,\ldots,i_{d_x^*}\}\subset \left\{1,\dots,d_x\right\}$, and a function $g:[0,1]^{d_x^*} \mapsto \mathbb{R}$ such that for all $\bm{x}=(x_1,\ldots,x_{d_x}) \in \XX$ and any $y \in \YY$,\footnote{Since $d_y=1$, we use a scalar $y$ instead of a vector $\bm y$ throughout the paper.} we have
	\begin{equation*}
	f(x_1,\dots,x_{d_x},y)=g(x_{i_1},\dots,x_{i_{d^*}},y).
	\end{equation*}
\end{assumption}

Assumption~\ref{ass:spa_cov} gives a rigorous definition of the sparsity.
We refer to the variables in $J$ as \textit{relevant} variables and those in $J^c\coloneqq\left\{1,\dots,d_x\right\}\setminus J$ as \textit{redundant} variables.
With a slight abuse of notations, we denote $J^{(i)}=1$ if $i \in J$ and $J^{(i)}=0$ otherwise.  Since the redundant variables do not affect $f$, their partial derivatives are always zero:
\begin{equation*}
J^c=\left\{i \in \{1,2,\ldots,d_x\}: \frac{\partial f(\bm{x},y)}{\partial x_i}=0, \quad \forall \bm{x} \in \XX, \forall y \in \YY \right\}.
\end{equation*}

However, in the nonparametric setting, Assumption~\ref{ass:spa_cov} alone is not sufficient to characterize the sparsity structure.
Suppose $f$ changes slightly along the direction of $x_1$, only when $y$ is in a small region. For example,
\begin{equation*}
f(\bm x, y)=g(x_2,\dots,x_{d_x}, y) \!+\! \Ind{0 \!\le\! y\!\le\! \frac{\epsilon}{2}}(3 \epsilon y^2-4y^3)x_1 \!+\! \Ind{\frac{\epsilon}{2} \!<\! y \!\le\! \epsilon} (3 \epsilon (\epsilon-y)^2-4(\epsilon-y)^3)x_1,
\end{equation*}
for an arbitrarily small $\epsilon>0$. The function $f$ satisfies Assumption \ref{ass:continuous} if $g$ is continuously differentiable. We see that $x_1$ plays a role when $y \leq \epsilon$, and technically speaking, it is a relevant variable. However, it is almost impossible for any methods to detect the relevance of $x_1$, since the partial derivatives $\partial f(\bm{x},y) / \partial x_1$ diminish for infinitesimal $\epsilon$. 
To resolve this issue, we impose a stronger assumption that $\partial f(\bm{x},y) / \partial x_i$ is non-vanishing for all $y \in \YY$ and all $i \in J$.


{\RV
\begin{assumption}[Global Relevance] \label{ass:pos_gra}
	There exists a constant $C>0$ such that
	\begin{equation}\label{equ:global-C}
	\left|\frac{\partial f(\bm{x},y)}{\partial x_i}\right| \geq C, \quad \forall i \in J, \bm{x} \in \XX, y \in \YY.
	\end{equation}

\end{assumption}
}
Assumption~\ref{ass:pos_gra} states that the relevant variables must play a \textit{global} role, not only for all $y \in \YY$, but also for all $\bm{x} \in \XX$.
Their partial derivatives are non-vanishing everywhere.
Note that Assumption~\ref{ass:pos_gra} includes functions that do not belong to any parametric family.
For example, the variables are allowed to have complex interactions.

\Copy{cop:global-rel-lit}{\RV{In the literature, some studies impose a similar global structure on the function.
For example, \citet{xu2016faithful} assume that $f$ is convex and \citet{rosasco2013nonparametric, ye2012learning, yang2016model, he2018scalable} assume $f$ in RKHS. These global asssumptions are typically stronger than Assumption~\ref{ass:pos_gra}.
}}

For certain applications, Assumption~\ref{ass:pos_gra} may be too strong, especially when some relevant variables are relevant locally but not globally in $\XX$.
Considering the dynamic pricing example,
even the variables that strongly predict consumer behavior are not always relevant.
For instance, the demand for a product may be significantly increased when the income ranges from ``low'' to ``medium'', while the income level becomes almost irrelevant when it is above a certain threshold.
Technically speaking, the partial derivatives are not always bounded away from zero,
in which case Assumption~\ref{ass:pos_gra} may fail.
To make our approach more practical, we relax Assumption~\ref{ass:pos_gra} below.

{\RV
\begin{lastassumption}{ass:pos_gra}[Local Relevance]\label{ass:gra-point}
	There exists a constant $C>0$
	such that
	\begin{equation}\label{equ:hypercube-H-C}
	\left|\frac{\partial f(\bm{x},y)}{\partial x_i}\right| \geq C, \quad  \forall i \in J,\bm{x} \in \mathcal{H}_i, y \in \YY,
	\end{equation}
	where $\mathcal{H}_i \subset \XX$ is a hypercube centred at $\bm{x}_{(i)}$.

\end{lastassumption}
}
Assumption \ref{ass:gra-point} is much weaker than Assumption~\ref{ass:pos_gra}.
For $i \in J$, it assumes non-vanishing partial derivatives at one point $\bm x_{(i)}$ in the domain, for all $y$.
By Assumption~\ref{ass:continuous}, Assumptions~\ref{ass:pos_gra} and~\ref{ass:gra-point} can be satistied by a simpler condition.
{\RV
\begin{lemma}[Generality of Assumptions \ref{ass:pos_gra} and \ref{ass:gra-point}]\label{lem:existence-C}
	\begin{enumerate}
		\item 	Under Assumptions~\ref{ass:continuous} and
		\begin{equation*}
			\frac{\partial f(\bm{x},y)}{\partial x_i} \neq 0, \quad \forall{\bm{x} \in \XX, y \in \YY, i \in J},
		\end{equation*}
		Assumption \ref{ass:pos_gra} holds.
		\item Suppose Assumptions~\ref{ass:continuous} holds and $f$ is twice-differentiable with respect to $\bm{x}$. In addition,
		for all $i\in J$, there exists $\bm{x}_{(i)} \in \XX $ such that
		\begin{equation*}
		\frac{\partial f(\bm{x}_{(i)},y)}{\partial x_i} \neq 0, \quad \forall y\in \YY.
		\end{equation*}
		Then Assumption \ref{ass:gra-point} holds.
	\end{enumerate}
\end{lemma}
Due to Lemma \ref{lem:existence-C}, Assumptions \ref{ass:pos_gra} and \ref{ass:gra-point}, especially Assumption~\ref{ass:gra-point}, hold for most functions that are used in practice.
}

For exposition, we first introduce our algorithm that works for Assumption~\ref{ass:pos_gra} in Section~\ref{sec:monotone_algorithm} and Section~\ref{sec:dec_spa}.
Then we show that with some adjustment, the algorithm has the same theoretical guarantee under Assumption~\ref{ass:gra-point} in Section \ref{sec:extensions}.

\subsection{Online Learning with Nonparametric Variable Selection}\label{sec:overview}
If the set of relevant variables $J$ were known a priori, then the decision maker would discard the redundant variables
and apply online learning algorithms only for the effective variables with dimension $d_x^*$.
For example, existing algorithms for contextual bandits in nonparametric settings \citep{kleinberg2005nearly,lu2009showing,slivkins2014contextual} can achieve the near-optimal regret of the order $\tilde O(T^{1-1/(d^*_x+3)})$. (Recall that we set $d_y=1$.)

We propose a two-phase approach to handle the problem.
In particular, we design a subroutine to select variables before applying the online learning algorithms.
We hope to collect data to provide an estimated set of relevant variables, $\hat J$, within the first $n<T$ periods.
If $\hat J=J$ with high probability and $n\ll T$, then the online learning algorithms can be executed as if $J$ were known and the regret does not deteriorate significantly.
We elaborate this idea below.

\textbf{Variable Selection Phase.}
We refer to the first $n$ periods devoted to variable selection as the variable selection phase.
In this phase, the main goal of the algorithm is to correctly identify the set of relevant variables $J$ with high probability.
By Assumption \ref{ass:spa_cov}, the sparsity structure remains identical for all $y$.
Therefore, in this phase, the decision maker may simply use a fixed decision $y \in \YY$.

Therefore, the observed reward is generated by
\begin{equation}\label{equ:fix-decision}
Z_t = f(\bm X_t, y)+\epsilon_t, \quad t=1,\dots,n.
\end{equation}
Our goal is to use $\left\{(\bm X_t, Z_t)\right\}_{t=1}^n$ to select relevant variables.
We describe the variable selection algorithm in details in Section~\ref{sec:monotone_algorithm}.

\textbf{Online Learning Phase.}
We refer to the remaining $T-n$ periods as the online learning phase.
Given that the relevant variables in the covariate have been correctly identified,
we may apply the existing algorithms \citep{kleinberg2005nearly,lu2009showing,slivkins2014contextual} for contextual bandits.
Denote the expected cumulative regret in the remaining $T-n$ periods as $R_2(T-n)$.
On the correctly selected covariate space, the Uniform algorithm in \citet{kleinberg2005nearly,lu2009showing} and the Contextual Zooming algorithm in \citet{slivkins2014contextual} can achieve regret
\begin{equation}\label{eq:learning-regret}
R_2(T-n)=O\left((T-n)^{1-1/(d_x^*+3)} \log(T-n) \right).
\end{equation}
We may use either as a subroutine in the online learning phase.

{\RV
\begin{remark}[Fixed Decision and Separated Phases]
	It is not uncommon for the decision maker to commit to a fixed decision starting online learning.
	For instance, when the market condition shifts and the firm needs to use online learning to learn the new pricing policy,
        it may start from the incumbent pricing policy, which has proved to perform reasonably well in the past, before exploring risky policies.
        This initial phase of ``cautious price stickiness'' can be used for variable selection.
	Moreover, in Section~\ref{sec:rule-stagewise}, we extend the algorithm to allow the decision $y$ to be sampled from a probability distribution.
        We also propose an algorithm that integrates variable selection and online learning.
        For the ease of exposition, we focus on the two-phase approach here and defer the extensions to Section \ref{sec:rule-stagewise}.
\end{remark}
}

\textbf{Combined Regret.}
The cumulative regret of the two phases depends on the probability of successful variable selection in the first phase and the regret of the subroutine in the second phase.
More precisely, the expected cumulative regret of our algorithm over $T$ periods is
\begin{equation*}
R_{\pi}(T) \leq 2 n \max_{\bm x\in \XX, y\in \YY}|f(\bm x,  y)|+ \PR (\hat{J}=J) R_2(T-n) + 2 \max_{\bm x\in \XX, y\in \YY}|f(\bm x,  y)| \PR (\hat{J} \neq J) (T-n).
\end{equation*}
The first term reflects the regret incurred in the variable selection phase, because $f^{*}(\bm{X}_t)-f(\bm{X}_t,  y)\le 2 \max_{\bm{x},y}|f(\bm x,  y)|$ in a single period.
The regret in the online learning phase combines two scenarios: a ``good'' event that the variable selection phase correctly identifies the relevant variables and a ``bad'' event, where incorrect variable selection leads to linearly growing regret.
The following proposition shows a sufficient condition for the total regret of both phases to achieve the optimal rate of regret.

\begin{proposition}\label{prop:rate_selection}
	If $n \leq T^{1-1/(d^*_x+3)}$ and $\PR (\hat{J} \neq J) \leq n^{-1/(d^*_x+2)}$, then we have
	$R_\pi(T)=$
	$O\left(T^{1-1/(d_x^*+3)} \log(T) \right)$.
\end{proposition}
The proposition provides a guideline for the algorithmic design of the variable selection phase.
In the next two sections, we elaborate on the details.
\begin{remark}[The Case of $d_y>1$]\label{rem:stronger-dy}
	If the dimension of decision space is $d_y$, then Proposition \ref{prop:rate_selection} can be modified as:
	If $n \leq T^{1-1/(d^*_x+d_y+2)}$ and $\PR (\hat{J} \neq J) \leq n^{-1/(d^*_x+d_y+1)}$,
	we have
	$R_\pi(T)=
	O\left(T^{1-1/(d_x^*+d_y+2)} \log(T) \right)$.
\end{remark}

\section{Variable Selection for Global Relevance}\label{sec:monotone_algorithm}
In this section, we propose a new variable selection algorithm under Assumption~\ref{ass:pos_gra}, which is referred to as ``Binning and Voting LASSO'' (BV-LASSO).
The algorithm utilizes the idea of LASSO, a well-known method in statistics and machine learning, to achieve nonparametric variable selection and thus dimension reduction.

For linear models, LASSO has proved to be extremely successful in practice with strong theoretical guarantees and computational efficiency \citep{zhao2005boosted,zhao2006model}.
If applied to our data, the standard LASSO estimator solves the following problem:
\begin{equation}\label{equ:standard_lasso}
(\theta_0,\bm{\theta}^{lasso})=\argmin_{\theta_0,\bm\theta}\left\{\frac{1}{n} \sum_{t=1}^{n} \left( Z_t- \theta_0-\bm{X}_t^T\bm{\theta} \right)^2 +2\lambda \|\bm{\theta}\|_{1}\right\},
\end{equation}
where the hyper-parameter $\lambda$ penalizes the $\ell_1$-norm of the parameter $\bm\theta$.
The basic intuition of LASSO is that the $\ell_1$ loss function creates sparsity.
If $f$ is a linear function, then with properly chosen $\lambda$, the estimators $\bm\theta^{lasso}_i$ of redundant variables $x_i$ tend to be zero, while the estimators of relevant variables remain non-zero with high probability.
As a result, the set of relevant variables can be identified from the sign of $\bm\theta^{lasso}$.

However, in our setting $f$ is not necessarily linear and LASSO may fail.
For example, consider $d_x=d_y=1$ and $f(x_1,y)=(x_1-0.5)^2$ where $X_1$ has a uniform distribution in $[0,1]$.
The LASSO estimator $\theta_1$ returns zero, because it is the best linear estimator for the quadratic function,
thus falsely ruling out the relevant variable $x_1$.
On the flip side, LASSO may also return false positives for nonlinear functions, identifying redundant variables as relevant.
For instance, consider $d_x=3$, $d_y=1$ and let $X_1\sim~U[0,1]$, $X_2\sim~U[0,1] \perp X_1$, $X_3=0.5 X_1+0.5 X_2$.
{\RV \Copy{cop:example-LASSO-fail}{If the reward function is nonlinear $f(x_1,x_2,y)=-x_1+e^{2 x_2}$,
then $X_3$ would be identified as relevant by LASSO.
In particular, because of the correlation between $X_3$ and $X_2$ and the nonlinearity of $X_2$,
LASSO would return a linear model with a non-zero coefficient for $X_3$.
Note that this failure occurs even when $f$ satisfies Assumption \ref{ass:pos_gra}.}
}

Having highlighted the technical difficulties, we introduce two mild technical assumptions required for our algorithm.
\begin{assumption}[Second-order Smoothness] \label{ass:smooth_cov}
	The function $f$ is twice-differentiable with respect to $\bm{x}$, i.e., there exists $L>0$ such that
	\begin{equation*}
	|f(\bm{x_1},y)-f(\bm{x_2},y)-\nabla_{\bm{x}} f(\bm{x_2},y)^T (\bm{x_1}-\bm{x_2})| \leq L \|\bm{x_1}-\bm{x_2}\|_{\infty}^2,
	\end{equation*} for all $\bm{x_1},\bm{x_2} \in \XX, \ y \in \YY$.
\end{assumption}

Assumption~\ref{ass:smooth_cov} imposes the smoothness condition of $f$ and is widely adopted in many problems in statistics and optimization.
It allows for a second-order approximation for $f$ in a small area.
An implication of Assumption~\ref{ass:smooth_cov} is that the infinity norm of the Hessian matrix with respect to $\bm x$ is bounded by $2L$.

Next, we impose an assumption on the distribution of the covariate.
\begin{assumption}[Regular Covariate] \label{ass:cov_des2}
	The covariate $\bm X\in \XX$ has a probability density function $\mu(\bm{x})$ and there exist $\mu_m, \mu_M ,L_{\mu} >0 $ such that
	\begin{enumerate}
		\item $\mu_m \leq \mu(\bm{x}) \leq \mu_M$ for all $\bm{x} \in \XX$,
		\item The density function $\mu$ is $L_{\mu}$-Lipschitz, i.e., $\mu(\bm{x})-\mu(\bm{x}') \leq L_{\mu} \|\bm{x}-\bm{x}'\|_{\infty}$ for all $\bm{x},\bm{x}' \in \XX$.
	\end{enumerate}
\end{assumption}

Assumption~\ref{ass:cov_des2} imposes bounded and continuous density functions and is easy to satisfy in many cases.
There is a more general and less interpretable version of Assumption~\ref{ass:cov_des2}, which we defer to Appendix~\ref{app:irrepresentable} for the exposition.

Now that we have introduced all the assumptions, next we propose the BV-LASSO algorithm.
Before describing our algorithms in detail, we remark on the information available to the decision maker initially: the decision maker knows $T$, $d_x$,
$\sigma$, $\mu_m$, $\mu_M$, and $L$ but doesn't know $J$, $L_{\mu}$ or $C$. \label{page:emphasize-sparsity}{\RV \Copy{cop:emphasize-sparsity}{In particular, we do not use the information of $d_x^*$ in any step of the algorithm.}}

\subsection{Binning and Local Linear Approximation}

We first partition the covariate space regularly into $k^{d_x}$ hypercubes (bins), each with side length $h=1/k$, denoted by
\begin{equation*}
\mathcal{B}_h=\{B_j| \ j={1,2,\ldots,h^{-d_x}}\}.
\end{equation*}
The intuition is that, although $f$ is nonlinear, it can be approximated by a linear function in a small bin by the Taylor series expansion.
The approximation error can be controlled by the size of the bins.
More importantly, the approximation error becomes small \textit{relative to} the statistical error of LASSO when the side length $h$ is small enough.

To formalize the intuition, for a given bin $B$, we project the function $f$ to the functional vector space spanned by linear functions of the variables for a fixed $y$ (we omit the dependence on $y$ if it doesn't cause confusions):
\begin{equation}\label{equ:L2}
\theta_0=\int_{\bm{x} \in B} f(\bm{x},y) \dif \bm{x},\ 	\theta_i=\dfrac{\int_{\bm{x} \in B} [f(x_1,\dots,x_{d_x},y)-\theta_0] x_i \dif \bm{x}}{\int_{\bm{x} \in B} x_i^2 \dif \bm{x}}, \quad \text{for} \ i=1,2,\ldots,d_x
\end{equation}
The projection $\theta_0+\sum_{i=1}^{d_x} \theta_i x_i$ is the ``best'' linear approximation for $f(\bm x, y)$ with respect to the integrated squared error, i.e.,
\begin{equation*}
(\theta_0,\theta_1,\ldots,\theta_{d_x})=\displaystyle{\argmin_{\theta_0,\theta_1,\ldots,\theta_{d_x}} \int_{\bm{x} \in B} \left(f(x_1,\ldots,x_{d_x},y)-\theta_0-\sum_{i=1}^{d_x} \theta_i x_{i} \right)^2\ \dif x_1\ldots \dif x_{d_x}}.
\end{equation*}

If the sparsity structure of the projection maintains that of the original function $f$, then we may attempt to run LASSO on the projection and recover the sparsity of $f$.
To do so, we need to calibrate the approximation error, in order to compare it with the statistical properties of LASSO later.
The following lemma provides such calibration.
\begin{lemma}[Linear Approximation Error in a Bin]\label{lem:proj}
	Suppose $(\theta_0,\theta_1,\dots,\theta_{d_x})$ are the coefficients of the linear projection of $f$ in $B$ shown in \eqref{equ:L2}. Under Assumptions~\ref{ass:continuous}, \ref{ass:spa_cov}, \ref{ass:pos_gra} and~\ref{ass:smooth_cov}, we have
	\begin{enumerate}
		\item $|\theta_i| \geq C$ for any $i \in J$ and $|\theta_i|=0$ for any $i \notin J$, where $C$ is a constant satisfying \eqref{equ:global-C}.
		\item $|f(\bm x,y)-\theta_0-\sum_{i=1}^{d_x} \theta_i x_i| \leq (4\sqrt{3}+1) L d_x h^2$ for all $\bm{x}=(x_1,\ldots,x_{d_x}) \in B$, where the constant $L$ is presented in Assumption~\ref{ass:smooth_cov}.
	\end{enumerate}
\end{lemma}

The first point of the lemma shows that the linear projection maintains the sparsity structure of $f$.
More importantly, it doesn't diminish the partial derivatives.
The second point shows that the approximation error of the linear approximation is $O(h^2)$.
This is crucial in the subsequent analysis, as we would like to control the bias or the approximation error of LASSO by the bin size.

\label{page:empirical-l2}
{\RV \Copy{cop:empirical-L2}{
Note that the values of the coefficients in the linear approximation in \eqref{equ:L2} are not used in the subsequent analysis.
We merely check if they are statistically nonzero to identify the sparsity structure.
Using their values directly may lead to biased estimates of the nonparametric reward function and suboptimal regret.
}}

\subsection{Localized LASSO}\label{sec:local-LASSO}
Next, we apply LASSO to a given bin $B_j$.
Suppose there are $n_j$ periods in which the generated covariate falls in $B_j$.
With a slight abuse of notation, let $\bm X_t\in B_j$ for $t=1,2,\dots, n_j$.
We first \textit{normalize} the data by defining
\begin{equation}\label{eq:normalize}
\bm U_t \coloneqq (\bm X_t-C_{B_j})/h
\end{equation}
where $C_{B_j}$ is the geometric centre of $B_j$.
The LASSO selector for $B_j$ solves the penalized least square problem and identifies the non-zero coefficients:
\begin{equation} \label{equ:lasso}
\hat{J}_j=\supp\left\{ \argmin_{\theta_0,\bm{\theta}}\left\{\frac{1}{n_j} \sum_{t=1}^{n_j} \left( Z_t- \theta_0-\bm{U}_t^T \bm{\theta} \right)^2 +2\lambda \|\bm{\theta}\|_{1}\right\} \right\},
\end{equation}
where the operator $\supp$ selects the subset of $\bm{\theta}$ that are non-zero{\RV \footnote{ \Copy{cop:lasso-theta0}{Note that $\theta_0$ is the intercept term in LASSO regression. It does not matter whether $\theta_0$ is zero.}}}.
Note that the normalization is an affine mapping and thus doesn't change $\hat J_j$ as long as $\lambda$ is properly scaled.
Indeed, we normalize in order to keep a constant $\lambda$ that does not scale with $h$ in the analysis.

Our hope is that $\hat J_j$ would be identical to $J$ for small $h$.
As shown in the second point of Lemma~\ref{lem:proj}, the approximation error is $O(h^2)$.
If LASSO selects the relevant variables for the linear projection when the approximation error is small,
then $\hat J_j=J$ because of the first point of Lemma~\ref{lem:proj}.
This intuition is formalized below.

\begin{proposition}[Variable Selection by Localized LASSO]\label{prop:subset}
	For a given bin $B_j$ of side length $h$, under Assumptions~\ref{ass:sub_gau}, \ref{ass:continuous}, \ref{ass:spa_cov}, \ref{ass:pos_gra}, \ref{ass:smooth_cov}, \ref{ass:cov_des2}, and $h\leq b_3$, choosing $\lambda=b_2 h^2$ in \eqref{equ:lasso}, we have \footnote{ \RV \Copy{cop:local-LASSO-condition}{Strictly speaking, the probability here is conditional on the $n_j$ covariates falling in $B_j$. The rigorous definition is deferred to Appendix \ref{proof:subset}.}}
	\begin{equation}\label{equ:prop_subset}
	\PR \left(\hat{J}_j=J\right) \geq 1- p_j,
	\end{equation}
	where $p_j \coloneqq b_0 \exp(-b_1 n_j h^{4})$ and the constants $b_0$, $b_1$, $b_2$, and $b_3$ are presented in Section~\ref{sec:dec_spa}.
\end{proposition}

\begin{remark}\label{rmk:Compare-Bertin08}
	\citet{bertin2008selection} apply LASSO in a neighborhood of a given point to locally select variables.
        They prove that the false selection probability converges to zero at the rate $O(-\exp(n_j h^{d_x+2}))$. In Proposition \ref{prop:subset}, we improve the rate to $O(-\exp(n_j h^{4}))$.
        The distinction between $h^2$ and $h^4$ is caused by the different assumptions on the order of smoothness.
\end{remark}
{\RV
\begin{remark}[Alternatives to LASSO]\label{rmk:local-alternative}
We point out that OLS and thresholding may serve the same purpose to LASSO.
More precisely, one may apply OLS to the data $\{\bm U_t, Z_t\}_{t=1}^{n_j}$ and compare the estimated coefficients to a threshold.
A variable is identified as relevant if the absolute value of the coefficient is greater than the threshold.
We can prove that this alternative method can identify the relevant variables in a similar form to \eqref{equ:prop_subset} with different constants.
In Section~\ref{sec:OLS-threshold}, we provide a rigorous proof and discuss the differences between the two methods.
\end{remark}
}
Proposition~\ref{prop:subset} provides an accurate characterization of the probability of $\hat J_j=J$.
In particular, $h$ needs to be less than $b_3$, which itself depends on other constants.
For example, it is understandable that if $C$ is large, then $J$ is easier to identify and the requirement $b_3$ can be larger.
Once $h$ is sufficiently small, the probability of $\hat J_j\neq J$ diminishes exponentially in $n_jh^4$.
Proposition~\ref{prop:subset} serves as the backbone of the analysis of our algorithm.

Now that we have applied localized LASSO to a single bin, the next question is how to combine them to identify $J$.
Because of the sheer number of bins ($1/h^{d_x}$), it is very unlikely that the sets of selected variables $\hat J_j$ are identical for all $j$ despite the probability guarantee in Proposition~\ref{prop:subset}.
Next we introduce a scheme to aggregate $\hat J_j$ referred to as \textit{weighted voting}.


\subsection{Weighted Voting}\label{sec:voting}

After applying LASSO to all the bins, we have $h^{-d_x}$ selectors $\{\hat{J}_j$,  $j =1,2,\ldots,h^{-d_x}\}$, each representing a set of relevant variables.
A straightforward idea would be to only trust the bin with most observations and use the outcome in that bin as the global selector.
As $n$ increases with $T$, the bin contains at least $nh^{d_x}$ observations and Proposition~\ref{prop:subset} guarantees the correct selection with high probability.
However, this method performs inefficiently in terms of data utilization.
For small $h$, any single bin would contain only a tiny fraction of all the observations $\{\bm X_t\}_{t=1}^n$.
Such waste of data limits its practical use despite the asymptotic properties.

To fully exploit all the observations, we propose the idea of ``weighted voting.''
For variable $x_i$, the outcome of LASSO in bin $B_j$, $\hat{J}_j^{(i)}$,  is binary.
If $i \in \hat{J}_j$, then bin $B_j$ votes ``yes'' for $x_i$ and $\hat{J}_j^{(i)}=1$. Otherwise, the vote is ``no'' and $\hat{J}_j^{(i)}=0$.
If a majority of bins vote ``yes'', then $x_i$ is likely to be relevant.
Moreover, if $B_j$ contains more observations, then we would expect $\hat J_j$ to be more reliable.
This intuition is supported by Proposition~\ref{prop:subset}, as the probability of false selection diminishes in $n_j$.
Therefore, we assign more weights to the votes from the bins with more observations.
In this way, all the observations are exploited as votes from all the bins are aggregated.

Next we describe the details of the procedure.
For $x_i$, consider the linear combination of $\hat{J}_j^{(i)}$ over $j$:
\begin{equation}\label{eq:weighted-vote}
\hat{J}^{(i)}=\sum_{j=1}^{h^{-d_x}} w_j \hat{J}_j^{(i)},
\end{equation}
where the weights $\{w_j\}$ satisfy
$\sum_{j=1}^{h^{-d_x}} w_j=1$ with  $w_j \geq 0$.
If $\hat{J}^{(i)}$ is greater than $1/2$, implying that $x_i$ has a weighted majority of ``yes'' votes, then we classify it as ``relevant''.
Otherwise, we classify it as ``redundant''.
The key questions to address are (1) how to properly choose the weights, and (2) how to control the errors, i.e.,
$\PR \left( \hat{J}^{(i)}< 1/2 \big|J^{(i)}=1 \right)$ and $\PR \left( \hat{J}^{(i)} \geq 1/2 \big|J^{(i)}=0 \right)$.
Proposition~\ref{prop:global} answers both questions.

\begin{proposition}[Choice of Voting Weights]\label{prop:global}
	Suppose $n \geq \log(2 b_0)/(b_1 h^{d_x+4})$, $h \leq b_3$ and the weights are set to
	\begin{equation*}
	w_j =
	\begin{cases}
	\dfrac{\log 2+ \log p_j}{\sum_{k:p_k \leq 0.5} (\log 2 + \log p_k)} & \text{if} \ p_j \leq 0.5 \\
	0 & \text{if} \ p_j > 0.5
	\end{cases},
	\end{equation*}
	where $p_j$ is defined in Proposition \ref{prop:subset}.
	Then under Assumptions~\ref{ass:sub_gau}, \ref{ass:continuous}, \ref{ass:spa_cov}, \ref{ass:pos_gra},~\ref{ass:smooth_cov}, and~\ref{ass:cov_des2}, we have
	\begin{equation}\label{equ:global-bound}
	\PR \left(\left|\hat{J}^{(i)} - J^{(i)}\right| \geq \frac{1}{2}\right) \leq \exp\left\{\frac{1}{2}\left(h^{-d_x} (1+\log b_0+\log 2)-b_1 n h^{4}\right)\right\}.
	\end{equation}
	Moreover, the union bound implies
	\begin{equation*}
	\PR(\hat{J}=J) \geq 1-d_x\exp\left\{\frac{1}{2}\left(h^{-d_x} (1+\log b_0+\log 2)-b_1 n h^{4}\right)\right\}.
	\end{equation*}
\end{proposition}
{\RV \Copy{cop:dis-global}{Proposition \ref{prop:global} shows the probability guarantee for the global variable selector.
Compared to Proposition~\ref{prop:subset}, the probability bound improves from $\exp(-n_j h^4)$ to $\exp(a h^{-d_x}-b n h^4)$ for some positive constants $a$ and $b$ omitting terms independent of $n$ and $h$.
This is a significant improvement because on average there are $n_j\approx n h^{d_x}$ observations in a bin and we expect $ h^{-d_x}\ll n$ and $n_j\ll n$.
It demonstrates the power of weighted voting as it aggregates all the available data.}
}
\begin{remark}[The Convergence Rate]\label{rmk:rate-BV-LASSO}
	We provide some intuitions for the convergence rate $O(\exp(h^{-d_x}-n h^{4}))$.
	It is well known that the false selection probability of LASSO for linear functions is $O(\exp(-n))$ (Theorem 11.3 in \citealt{hastie2015statistical}).
	Our bound has an additional term $\exp(h^{-d_x})$, because we have to discretize the covariate space into $h^{-d_x}$ bins for the nonparametric setting.
	Also, there is another term $h^{4}$ in the convergence rate, which comes from approximating $f$ by a linear function.
	There are two inferior alternatives to weighted voting:
	(1) If we just focus on a single bin, then roughly $n h^{d_x}$ observations are used.
	So the convergence rate $O(\exp(h^{-d_x}-n h^{4+d_x}))$ is much worse than weighted voting.
	(2) If we assign the same weight to all the bins, then the votes from bins with fewer observations may tilt the outcome disproportionately, leading to noisy estimates.
\end{remark}

\subsection{BV-LASSO and the Regret Analysis}
After binning the observations, applying localized LASSO and weighted voting, the algorithm proceeds to the online learning phase and only focuses on the relevant variables in $\hat J$.
Algorithm~\ref{alg:CB_Non_LASSO} demonstrates the complete algorithm combining the two phases, which we refer to as ``BV-LASSO and Learning''.
\begin{algorithm}
	\caption{BV-LASSO and Learning}
	\label{alg:CB_Non_LASSO}
	\begin{algorithmic}[1]
		\STATE \textbf{Input}:
		$T,d_x,\mu_m,\mu_M,L,\sigma$
		\STATE \textbf{Tunable parameters}:
		$n,h,\lambda$
		\FOR{$t=1,2,\ldots,n$}
		\STATE Observe covariate $\bm{X}_t$
		\STATE Choose a fixed decision $Y_t=y$
		\STATE Observe $Z_t$ \COMMENT{colleting observations in the variable selection phase}
		\ENDFOR
		\STATE Partition the covariate space into $\mathcal B_h$
		\FOR{$j=1,2,\ldots,h^{-d_x}$}
		\STATE $\hat{J}_j=\supp\left\{ \argmin_{\theta_0,\bm{\theta}}\left\{\frac{1}{n_j} \sum_{t=1}^{n_j} \left( Z_t- \theta_0-\bm{U}_t^T \bm{\theta} \right)^2 +2\lambda \|\bm{\theta}\|_{1}\right\} \right\}$ \COMMENT{applying LASSO to bin $B_j$}
		\ENDFOR
		\FOR{$i=1,2,\ldots,d_x$}
		\STATE $\hat{J}^{(i)}=\sum_{j=1}^{h^{-d_x}} w_j \hat{J}_j^{(i)}$
		\COMMENT {$w_j$ defined in Proposition \ref{prop:global}}
		\ENDFOR
		\STATE Let $\hat{J}=\{i: \hat{J}^{(i)} \geq 0.5\}$
		\COMMENT {the set of selected coordinates}
		\FOR{$t=n+1,n+2,\ldots,T$}
		\STATE{Apply contextual bandits algorithm to the variables in $\hat{J}$}
		\ENDFOR
	\end{algorithmic}
\end{algorithm}

The regret analysis of the algorithm follows from Proposition \ref{prop:global}.
Since the false selection probability decreases exponentially with $n h^4$, BV-LASSO easily meets the rate required in Proposition \ref{prop:rate_selection}.
For properly chosen parameters, we have

{\RV

\begin{theorem}[Regret of BV-LASSO]\label{theo:cov_reg}
	Suppose \begin{equation}\label{equ:warm-up-T}
	T \geq \max\left\{\left((3+\log 2+\log b_0)/b_1\right)^{3(1+2/d_x)}, (b_3)^{-3(d_x+2)}, (\log T)^{3(1+2/d_x)}\right\},
	\end{equation} and Assumptions~\ref{ass:sub_gau}, \ref{ass:continuous}, \ref{ass:spa_cov}, \ref{ass:pos_gra}, \ref{ass:smooth_cov} and \ref{ass:cov_des2} hold.
	Taking $n=T^{2/3}$, $h=n^{-1/(2 d_x+4)}=T^{-1/(3d_x+6)}$ and $\lambda=b_2 h^2$, we have
	\begin{equation*}
	R_\pi(T)=
	O\left(T^{1-1/(d_x^*+3)} \log(T) \right).
	\end{equation*}
\end{theorem}
We have some flexibility in the choice of $n$ as long as it satisfies Proposition \ref{prop:rate_selection}.
In Theorem \ref{theo:cov_reg}, we choose $n$ as a polynomial of $T$, where the warm-up periods \eqref{equ:warm-up-T} have a polymonial dependence on the constants $b_0,b_1,b_3$. We can also choose $n=O(\log T)$, which will shorten the variable selection phase. But it comes at the cost of a longer warm-up period that depends on $b_0,b_1,b_3$ at a higher order.
This is shown in Corollary \ref{cor:cov_reg_n}.

\begin{corollary}\label{cor:cov_reg_n}
	Suppose \begin{equation*}
	T \geq \max\left\{\exp\left\{(3+\log 2+\log b_0)/b_1, (b_3)^{-d_x}\right\}\right\},
	\end{equation*} and Assumptions~\ref{ass:sub_gau}, \ref{ass:continuous}, \ref{ass:spa_cov}, \ref{ass:pos_gra}, \ref{ass:smooth_cov} and \ref{ass:cov_des2} hold.
	Taking $n=(\log T)^{2+4/d_x}$, $h=n^{-1/(2 d_x+4)}=(\log T)^{-1/d_x}$ and $\lambda=b_2 h^2$, we have
	\begin{equation*}
	R_\pi(T)=
	O\left(T^{1-1/(d_x^*+3)} \log(T) \right).
	\end{equation*}
\end{corollary}
}
Note that the constants $b_0,b_1,b_2,b_3$, similar to Proposition~\ref{prop:subset}, are given in Section~\ref{sec:dec_spa}.
We do point out that to set the values of $\lambda$ and $w_j$, we need to be able to access some model parameters ($\sigma,\mu_m,\mu_M,L$) and compute those constants.
We discuss this point in Remark~\ref{rmk:known-para}.

We have shown that BV-LASSO doesn't significantly increase the regret relative to the regret incurred in the online learning phase, demonstrated by the optimal rate of regret.
As a general tool, we believe it has potential to be implemented for other nonparametric variable selection problems outside online learning.



\begin{remark}[Smooth Reward Function]\label{rmk:lower-bound}
	Note that the regret achieved in Theorem \ref{theo:cov_reg} matches the optimal rate \eqref{eq:formu_lb} for $f$ that is Lipschitz continuous.
        But Assumption \ref{ass:smooth_cov} (second-order smoothness in $\bm{x}$) is stronger and may lower the optimal rate of regret to $\tilde{O}(T^{(d^*_x+2)/(d^*_x+3)})$.
	{\RV \Copy{cop:hu-discrete}{In the setting of finite-armed contextual bandits, \citet{hu2019smooth} show the minimax regret to be $\Theta(T^{(d^*_x+2)/(d^*_x+4)})$ when the reward functions are second-order smooth in $\bm{x}$.
        Under their setting, we can use BV-LASSO to select the relevant variables before using their algorithm to achieve the optimal regret.
	But in the continuum-armed bandit setting, as far as we know, no online learning algorithm are designed to adapt to the smoothness.
	Since we focus on the variable selection, we omit the technical subtlety in the paper. }}
\end{remark}

\section{Theoretical Analysis}\label{sec:dec_spa}
In this section, we provide the detailed analysis for Theorem~\ref{theo:cov_reg}. 

\subsection{Analysis of Localized LASSO}\label{sec:analysis_local_LASSO}
In this section, we provide the major steps of the proof for Proposition~\ref{prop:subset}.
The proof is related to the variable selection consistency of LASSO \citep{zhao2006model,meinshausen2006high,wainwright2009sharp}.
We use some of the core ideas in proving the theoretical properties of LASSO and
adapt them to the case when $f$ is not necessarily linear.

\textbf{Notations and Characterizations of LASSO.}
We rewrite the observations in bin $B_j$ in the following form:
\begin{equation}\label{eq:lr-form}
Z_t=f(\bm X_t, y)+\epsilon_t = \bar{\bm{U}}_t^T \bm{\theta}^*+ \Delta_t + \epsilon_t\eqqcolon\bar{\bm{U}}_t^T \bm{\theta}^*+ \rho_t,
\end{equation}
where $\bar{\bm U}=(1, \bm U)\in \mathbb R^{d_x+1}$ incorporates the constant term,
$\bm\theta^*$ is the coefficients of the linear projection of $f$ in $B$ scaled by $h$ because of the normalization, i.e., $\bm{\theta}^*=(\theta_0,h \theta_1,\ldots,h \theta_{d_x})^T$ where $(\theta_0,\ldots,\theta_{d_x})$ is the solution to \eqref{equ:L2}, and $\Delta_t\coloneqq f(\bm X_t,y)-\bar{\bm{U}}_t^T \bm{\theta}^*$ is the approximation error.
In other words, we combine the random error $\epsilon_t$ and the approximation error $\Delta_t$ into $\rho_t$ and transform the problem into a linear regression.
It is still not a standard linear regression, as $\rho_t$ is no longer i.i.d. and does not have mean zero.
We hope to control $\Delta_t$ and thus $\rho_t$ in the subsequent analysis because of Lemma~\ref{lem:proj}.

The new form allows us to utilize the techniques developed for linear regression.
More precisely, we define the design matrix $A \coloneqq (1/\sqrt{n_j}) (\bar{\bm U}_1, \dots,\bar{\bm U}_{n_j})^T$ and vectorize the observations $\bm Z \coloneqq (1/\sqrt{n_j})(Z_1,\dots,Z_{n_j})^T$ and the error term $\bm\rho \coloneqq (1/\sqrt{n_j})(\rho_1,\dots,\rho_{n_j})$.
Then \eqref{eq:lr-form} can be written as $\bm Z = A\bm\theta^*+\bm\rho$.
We also introduce the empirical version of the covariance matrix $\Psi$ defined in Assumption~\ref{ass:pos_gra}, which will be useful in our analysis:
\begin{equation*}
\hat{\Psi}
= A^T A
= \frac{1}{n_j} \sum_{i=1}^{n_j} \bar{\bm{U}}_i \bar{\bm{U}}_i^T.
\end{equation*}
We also rearrange the order of the variables so that $J=\{1,\ldots,d^*_x\}$ and $J^c=\{d^*_x+1,\ldots,d_x\}$ and partition the vectors and matrices into ``relevant'' and ``redundant'' blocks:
\begin{equation}
A=\left(A_{(1)} A_{(2)}\right),\ {\bm{\theta}^*}= \begin{pmatrix}
\bm{\theta}^*_{(1)}\\ \bm{\theta}^*_{(2)}
\end{pmatrix},\
\hat{\Psi}=\begin{pmatrix}
\hat{\Psi}_{11} & \hat{\Psi}_{12} \\
\hat{\Psi}_{21} & \hat{\Psi}_{22} \\
\end{pmatrix}
=
\begin{pmatrix}\label{equ:cov-mat-part}
A_{(1)}^T A_{(1)} & A_{(1)}^T A_{(2)} \\
A_{(2)}^T A_{(1)} & A_{(2)}^T A_{(2)} \\
\end{pmatrix}
,
\end{equation}
where the dimensions are clear from the context (e.g., $A_{(1)}\in \mathbb R^{n_j\times(d_x^*+1)}$ because of the constant vector $\bm e$).

It is proved in Lemma~1 of \citet{zhao2006model} that ${\bm\theta}$ solves \eqref{equ:lasso} if and only if it satisfies the following KKT (Karush-Kuhn-Tucker) conditions:
\begin{align}
(A_{.i})^T (Z - A {\bm{\theta}})= \lambda \sign(\theta_i) \quad \text{if} \quad  {\theta}_i \neq 0 \label{eq:zhao-yu-lem1}\\
\left|(A_{.i})^T (Z- A {\bm{\theta}})\right| \leq \lambda \quad \quad \quad \quad \, \, \text{if} \quad {\theta}_i=0\notag
\end{align}
for all $i=1,2,\dots,d_x$.
Here $\sign(\cdot)$ stands for the sign function for each entry of a vector and $A_{.i}$ stands for the $i$-column of $A$.
Thus, our goal is to show that any $\bm\theta$ satisfying the above equations has the same signs as $\bm\theta^*$, which in turn matches the signs of the partial derivatives of $f$ by Lemma~\ref{lem:proj}. The following parts accomplish this goal.

\textbf{``Good'' Events for Sign Consistency.}
Suppose $\hat{\bm\theta}$ is the LASSO estimator for~\eqref{equ:lasso}, or equivalently, a solution to \eqref{eq:zhao-yu-lem1}.
As $\hat{\bm\theta}$ doesn't have a closed form, we then define a set of events $\Omega_i$, $i=1,\dots,4$, and argue that if $\cap_{i=1}^4\Omega_i$ occurs, then $\hat{\bm\theta}$ has the same signs as $\bm\theta^*$.
The first two events are defined as
\begin{align*}
\Omega_{1}&\coloneqq\left\{(1-\alpha) \underline{\lambda} \leq \lambda_{\min}(\hat{\Psi}) \leq  \lambda_{\max}(\hat{\Psi}) \leq (1+\alpha) \overline{\lambda}\right\}\\
\Omega_{2}&\coloneqq\left\{|(\hat{\Psi}_{21})_{ik}| \leq (1+\delta) \gamma \underline{\lambda}/d_x^*,\  \forall i \in J^c, k \in J\right\},
\end{align*}
where $\alpha\coloneqq\frac{1-\gamma}{2(1+\gamma)}$ and $\delta\coloneqq\frac{1-\gamma}{4\gamma}$, and $\overline{\lambda},\underline{\lambda},\gamma$ are defined in condition two of Assumption~\ref{ass:cov_des}, a weaker version of Assumption~\ref{ass:cov_des2} (discussed in Appendix \ref{app:irrepresentable}). 
{\RV Note that $\hat{\Psi}$ is the empirical estimate of the conditional covariance matrix $\Psi=\EX[\bm{U} \bm{U}^T | \bm{X} \in B_j]$, given that $X_1,\ldots,X_{n_j} \in B_j$.}
Compared to Assumption~\ref{ass:cov_des}, it is clear that $\Omega_1$ and $\Omega_2$ characterize the concentration of the empirical covariance matrix $\hat\Psi$ around the mean $\Psi$.
In particular, $\Omega_1$ corresponds to condition one of Assumption~\ref{ass:cov_des} and $\Omega_2$ corresponds to condition two.
Both events have error margins $\alpha$ and $\delta$ to accommodate the random error.

The events $\Omega_3$ and $\Omega_4$ are less straightforward to interpret:
\begin{align*}
\Omega_3&\coloneqq\left\{\Big|(\hat{\Psi}_{11}^{-1} A^T_{(1)} \bm{\rho})_i-\lambda (\hat{\Psi}_{11}^{-1} \sign (\bm{\theta}^*_{(1)}))_i\Big| \leq |(\bm{\theta}^*_{(1)})_i|, \  \forall i \in J\right\}\\
\Omega_4&\coloneqq\left\{\Big|\left(\hat{\Psi}_{21} \hat{\Psi}_{11}^{-1} A^T_{(1)}\bm\rho-A^T_{(2)}\bm\rho\right)_i\Big| \leq \frac{1}{2}(1-\gamma) \lambda, \ \forall i \in J^c \right\}.
\end{align*}
Since LASSO is a shrinkage estimation method, all the estimators $\hat{\bm\theta}$ are biased towards zero.
Roughly speaking, $\Omega_3$ guarantees that the estimators for the coefficients of relevant variables are not shrunk too much,
while $\Omega_4$ guarantees that the estimators for coefficients of redundant variables are shrunk sufficiently.
The degree of the shrinkage is precisely controlled by the penalty term $\lambda$.
After algebraic manipulations, one can show that $\Omega_3\cap\Omega_4$ is equivalent to \eqref{eq:zhao-yu-lem1}. When the joint event $\cap_{i=1}^4\Omega_i$ occurs, we have
\begin{lemma}\label{lem:sign-events}
	On the event $\cap_{i=1}^4\Omega_i$, the LASSO estimator $\hat{\bm{\theta}}$ for~\eqref{equ:lasso} is unique and $\sign({\hat{\bm{\theta}}})=\sign(\bm{\theta}^*)$.
\end{lemma}

Note that the techniques used in the proof are more or less standard in the LASSO literature.
We present the complete proof in Appendix \ref{app:proof-local}.

\textbf{Probability Bound for ``Good'' Events.}
By Lemma \ref{lem:sign-events}, we know the LASSO estimator has the desired property under the ``good'' events. The last step to prove Proposition  \ref{prop:subset} is to show $\cap_{i=1}^4\Omega_i$ occurs with high probability.

\begin{lemma}\label{lem:prob-bound-good}
	Under Assumptions~\ref{ass:sub_gau}, \ref{ass:continuous}, \ref{ass:spa_cov}, \ref{ass:pos_gra}, \ref{ass:smooth_cov}, and \ref{ass:cov_des2}, choosing $h\leq b_3$ and $\lambda=b_2 h^2$, we have
	\begin{equation*}
	\PR (\cap_{i=1}^4\Omega_i) \geq 1-b_0 \exp(b_1 n_j h^4).
	\end{equation*}
\end{lemma}
The constants in Lemma~\ref{lem:prob-bound-good} are the same as Proposition~\ref{prop:subset}, which are presented below
\begin{align*}
&b_0(d_x)=2\max\{2 (d_x+1), d_x^2/4\},\\
&b_1(d_x,\mu_m, \mu_M,L,\sigma)=\frac{11 \mu_m}{10^4 (1+d_x/4)} \wedge \mu_m^2/(4608 d_x^2) \wedge 64 L^2 d_x^2/(2 \sigma^2) \wedge 22400 \mu_M L^2 d_x^3/\sigma^2,\\
&b_2(d_x,\mu_M)=64 \sqrt{7 \mu_M/3} L d_x,\\
&b_3(d_x,\mu_m,\mu_M,L_{\mu},C) =\min\left\{C \mu_m/(768 \sqrt{21 \mu_m d_x}), \mu_m^2/(3 d_x L_{\mu})\right\}.
\end{align*}
Their derivations can be found in the proof, which is provided in Appendix \ref{app:proof-local}.

\begin{remark}\label{rmk:known-para}
	The constants $\mu_m$, $\mu_M$, $L_{\mu}$, $L$, and $\sigma$ appearing in Proposition~\ref{prop:subset} are defined in Assumptions~\ref{ass:sub_gau},~\ref{ass:spa_cov},~\ref{ass:pos_gra},~\ref{ass:smooth_cov},~\ref{ass:cov_des2} and the constant $C$ is defined in \eqref{equ:global-C} of Assumption \ref{ass:pos_gra}.
	To implement the localized LASSO in a bin, the decision maker needs to know $\mu_M$, $d_x$ and $L$ to obtain the penalty $\lambda$.
	To get the misidentification probability $p_j$ for weighted voting, the decision maker in addition needs to know $\mu_m$ and $\sigma$.
	The implementation of Algorithm~\ref{alg:CB_Non_LASSO} does not need the value of $C$ and $L_{\mu}$, which appear in the bound for $h$ that is satisfied automatically if $n$ is large enough.
\end{remark}

The proof of Lemma \ref{lem:prob-bound-good} deviates significantly from the LASSO literature, as the error $\bm\rho$ is not i.i.d. due to the approximation error.
The bound for $\PR(\Omega_1\cap \Omega_2)$ arises from random matrix concentration inequalities:
the empirical covariance matrix $\hat\Psi$ can be viewed as the average of independent copies of $\bar{\bm U}\bar{\bm U}^T$, whose mean is $\Psi$.
Therefore, we can guarantee that the spectrum (eigenvalues) and entries of the matrix do not deviate too much from the mean.
The bound for $\PR(\Omega_3\cap \Omega_4)$ is harder to analyze, as it involves the matrix inverse and multiplications such as $\hat\Psi_{21}\hat\Psi_{11}^{-1}$.
The left-hand sides of the inequalities in $\Omega_3$ and $\Omega_4$ are linear transformations of the error $\bm\rho$, but the coefficients are not tractable.
To analyze $\Omega_3$ and $\Omega_4$, we use the bounds for the eigenvalues conditional on $\Omega_1$.
In particular, we exploit the following inequalities: for a square matrix $A$ and a vector $\bm x$, we have $\|A\|_2\le \lambda_{\max}(A)$ and $\|A\bm x\|_2\le \|A\|_2\|\bm x\|_2$.
They help to reduce matrix multiplications to the eigenvalues, which is explicitly bounded in $\Omega_1$.
Eventually, we can transform $\Omega_3$ and $\Omega_4$ to a bound for a simple linear combination of sub-Gaussian random variables, for which we can apply standard concentration bounds.

\subsection{Analysis of Weighted Voting}\label{sec:analysis-voting}
Now that we have obtained the probability of making mistakes in selecting relevant variables in a single bin from Proposition~\ref{prop:subset}, we proceed to analyze the effect of weighted voting, i.e., Proposition~\ref{prop:global}.
Note that for a certain variable $x_i$, the outcome of a bin $\hat J^{(i)}_j$ can be treated as a Bernoulli random variable with $\PR(\hat{J}^{(i)}_j =0 | J^{(i)}=1)<p_j$ and $\PR(\hat{J}^{(i)}_j =1 | J^{(i)}=0)<p_j$.
{\RV \Copy{cop:why-concentration}{Therefore, $\hat J^{(i)}$ in \eqref{eq:weighted-vote} is a weighted average of $h^{-d_x}$ Bernoulli random variables with different success probabilities. So the optimal $w_j$ doesn't have a closed form.
To analyze the error probabilities $\PR(\hat{J}^{(i)} \geq \xi | J^{(i)}=0)$ or $\PR(\hat{J}^{(i)} <\xi | J^{(i)}=1)$ for some $\xi>0$, we, we use the concentration inequalities to obtain an upper bound for the error and then calculate the optimal weights for the upper bound. }}
In particular, we have that for all $\eta>0$,
\begin{align} \label{equ:lasso_global_prob}
\PR(\hat{J}^{(i)} \geq \xi | J^{(i)}=0)
&=\PR (e^{\eta \hat{J}^{(i)}} \geq e^{\eta \xi} | J^{(i)}=0)\notag\\
&\le \exp(-\eta\xi)\prod_{j=1}^{h^{-d_x}}\EX[\exp(\eta w_j X_j)]\notag\\
&\leq \exp\left\{\sum_{j=1}^{h^{-d_x}} (e^{\eta w_j}-1)p_j- \eta \xi\right\},
\end{align}
where $X_j$ is a Bernoulli random variable with $\PR(X_j=1)<p_j$.
The last inequality follows from the moment generating function of Bernoulli random variables: $\EX[\exp(\eta X_j)]\le 1+p_j (\exp(\eta)-1)\le \exp(p_j(e^\eta-1)).$
The inequality \eqref{equ:lasso_global_prob} holds for all non-negative $\eta$ and $w_j$. {\RV \Copy{cop:weighted-voting-main}{Note that the probability in \eqref{equ:lasso_global_prob} is conditional on the information of the bins which $X_1,\ldots,X_n$ fall in. We omit it here for the ease of exposition, and defer the rigorous analysis to Appendix \ref{app:weighted-voting-main}. }}
Our objective is to find $\eta$ and $w_j$ that minimize the {\RV logarithm} of the error, i.e.,
\begin{equation}\label{equ:global_obj}
\begin{aligned}
\min_{\eta,\bm{w}} \quad & V(\eta,\bm{w})\coloneqq \sum_{j=1}^{h^{-d_x}} (e^{\eta w_j}-1)p_j- \eta \xi \\
\mbox{s.t.}\quad & w_j \geq 0, \ \forall j \in \{1,2,\ldots,h^{-d_x}\}, \\
& \eta \ge 0,\\
& \sum_{j=1}^{h^{-d_x}} w_j=1.\\
\end{aligned}
\end{equation}

{\RV \Copy{cop:KKT-minimal}{
Since $V \rightarrow +\infty$ as $\eta \rightarrow +\infty$ and $V(\eta,\bm w)$ is a continuous function, the global minimum is obtained in a compact set when $\eta$ is finite.
Therefore, the global minimum necessarily satisfies the KKT condition,
although the objective function $V(\eta,\bm{w})$ may not be convex in $(\eta,\bm{w})$. 
Recall that the KKT condition is a necessary condition for all the local minima and maxima. 
In the proof of Lemma \ref{lem:global_min} in Appendix \ref{app:weighted-voting}, we prove that the KKT condition admits a unique solution. Then the unique solution must be a global minimum for problem \eqref{equ:global_obj}. 
}}


\begin{lemma}[Optimal Weights]\label{lem:global_min}
	The optimal solution $\eta^*, \bm{w}^*$ of the optimization problem \eqref{equ:global_obj} satisfies:
	\begin{enumerate}[itemsep= 1 pt,topsep = 4 pt]
		\item $\eta^*=\sum_{j=1}^{h^{-d_x}} \left(\log \xi- \log p_j\right) \mathbb{I}\left(p_j < \xi\right)$;
		\item If $p_j < \xi$, then $w_j^*=(\log \xi- \log p_j)/\eta^*$;
		\item If $p_j \ge \xi$, then $w_j^*=0$;
		\item The optimal value $V(\eta^*,\bm{w}^*)=\sum_{j=1}^{h^{-d_x}}\left(\xi-\xi \log \xi-p_j+\xi \log p_j\right) \mathbb{I}\left(p_j < \xi\right).$
	\end{enumerate}
\end{lemma}

Lemma~\ref{lem:global_min} implies an intuitive structure of the weights.
If bin $B_j$ has a high misidentification error $p_j>\xi$, then the variable selection output by $B_j$ is not counted in the vote ($w_j=0$).
Otherwise, the weight assigned is proportional to $\log (\xi/p_j)$.
Clearly, the weights are biased toward the bins with higher confidence (smaller $p_j$).
Moreover, recall that $p_j=b_0\exp(-b_1 n_j h^4)$. So $\log (\xi/p_j)$ roughly grows in the order of $n_j$.
In other words, the voting power from $B_j$ is almost proportional to the number of observations $n_j$ in each bin. Therefore, each observation contributes equally to the global selector of the covariates.

Lemma~\ref{lem:global_min} provides a weighting mechanism after the covariates have been generated and observed (after calculating $p_j$).
What about the ex ante performance of the mechanism?
Note that $p_j$ depends on $n_j$, the number of observations in a bin.
If the distribution of $\bm X$ were known, then $p_j$ might be estimated.
However, this is usually too strong an assumption in typical learning problems.
Instead, we investigate the worst-case scenario in which $V(\eta,\bm w)$ attains the maximum for all possible values of $p_j$ (or equivalently, $n_j$).
Using the form of $V(\eta^*,\bm w^*)$ from Lemma~\ref{lem:global_min}, we have
\begin{equation}\label{equ:lasso_global_max}
\begin{aligned}
\max_{\bm{n}} \quad & V(\bm{n})\coloneqq\sum_{j=1}^{h^{-d_x}} (\xi-p_j-\xi \log \xi + \xi \log p_j)\mathbb{I} (p_j < \xi)\\
\mbox{s.t.}\quad & p_j = b_0 \exp\left(-b_1 n_j h^{4}\right) \\
& \sum_{j=1}^{h^{-d_x}} n_j=n\\
& n_j \in N^{+}, \quad \forall j \in \{1,2,\ldots,h^{-d_x}\}.
\end{aligned}
\end{equation}
Note that the discontinuity in the objective function introduced by the indicator $\mathbb{I} (p_j < \xi)$ presents a challenge.
To address the issue, we treat it as a budget allocation problem.
Then, after analyzing the optimal budget allocation rule, we reformulate it as a concave optimization problem.
The optimal solution is demonstrated in the following lemma.
\begin{lemma}[Worst-case Covariate Distribution]\label{lem:global_max}
	The optimal solution $\bm{n}^*$ of the optimization problem \eqref{equ:lasso_global_max} satisfies $n_1^*=n_2^*=\ldots=n^*_{h^{-d_x}}=n h^{d_x}$, and the optimal value satisfies
	\begin{equation}\label{equ:global_max}
	V(\bm{n}^*) \leq \xi\left(h^{-d_x} (1+\log b_0-\log \xi)-b_1 n h^{4}\right).
	\end{equation}
\end{lemma}

Lemma~\ref{lem:global_max} shows that the worst case occurs when the covariates are equally distributed across the bins.
Combining Lemma~\ref{lem:global_min} and Lemma~\ref{lem:global_max} and setting $\xi=0.5$, we can prove Proposition~\ref{prop:global}.

\section{BV-LASSO for Local Relevance}\label{sec:extensions}

In this section, we relax Assumption~\ref{ass:pos_gra} to Assumption~\ref{ass:gra-point}.
Recall from the analysis in Section \ref{sec:dec_spa} that Assumption~\ref{ass:pos_gra} plays an important role in the successful variable selection (Proposition~\ref{prop:subset}).
However, the theoretical guarantee only requires that $|\partial f(\bm{x},y) /\partial x_i| \ge  C$ always holds locally in a bin.
If this is the case for a large number of bins under Assumption~\ref{ass:gra-point}, then one may still be able to select the variables by weighing the votes from these bins more, given that there is a mechanism to do so.

To see the intuition, note that Assumption \ref{ass:gra-point} implies that the hypercube $\mathcal H_i$ is contained in the \textit{level set} of variable $i$, defined as
\begin{equation}\label{equ:level-set}
\mathcal H_i\subset A_i (C) \coloneqq \left\{\bm{x}: \left|\frac{\partial f(\bm{x},y)}{\partial x_i}\right| \geq C, \ \forall \  y \in \YY \right\}.
\end{equation}
This implies that as $h\to 0$, there are always at least a constant fraction of the $h^{-d_x}$ bins entirely inside $\mathcal H_i$, or $A_i(C)$.
For those bins, which we refer to as ``informative bins'', $|\partial f(\bm{x},y) /\partial x_i| \ge  C$ holds locally and the probability guarantee in Proposition~\ref{prop:subset} holds for the bin.
On the other hand, for ``uninformative bins'' which are partially or entirely outside $A_i(C)$, Assumption~\ref{ass:pos_gra} fails and we no longer have any theoretical guarantee for the output of localized LASSO.

To formalize the idea, given $h$ and the partition $\mathcal{B}_h$, we define the informative area as the union of informative bins:
\begin{align*}
Q_i(C) &\coloneqq \cup \left\{B_j: B_j \subset A_i(C),\ B_j\in \mathcal{B}_h,\ j= 1,2,\ldots,h^{-d_x}\right\},
\end{align*}
while $Q_i^c(C)$ denotes the complimentary area.
One would expect that when aggregating the outputs of localized LASSO from the bins, the BV-LASSO algorithm should still work if the area of $Q_i(C)$ does not vanish for $i\in J$.
This is indeed the case, as $\mathcal H_i$ itself doesn't scale with $h$.

\begin{proposition}[Informative Area]
	\label{prop:inf_bins}
	Suppose Assumptions~\ref{ass:continuous}, \ref{ass:gra-point}, \ref{ass:smooth_cov} and \ref{ass:cov_des2} hold. Then for the constant $C$, the hypercubes $\{\mathcal{H}_i\}_{i \in J}$ in Assumption \ref{ass:gra-point} and $h \leq C/(3L)$,
	we have
	\begin{equation*}
	\PR \left(\bm{X} \in Q_i(C)\right)\geq (1/3)^{d_x}\PR \left(\bm{X} \in \mathcal{H}_i \right) \geq \mu_{m} \left(\frac{C}{3 L}\right)^{d_x}\eqqcolon p_Q,
	\end{equation*}
	for all $i\in J$. Note that $p_Q \in (0,1]$ is a constant.
\end{proposition}
Proposition~\ref{prop:inf_bins} states that as $h\to 0$, there are always at least a constant fraction of bins for which Proposition~\ref{prop:subset} holds.
Under the stronger Assumption~\ref{ass:pos_gra}, we always have $p_Q=1$.
As we shall see next, as long as $p_Q$ is bounded away from zero, our algorithm can be adjusted to successfully select the relevant variables.

\begin{remark}[Intuition of the Informative Area]
	Although Assumption~\ref{ass:gra-point} is very weak and sufficient for Proposition~\ref{prop:inf_bins},
	one may be concerned that the hypercube $\mathcal H_i$ is small and leads to a small $p_Q$, which may affect the performance of the algorithm (see Proposition~\ref{prop:small-xi} below).
	In practice, the level set $A_i(C)$ and informative area $Q_i(C)$ can be much larger than $\mathcal H_i$ and the value of $p_Q$ in Proposition~\ref{prop:inf_bins} can be too conservative.
	Nevertheless, since our algorithm doesn't need to take $p_Q$ as an input, the actual performance may be much better than the theoretical guarantee.

	\begin{figure}[ht]
		\centering
		\pgfplotsset{
			colormap={selfmild}{
				rgb255(0cm)=(128,128,128);
				rgb255(1cm)=(160,160,0);
				rgb255(3cm)=(255,100,0)},
			tick label style={font=\normalsize},
			major grid style={dotted,color=black},
			tick align=center,
			width=7.3cm}

		\begin{tikzpicture}
		\begin{axis}[
		grid=major,
		axis x line*=left,
		axis y line*=left,
		axis z line*=left,
		view={20}{25},
		xlabel=$x_1$,
		ylabel=$x_2$,
		xtick={0,0.2,0.4,0.6,0.8,1.0},
		ytick={0,0.5,1.0},
		ztick={0,0.5,1.0},
		]
		\addplot3[
		surf,
		fill=white,
		samples=30,
		domain=0:1,
		]
		{exp(-15*(x-0.5)^2-15*(y-0.5)^2)};
		\end{axis}
		\end{tikzpicture}
		\hskip 10pt
		\begin{tikzpicture}
		\begin{axis}[
		grid=major,
		axis x line*=left,
		axis y line*=left,
		axis z line*=left,
		view={20}{25},
		xlabel=$x_1$,
		ylabel=$x_2$,
		xtick={0,0.2,0.4,0.6,0.8,1.0},
		ytick={0,0.5,1.0},
		ztick=\empty,
		zmax=3,
		zmin=-6,
		enlarge z limits,
		]

		\addplot3[
		surf,
		fill=white,
		samples=30,
		domain=0:1,
		]
		{-30*(x-0.5)*exp(-15*(x-0.5)^2-15*(y-0.5)^2)};

		\addplot3 [
		contour gnuplot={
			output point meta=rawz,
			levels={-3,-2,-0.9,0.9,2,3},
			contour label style={
				inner sep=0pt,
				every node/.append style={text=black},
				font=\normalsize,
			}
		},
		z filter/.code={\def\pgfmathresult{-6.7}},
		domain=0:1,
		domain y=0:1,
		very thick,
		] {-30*(x-0.5)*exp(-15*(x-0.5)^2-15*(y-0.5)^2)};
		\end{axis}
		\end{tikzpicture}
		\caption{An illustration of the level set.}
		\label{fig:level-set}
	\end{figure}
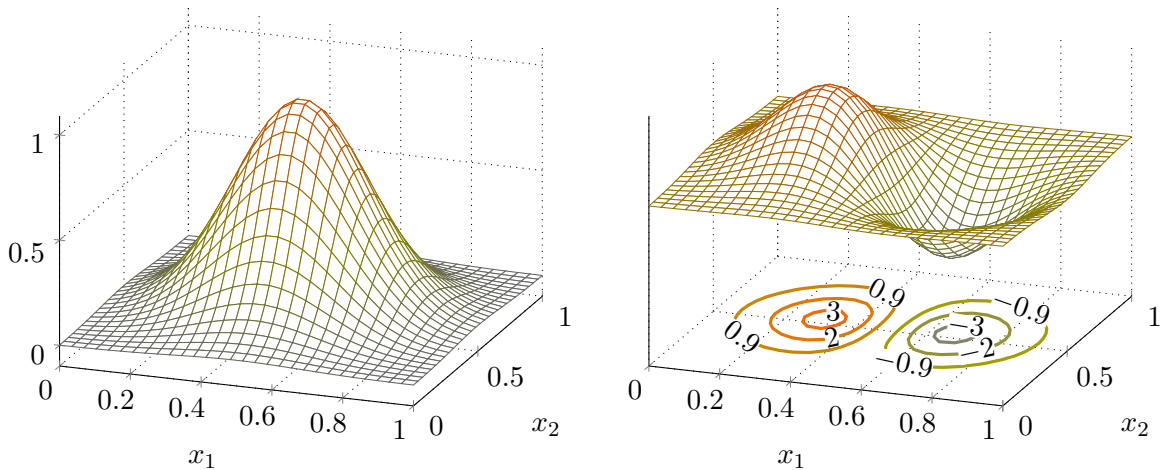
	To give some intuition, consider
	$f(x_1,x_2,y)=\exp\left(-15(x_1-0.5)^2-15(x_2-0.5)^2\right)$, which is illustrated in the left panel of Figure~\ref{fig:level-set}.
	The partial derivative of $x_1$
	and its contour map are illustrated in the right panel of Figure \ref{fig:level-set}.
	If we set $C=0.9$, then the level set $A_1(0.9)$ is the area inside the contour line labelled $0.9$ and $-0.9$.

	Next consider $Q_1(0.9)$ for given $h=0.2$ and $h=0.1$, which is illustrated in Figure~\ref{fig:informative-area}.
	The bins completely inside $A_1(0.9)$ are informative bins (heavily shaded bins) and the bins fully outside $A_1(0.9)$ (lightly shaded bins) are uninformative bins.
	There are some bins (white) intersecting with the boundaries of $A_1(0.9)$, also counted as uninformative.
	As $h\to 0$, $Q_1(0.9)$ approximates $A_1(0.9)$ and $\PR(\bm{X} \in Q_1(0.9))$ converges to $\PR(\bm{X} \in A_1(0.9))$.
	\vspace{20pt}

	\begin{figure}[ht]
		\centering
		\subfigure{
			\begin{minipage}[t]{0.95\textwidth}
				\centering
				\includegraphics[width=10cm]{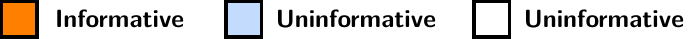}
		\end{minipage}}
		\subfigure{
			\centering
			\begin{minipage}[t]{0.45\textwidth}
				\centering
				\includegraphics[width=6.2cm]{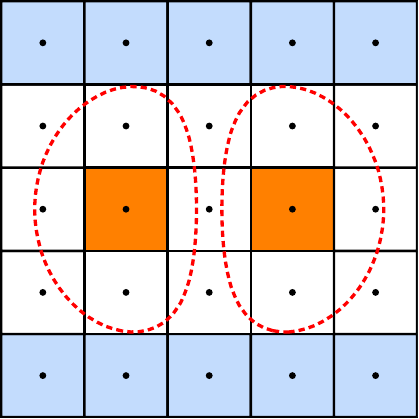}
		\end{minipage}}
		\subfigure{
			\centering
			\begin{minipage}[t]{0.45\textwidth}
				\centering
				\includegraphics[width=6.2cm]{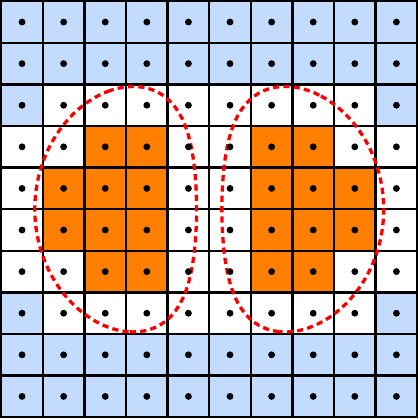}
		\end{minipage}}
		\caption{An illustration of the informative area.}
		\label{fig:informative-area}
	\end{figure}

\end{remark}

Note that Proposition \ref{prop:inf_bins} guarantees that roughly at least $p_Q n$ observations fall into informative bins.
If the decision maker knows which bins are informative a priori, s/he can assign zero weights to uninformative bins and only allows the informative bins to vote, then the problem is simplified to the problem analyzed in Section~\ref{sec:monotone_algorithm} with $p_Q n$ observations.
The challenge, of course, is that the decision maker does not know which bins are informative. If a majority of bins are uninformative and they vote that the variable is redundant, then it is hard for the decision maker to screen out the noisy votes.
In order to bias toward the informative bins in the weighted voting, the key is to tune $\xi$ in Lemma~\ref{lem:global_min} and \ref{lem:global_max}.
To see this, note that the threshold $\xi$ balances the false positive probability $\PR(\hat{J}^{(i)} \geq \xi | J^{(i)}=0)$ and the false negative probability $\PR(\hat{J}^{(i)} \leq \xi | J^{(i)}=1)$.
A smaller $\xi$ leads to a higher false positive rate and a lower false negative rate.
Because uninformative bins tend to vote ``redundant'' (or negative) even if $i\in J$,
we want to set $\xi$ to be smaller to reduce the false negative rate, which is biased toward informative bins.
That is, a small $\xi$ assigns more importance to the bins that vote ``relevant'' and less importance to the bins that vote ``redundant''.
For a relevant variable, if $\xi$ is sufficiently small, then the ``relevant'' votes from the informative bins eventually outweigh the ``redundant'' votes.
For a redundant variable, although the importance of ``redundant'' votes shrinks, there are no bins systematically voting ``relevant'' and the probability is still guaranteed.
The next proposition shows the probability guarantee of the modified voting scheme.
\begin{proposition}[Weights Under Local Relevance]\label{prop:small-xi}
	Suppose that $n \geq \log(2 b_0)/(b_1 h^{d_x+4})$, $h < \min\{C/(3L),b_3/2\}$ and the weights satisfy {\RV \footnote{\Copy{cop:weight-local}{We can use the weights here in Proposition \ref{prop:global}, but the constants in the bound \eqref{equ:local-false-posi} are much looser.}}}
	\begin{equation*}
	w_j =
	\left\{
	\begin{array}{lr}
	\dfrac{\log \xi+\log(1-p_j)- \log p_j-\log(1-\xi)}{\sum_{k:p_k \leq \xi} \log \xi+\log(1-p_k)- \log p_k-\log(1-\xi)}, & \text{if} \ p_j \leq \xi \\
	0, & \text{if} \ p_j > \xi \\
	\end{array}
	\right..
	\end{equation*}
	Then, under Assumptions~\ref{ass:sub_gau}, \ref{ass:continuous}, \ref{ass:spa_cov}, \ref{ass:gra-point}, \ref{ass:smooth_cov} and  \ref{ass:cov_des2},  the misidentification probability of $x_i$ is bounded by
	\begin{align}
	\PR(\hat{J}^{(i)} \geq \xi | J^{(i)}=0) &\leq \exp\left\{\left(2\xi \log b_0\!\!-2\xi \log \xi\!-\!(1\!-\!\xi) \log(1\!-\!\xi)\right) h^{-d_x}\!-\!\xi b_1 h^4 n \right\}, \label{equ:local-false-posi}\\
	\PR(\hat{J}^{(i)} \leq \xi | J^{(i)}=1) &\leq \exp\left\{\left(2(1\!-\!\xi) \log b_0\!-\!\xi \log  \xi\!+\!2(1\!-\!\xi)\log(1\!-\!\xi)\right) h^{-d_x} \!-\!\left(\frac{2p_Q}{3}\!-\!\xi\right)b_1 h^4 n\right\} \notag \\
	{}&\ +\exp\left(-\frac{2}{9} p_Q^2 n\right). \label{equ:local-false-nega}
	\end{align}

\end{proposition}
From \eqref{equ:local-false-posi} and \eqref{equ:local-false-nega}, we can see how $\xi$ balances the false positive and false negative probabilities. The false positive probability $\PR(\hat{J}^{(i)} \geq \xi | J^{(i)}=0)$ converges at the rate $O(\exp(-\xi b_1 h^4 n))$ while the false negative probability $\PR(\hat{J}^{(i)} \leq \xi | J^{(i)}=1)$ converges at the rate $O(\exp(-(2 p_Q/3-\xi)b_1 h^4 n))$.
If the value of $p_Q$ is known, then setting $\xi=p_Q/3$ leads to a bound of $\exp(-n h^4 p_Q/3)$ for both probabilities $\PR(\hat{J}^{(i)} \geq \xi | J^{(i)}=0)$ and $\PR(\hat{J}^{(i)} \leq \xi | J^{(i)}=1)$.
{\RV \Copy{cop:local-xi}{
If $p_Q$ is unknown, then we may set $\xi$ as a function of $n$ (such as $1/n^a$ for a constant $a$).
It guarantees that for a sufficiently large $n$ (or $T$), $\xi$ is less than $2p_Q/3$.}
\begin{corollary}\label{cor:local-relevance}
	Under the conditions of Proposition~\ref{prop:small-xi}, we have
	\begin{equation*}
	\PR(\hat{J}\!=\!J) \!\geq\!
	\left\{
	\begin{array}{lr}
	1\!-\! d_x\exp \left\{2(\log b_0\!-\!\log (\frac{p_Q}{3})) h^{-d_x}\! -\!\frac{1}{3} p_Q b_1 h^4 n\right\}
	\!-\!d_x \exp\left(\!-\frac{2}{9} p_Q^2 n\right), \:   \text{if} \;  \xi\!=\! \frac{p_Q}{3},  & \\
	1\!-\!d_x \exp \left\{2(\log b_0\!+\!2 e^{-1})h^{-d_x}\!-\!b_1 h^4 n^{1-a} \right\}\!-\!d_x \exp\left(\!-\frac{2}{9} p_Q^2 n\right), \: \text{if} \;  \xi\!=\! \frac{1}{n^a}. & \\
	\end{array}
	\right.
	\end{equation*}
\end{corollary}

Corollary~\ref{cor:local-relevance} generalizes the theoretical guarantee of Proposition \ref{prop:global} to local relevance (Assumption~\ref{ass:gra-point}).
The new bound still guarantees the regret in Theorem~\ref{theo:cov_reg}, which shows in the following theorem.


\begin{theorem}\label{theo:local-cov_reg}
	Suppose \begin{equation}\label{equ:local-warm-up-T}
	T \geq \max\left\{\left((1\!+\!6 e^{-1}\!+\!4 \log b_0)/b_1\right)^{4.5+6/d_x}, (b_3)^{-(4.5 d_x+6)}, (\log T)^{4.5+6 d_x}, \left(3/(2 p_Q)\right)^{4.5+6/d_x}\right\},
	\end{equation}
	where $p_Q$ shows in Proposition \ref{prop:inf_bins}.
	Under Assumptions~\ref{ass:sub_gau}, \ref{ass:continuous}, \ref{ass:spa_cov}, \ref{ass:gra-point}, \ref{ass:smooth_cov} and \ref{ass:cov_des2},
	then taking $n=T^{2/3}$, $h=n^{-1/(3 d_x+4)}$, $\xi=0.5 n^{-d_x/(3d_x+4)}$ and $\lambda=b_2 h^2$, we have
	\begin{equation*}
	R_\pi(T)=
	O\left(T^{1-1/(d_x^*+3)} \log(T) \right).
	\end{equation*}
\end{theorem}
We have shown that under the local relevance assumption (Assumption \ref{ass:gra-point}), BV-LASSO still achieves the optimal rate of regret. But the weaker assumption comes at the cost of a longer warm-up period \eqref{equ:local-warm-up-T}. Comparing with \eqref{equ:warm-up-T}, it has a higher order of dependence on $b_0,b_1,b_3$, also an additional term $\left(3/(2 p_Q)\right)^{4.5+6/d_x}$, which is added to make sure $\xi \leq 2 p_Q/3$. Note that the additional term is at most $\left(3/(2 p_Q)\right)^{10.5}$ since $d_x \geq 1$.

}
\section{Numerical Experiments}\label{sec:numerical}
In this section, we conduct numerical experiments to validate the theoretical performances of BV-LASSO.
We attempt to address three questions in practice: (1) Can the BV-LASSO algorithm successfully select relevant variables?
(2) How does the BV-LASSO and Learning algorithm perform against existing algorithms without considering the sparsity structure?
(3) How does BV-LASSO perform when $f$ is a linear function of $\bm{x}$?
We first introduce the setups below.

\textbf{Reward functions.} Supposing $d_x=3$ and $d_x^*=1$, we consider two functions.\footnote{We also consider the setting of larger $d_x$ and more complicated functions, seeing Appendix \ref{app:numerical}.}
The first function is nonlinear:
\begin{equation}\label{equ:numerical-f1}
f_1(\bm{x},y)=\exp\left(-10(x_1-0.5)^2-15(x_1-y)^2\right),
\end{equation}
where $\bm{x}=(x_1,x_2,x_3)$ and only $x_1$ is relevant.
Note that its optimal decision $y^*(\bm{x})=x_1$, and the optimal value $f_1^*(\bm{x})=\exp(-10(x_1-0.5)^2)$. The second function is linear with respect to $x_1$ when $y$ is fixed:
\begin{equation}\label{equ:numerical-f2}
f_2(\bm{x},y)=3(1-2 x_1)y+3x_1.
\end{equation}
When $x_1 <0.5$, its optimal solution $y^*(\bm{x})=1$ and $f^*_2(\bm{x})=3-3x_1$; when $x_1 \geq 0.5$, its optimal solution $y^*(\bm{x})=0$ and $f^*_2(\bm{x})=3x_1$.
At time $t$, the covariate $\bm{X}_t$ is independently sampled from a uniform distribution in $[0,1]^3$. The noise $\epsilon_t$ are generated from a Gaussian distribution $N(0,\sigma^2)$, where we vary the value of $\sigma$ as a robustness check.
{\RV Note that $f_1$ satisfies local relevance (Assumption~\ref{ass:gra-point}) and $f_2$ doesn't satisfy global (Assumption~\ref{ass:pos_gra}) or local relevance.}

\textbf{BV-LASSO inputs.} To implement the algorithm, we need to specify a set of hyper-parameters: $T, d_x, n, h, \lambda, \xi$.
Among them, $T$ and $d_x$ are known to the decision-maker; $\xi$ can be set to $0.5$ as the partial derivatives are non-vanishing in most area; $n$,$h$ are chosen as in Theorem \ref{theo:cov_reg}.
We also set $h=1/\lfloor{n^{1/(2 d_x+4)}\rfloor}$ for the bin size.
To determine the value of $\lambda$, the $l_1$-penalty in localized LASSO, one is required to know $L$ and $\mu_M$ as in Proposition \ref{prop:subset}.
To avoid this scenario, we use a heuristic approach by noting that $\lambda=\Theta(h^2)$ in Proposition \ref{prop:subset}.
We set $\lambda=c_{\lambda} h^2$ for some constant $c_\lambda$.
We vary $c_{\lambda}$ to better understand the sensitivity of the algorithm's performance to the choice.

To choose the weights $w_j$ of the bins, if we follow Proposition \ref{prop:global}, then the knowledge of $\mu_m,\mu_M,L,\sigma$ are required, which is often unknown in practice.
Instead, we simply set $w_j$ to be proportional to $n_j$ (number of observations in bin $B_j$), $w_j=n_j / \sum_{j=1}^{h^{-d_x}} n_j$,
which is still consistent with Propositions~\ref{prop:subset} and~\ref{prop:global} to a large degree.
Our numerical results indicate a good performance.

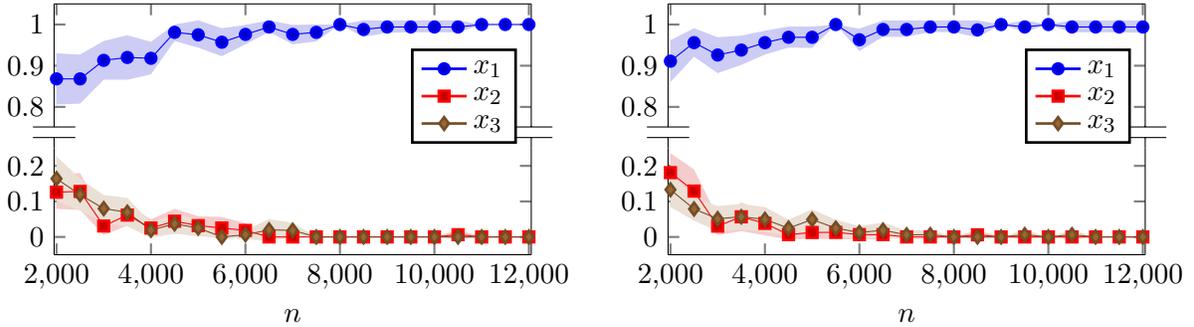
\begin{figure}[htbp]
	\centering
	\setlength{\abovecaptionskip}{-30pt}
	\pgfplotsset{
		every axis/.append style={
			line width=1pt,
			tick style={line width=0.8pt}},
		scaled x ticks=false,
		every non boxed x axis/.style={} }
	\begin{tikzpicture}
	\begin{groupplot}[
	group style={
		group name=my fancy plots,
		group size=1 by 2,
		xticklabels at=edge bottom,
		vertical sep=0pt
	},
	width=0.48\textwidth,
	xmin=1950, xmax=12050,
	cycle list name=mylist
	]

	\nextgroupplot[ymin=0.7,ymax=1.05,
	ytick={0.8,0.9,1.0},
	axis x line=top,
	axis y discontinuity=parallel,
	height=3.5cm,
	legend pos = south east,
	]
	\addplot table[x = n, y = x1m, col sep=comma] {data/Nonlinear_selection.csv};
	\addlegendentry{$x_1$}
	\addplot table[x = n, y = x2m, col sep=comma] {data/Nonlinear_selection.csv};
	\addlegendentry{$x_2$}
	\addplot table[x = n, y = x3m, col sep=comma] {data/Nonlinear_selection.csv};
	\addlegendentry{$x_3$}
	\addplot [name path=upper,draw=none] table[x=n,y=x1l, col sep=comma] {data/Nonlinear_selection.csv};
	\addplot [name path=lower,draw=none] table[x=n,y=x1h, col sep=comma] {data/Nonlinear_selection.csv};
	\addplot[fill=blue!80!black!20] fill between[of=upper and lower];
	\nextgroupplot[ymin=-0.05,ymax=0.25,
	ytick={0.0,0.1,0.2},
	axis x line=bottom,
	xlabel=$n$,
	height=3.0cm,
	]
	\addplot table[x = n, y = x1m, col sep=comma] {data/Nonlinear_selection.csv};
	\addplot table[x = n, y = x2m, col sep=comma] {data/Nonlinear_selection.csv};
	\addplot table[x = n, y = x3m, col sep=comma] {data/Nonlinear_selection.csv};
	\addplot [name path=upper,draw=none] table[x=n,y=x2l, col sep=comma] {data/Nonlinear_selection.csv};
	\addplot [name path=lower,draw=none] table[x=n,y=x2h, col sep=comma] {data/Nonlinear_selection.csv};
	\addplot[fill=red!80!black!20] fill between[of=upper and lower];
	\addplot [name path=upper,draw=none] table[x=n,y=x3l, col sep=comma] {data/Nonlinear_selection.csv};
	\addplot [name path=lower,draw=none] table[x=n,y=x3h, col sep=comma] {data/Nonlinear_selection.csv};
	\addplot[fill=brown!80!black!20] fill between[of=upper and lower];
	\end{groupplot}
	\end{tikzpicture}
	\hspace{5pt}
	\begin{tikzpicture}
	\begin{groupplot}[
	group style={
		group name=my fancy plots,
		group size=1 by 2,
		xticklabels at=edge bottom,
		vertical sep=0pt
	},
	width=0.48\textwidth,
	xmin=1950, xmax=12050,
	cycle list name=mylist
	]

	\nextgroupplot[ymin=0.7,ymax=1.05,
	ytick={0.8,0.9,1.0},
	axis x line=top,
	axis y discontinuity=parallel,
	height=3.5cm,
	legend pos = south east,
	]
	\addplot table[x = n, y = x1m, col sep=comma] {data/Linear_selection.csv};
	\addlegendentry{$x_1$}
	\addplot table[x = n, y = x2m, col sep=comma] {data/Linear_selection.csv};
	\addlegendentry{$x_2$}
	\addplot table[x = n, y = x3m, col sep=comma] {data/Linear_selection.csv};
	\addlegendentry{$x_3$}
	\addplot [name path=upper,draw=none] table[x=n,y=x1l, col sep=comma] {data/Linear_selection.csv};
	\addplot [name path=lower,draw=none] table[x=n,y=x1h, col sep=comma] {data/Linear_selection.csv};
	\addplot[fill=blue!80!black!20] fill between[of=upper and lower];
	\nextgroupplot[ymin=-0.05,ymax=0.25,
	ytick={0.0,0.1,0.2},
	axis x line=bottom,
	xlabel=$n$,
	height=3.0cm,
	]
	\addplot table[x = n, y = x1m, col sep=comma] {data/Linear_selection.csv};
	\addplot table[x = n, y = x2m, col sep=comma] {data/Linear_selection.csv};
	\addplot table[x = n, y = x3m, col sep=comma] {data/Linear_selection.csv};
	\addplot [name path=upper,draw=none] table[x=n,y=x2l, col sep=comma] {data/Linear_selection.csv};
	\addplot [name path=lower,draw=none] table[x=n,y=x2h, col sep=comma] {data/Linear_selection.csv};
	\addplot[fill=red!80!black!20] fill between[of=upper and lower];
	\addplot [name path=upper,draw=none] table[x=n,y=x3l, col sep=comma] {data/Linear_selection.csv};
	\addplot [name path=lower,draw=none] table[x=n,y=x3h, col sep=comma] {data/Linear_selection.csv};
	\addplot[fill=brown!80!black!20] fill between[of=upper and lower];
	\end{groupplot}
	\end{tikzpicture}
	\caption{Variable selection of BV-LASSO for $f_1$ (left) and $f_2$ (right).}
	\label{fig:voting-score}
\end{figure}

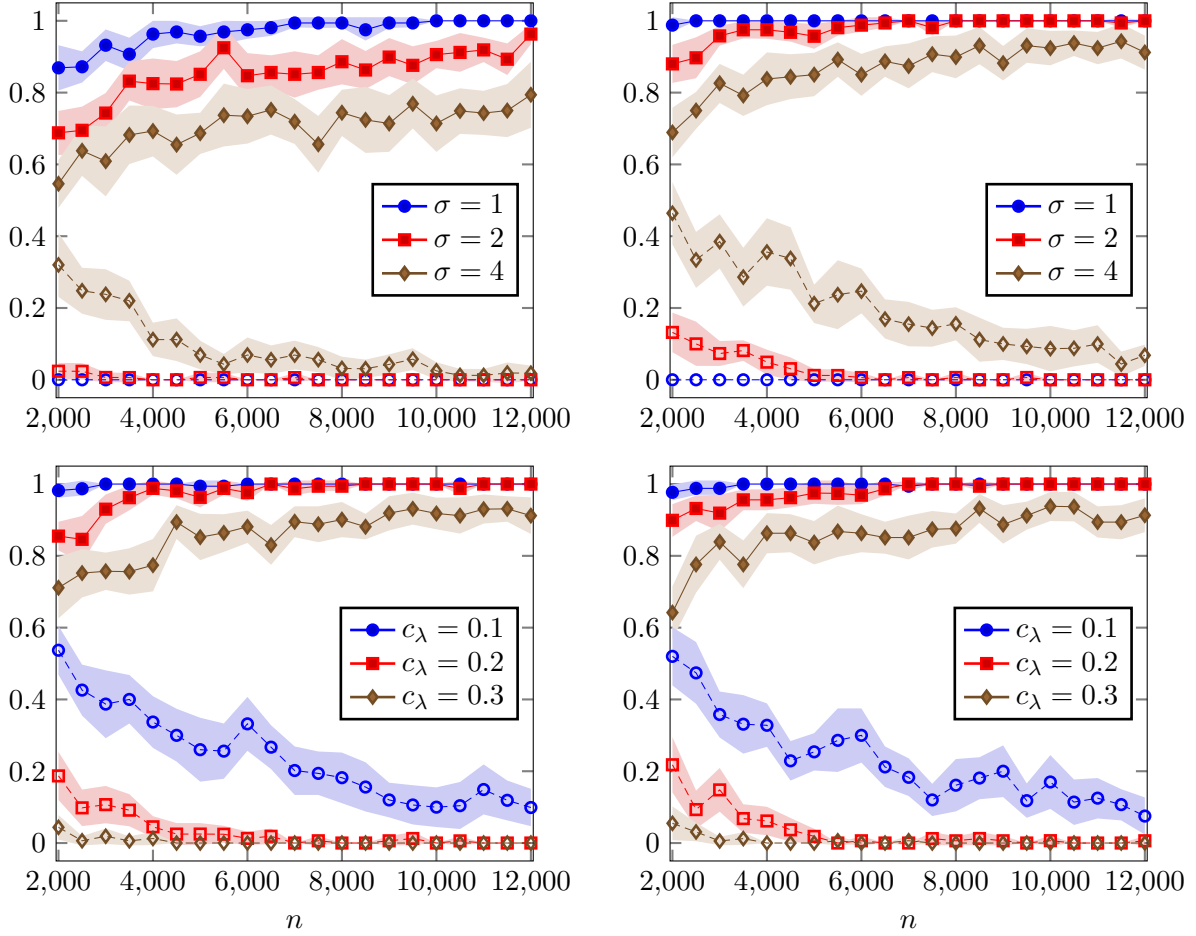
\begin{figure}[htp]
	\centering
	\pgfplotsset{
		every axis/.append style={
			line width=1pt,
			tick style={line width=0.8pt}},
		scaled x ticks=false}
	\begin{tikzpicture}
	\begin{axis}[width=0.48\textwidth,xmin=1950, xmax=12050, ymin = -0.05, ymax = 1.05,
	ytick={0.0,0.2,0.4,0.6,0.8,1.0},
	xlabel=\ ,
	legend style={at={(0.97,0.4)},anchor=east},
	cycle list name=mylist
	]
	\addplot table[x = n, y = 1x1m, col sep=comma] {data/Nonlinear_sigma.csv};
	\addlegendentry{$\sigma=1$}
	\addplot table[x = n, y = 2x1m, col sep=comma] {data/Nonlinear_sigma.csv};
	\addlegendentry{$\sigma=2$}
	\addplot table[x = n, y = 3x1m, col sep=comma] {data/Nonlinear_sigma.csv};
	\addlegendentry{$\sigma=4$}
	\addplot [name path=upper,draw=none] table[x=n,y=1x1l, col sep=comma] {data/Nonlinear_sigma.csv};
	\addplot [name path=lower,draw=none] table[x=n,y=1x1h, col sep=comma] {data/Nonlinear_sigma.csv};
	\addplot[fill=blue!80!black!20] fill between[of=upper and lower];
	\addplot [name path=upper,draw=none] table[x=n,y=2x1l, col sep=comma] {data/Nonlinear_sigma.csv};
	\addplot [name path=lower,draw=none] table[x=n,y=2x1h, col sep=comma] {data/Nonlinear_sigma.csv};
	\addplot[fill=red!80!black!20] fill between[of=upper and lower];
	\addplot [name path=upper,draw=none] table[x=n,y=3x1l, col sep=comma] {data/Nonlinear_sigma.csv};
	\addplot [name path=lower,draw=none] table[x=n,y=3x1h, col sep=comma] {data/Nonlinear_sigma.csv};
	\addplot[fill=brown!80!black!20] fill between[of=upper and lower];
	\pgfplotsset{cycle list shift=-3}
	\addplot table[x = n, y = 1x2m, col sep=comma] {data/Nonlinear_sigma.csv};
	\addplot table[x = n, y = 2x2m, col sep=comma] {data/Nonlinear_sigma.csv};
	\addplot table[x = n, y = 3x2m, col sep=comma] {data/Nonlinear_sigma.csv};
	\addplot [name path=upper,draw=none] table[x=n,y=1x2l, col sep=comma] {data/Nonlinear_sigma.csv};
	\addplot [name path=lower,draw=none] table[x=n,y=1x2h, col sep=comma] {data/Nonlinear_sigma.csv};
	\addplot[fill=blue!80!black!20] fill between[of=upper and lower];
	\addplot [name path=upper,draw=none] table[x=n,y=2x2l, col sep=comma] {data/Nonlinear_sigma.csv};
	\addplot [name path=lower,draw=none] table[x=n,y=2x2h, col sep=comma] {data/Nonlinear_sigma.csv};
	\addplot[fill=red!80!black!20] fill between[of=upper and lower];
	\addplot [name path=upper,draw=none] table[x=n,y=3x2l, col sep=comma] {data/Nonlinear_sigma.csv};
	\addplot [name path=lower,draw=none] table[x=n,y=3x2h, col sep=comma] {data/Nonlinear_sigma.csv};
	\addplot[fill=brown!80!black!20] fill between[of=upper and lower];
	\end{axis}
	\end{tikzpicture}
	\hspace{5pt}
	\begin{tikzpicture}
	\begin{axis}[width=0.48\textwidth,xmin=1950, xmax=12050, ymin = -0.05, ymax = 1.05,
	ytick={0.0,0.2,0.4,0.6,0.8,1.0},
	xlabel=\ ,
	legend style={at={(0.97,0.4)},anchor=east},
	cycle list name=mylist,
	]
	\addplot table[x = n, y = 1x1m, col sep=comma] {data/Linear_sigma.csv};
	\addlegendentry{$\sigma=1$}
	\addplot table[x = n, y = 2x1m, col sep=comma] {data/Linear_sigma.csv};
	\addlegendentry{$\sigma=2$}
	\addplot table[x = n, y = 3x1m, col sep=comma] {data/Linear_sigma.csv};
	\addlegendentry{$\sigma=4$}
	\addplot [name path=upper,draw=none] table[x=n,y=1x1l, col sep=comma] {data/Linear_sigma.csv};
	\addplot [name path=lower,draw=none] table[x=n,y=1x1h, col sep=comma] {data/Linear_sigma.csv};
	\addplot[fill=blue!80!black!20] fill between[of=upper and lower];
	\addplot [name path=upper,draw=none] table[x=n,y=2x1l, col sep=comma] {data/Linear_sigma.csv};
	\addplot [name path=lower,draw=none] table[x=n,y=2x1h, col sep=comma] {data/Linear_sigma.csv};
	\addplot[fill=red!80!black!20] fill between[of=upper and lower];
	\addplot [name path=upper,draw=none] table[x=n,y=3x1l, col sep=comma] {data/Linear_sigma.csv};
	\addplot [name path=lower,draw=none] table[x=n,y=3x1h, col sep=comma] {data/Linear_sigma.csv};
	\addplot[fill=brown!80!black!20] fill between[of=upper and lower];
	\pgfplotsset{cycle list shift=-3}
	\addplot table[x = n, y = 1x2m, col sep=comma] {data/Linear_sigma.csv};
	\addplot table[x = n, y = 2x2m, col sep=comma] {data/Linear_sigma.csv};
	\addplot table[x = n, y = 3x2m, col sep=comma] {data/Linear_sigma.csv};
	\addplot [name path=upper,draw=none] table[x=n,y=1x2l, col sep=comma] {data/Linear_sigma.csv};
	\addplot [name path=lower,draw=none] table[x=n,y=1x2h, col sep=comma] {data/Linear_sigma.csv};
	\addplot[fill=blue!80!black!20] fill between[of=upper and lower];
	\addplot [name path=upper,draw=none] table[x=n,y=2x2l, col sep=comma] {data/Linear_sigma.csv};
	\addplot [name path=lower,draw=none] table[x=n,y=2x2h, col sep=comma] {data/Linear_sigma.csv};
	\addplot[fill=red!80!black!20] fill between[of=upper and lower];
	\addplot [name path=upper,draw=none] table[x=n,y=3x2l, col sep=comma] {data/Linear_sigma.csv};
	\addplot [name path=lower,draw=none] table[x=n,y=3x2h, col sep=comma] {data/Linear_sigma.csv};
	\addplot[fill=brown!80!black!20] fill between[of=upper and lower];
	\end{axis}
	\end{tikzpicture}
	\begin{tikzpicture}
	\begin{axis}[width=0.48\textwidth,xlabel=$n$,xmin=1950, xmax=12050, ymin = -0.05, ymax = 1.05,
	ytick={0.0,0.2,0.4,0.6,0.8,1.0},
	legend style={at={(0.97,0.5)},anchor=east},
	cycle list name=mylist
	]
	\addplot table[x = n, y = 1x1m, col sep=comma] {data/Nonlinear_lambda.csv};
	\addlegendentry{$c_\lambda=0.1$}
	\addplot table[x = n, y = 2x1m, col sep=comma] {data/Nonlinear_lambda.csv};
	\addlegendentry{$c_\lambda=0.2$}
	\addplot table[x = n, y = 3x1m, col sep=comma] {data/Nonlinear_lambda.csv};
	\addlegendentry{$c_\lambda=0.3$}
	\addplot [name path=upper,draw=none] table[x=n,y=1x1l, col sep=comma] {data/Nonlinear_lambda.csv};
	\addplot [name path=lower,draw=none] table[x=n,y=1x1h, col sep=comma] {data/Nonlinear_lambda.csv};
	\addplot[fill=blue!80!black!20] fill between[of=upper and lower];
	\addplot [name path=upper,draw=none] table[x=n,y=2x1l, col sep=comma] {data/Nonlinear_lambda.csv};
	\addplot [name path=lower,draw=none] table[x=n,y=2x1h, col sep=comma] {data/Nonlinear_lambda.csv};
	\addplot[fill=red!80!black!20] fill between[of=upper and lower];
	\addplot [name path=upper,draw=none] table[x=n,y=3x1l, col sep=comma] {data/Nonlinear_lambda.csv};
	\addplot [name path=lower,draw=none] table[x=n,y=3x1h, col sep=comma] {data/Nonlinear_lambda.csv};
	\addplot[fill=brown!80!black!20] fill between[of=upper and lower];
	\pgfplotsset{cycle list shift=-3}
	\addplot table[x = n, y = 1x2m, col sep=comma] {data/Nonlinear_lambda.csv};
	\addplot table[x = n, y = 2x2m, col sep=comma] {data/Nonlinear_lambda.csv};
	\addplot table[x = n, y = 3x2m, col sep=comma] {data/Nonlinear_lambda.csv};
	\addplot [name path=upper,draw=none] table[x=n,y=1x2l, col sep=comma] {data/Nonlinear_lambda.csv};
	\addplot [name path=lower,draw=none] table[x=n,y=1x2h, col sep=comma] {data/Nonlinear_lambda.csv};
	\addplot[fill=blue!80!black!20] fill between[of=upper and lower];
	\addplot [name path=upper,draw=none] table[x=n,y=2x2l, col sep=comma] {data/Nonlinear_lambda.csv};
	\addplot [name path=lower,draw=none] table[x=n,y=2x2h, col sep=comma] {data/Nonlinear_lambda.csv};
	\addplot[fill=red!80!black!20] fill between[of=upper and lower];
	\addplot [name path=upper,draw=none] table[x=n,y=3x2l, col sep=comma] {data/Nonlinear_lambda.csv};
	\addplot [name path=lower,draw=none] table[x=n,y=3x2h, col sep=comma] {data/Nonlinear_lambda.csv};
	\addplot[fill=brown!80!black!20] fill between[of=upper and lower];
	\end{axis}
	\end{tikzpicture}
	\hspace{5pt}
	\begin{tikzpicture}
	\begin{axis}[width=0.48\textwidth,xlabel=$n$,xmin=1950, xmax=12050, ymin = -0.05, ymax = 1.05,
	ytick={0.0,0.2,0.4,0.6,0.8,1.0},
	legend style={at={(0.97,0.5)},anchor=east},
	cycle list name=mylist
	]
	\addplot table[x = n, y = 1x1m, col sep=comma] {data/Linear_lambda.csv};
	\addlegendentry{$c_\lambda=0.1$}
	\addplot table[x = n, y = 2x1m, col sep=comma] {data/Linear_lambda.csv};
	\addlegendentry{$c_\lambda=0.2$}
	\addplot table[x = n, y = 3x1m, col sep=comma] {data/Linear_lambda.csv};
	\addlegendentry{$c_\lambda=0.3$}
	\addplot [name path=upper,draw=none] table[x=n,y=1x1l, col sep=comma] {data/Linear_lambda.csv};
	\addplot [name path=lower,draw=none] table[x=n,y=1x1h, col sep=comma] {data/Linear_lambda.csv};
	\addplot[fill=blue!80!black!20] fill between[of=upper and lower];
	\addplot [name path=upper,draw=none] table[x=n,y=2x1l, col sep=comma] {data/Linear_lambda.csv};
	\addplot [name path=lower,draw=none] table[x=n,y=2x1h, col sep=comma] {data/Linear_lambda.csv};
	\addplot[fill=red!80!black!20] fill between[of=upper and lower];
	\addplot [name path=upper,draw=none] table[x=n,y=3x1l, col sep=comma] {data/Linear_lambda.csv};
	\addplot [name path=lower,draw=none] table[x=n,y=3x1h, col sep=comma] {data/Linear_lambda.csv};
	\addplot[fill=brown!80!black!20] fill between[of=upper and lower];
	\pgfplotsset{cycle list shift=-3}
	\addplot table[x = n, y = 1x2m, col sep=comma] {data/Linear_lambda.csv};
	\addplot table[x = n, y = 2x2m, col sep=comma] {data/Linear_lambda.csv};
	\addplot table[x = n, y = 3x2m, col sep=comma] {data/Linear_lambda.csv};
	\addplot [name path=upper,draw=none] table[x=n,y=1x2l, col sep=comma] {data/Linear_lambda.csv};
	\addplot [name path=lower,draw=none] table[x=n,y=1x2h, col sep=comma] {data/Linear_lambda.csv};
	\addplot[fill=blue!80!black!20] fill between[of=upper and lower];
	\addplot [name path=upper,draw=none] table[x=n,y=2x2l, col sep=comma] {data/Linear_lambda.csv};
	\addplot [name path=lower,draw=none] table[x=n,y=2x2h, col sep=comma] {data/Linear_lambda.csv};
	\addplot[fill=red!80!black!20] fill between[of=upper and lower];
	\addplot [name path=upper,draw=none] table[x=n,y=3x2l, col sep=comma] {data/Linear_lambda.csv};
	\addplot [name path=lower,draw=none] table[x=n,y=3x2h, col sep=comma] {data/Linear_lambda.csv};
	\addplot[fill=brown!80!black!20] fill between[of=upper and lower];
	\end{axis}
	\end{tikzpicture}
	\caption{Variable selection of BV-LASSO  for varying $\sigma$ and $c_\lambda$. The left and right panels demonstrate the result for $f_1$ \eqref{equ:numerical-f1} and $f_2$ \eqref{equ:numerical-f2}, respectively.}
	\label{fig:voting-robust}
\end{figure}

\textbf{Variable selection.}
First, we test the performance of BV-LASSO in terms of variable selection.
The performance of BV-LASSO is affected by $n$, $\sigma$ and $\lambda$.
As $n$ increases, the space is partitioned more granularly and there are more observations in each bin.
Thus, we expect the performance to improve.
The sub-Gaussian parameter $\sigma$ reflects the signal-to-noise ratio.
The penalty $\lambda$ controls the balance between false positive and false negative.
We show the results for varying $n$ in Figure~\ref{fig:voting-score} while fixing $\sigma=2$, $c_\lambda=0.22$ and show the results for varying $\sigma \in\{1,2,4\}$ and $c_\lambda \in \{0.1,0.2,0.3\}$ in Figure~\ref{fig:voting-robust}.


Figure \ref{fig:voting-score} compares the value of $\hat{J}$ of the three variables according to \eqref{eq:weighted-vote} based on the average of $20$ trials, in which only $x_1$ is relevant. The left (right) panel corresponds to $f_1(x,y)$ ($f_2(x,y)$) and the shaded region corresponds to the $95\%$ confidence interval of $20$ trials.
The results show $\hat{J}^{(1)}$ is significantly greater than $0.75$ and $\hat{J}^{(2)},\hat{J}^{(3)}$ are significantly less than $0.25$.
Choosing the threshold as $\xi=0.5$, the relevant variable can be successfully selected, even if $n$ is not large.
The numerical example demonstrates that the BV-LASSO algorithm can successfully select relevant variables.

Figure~\ref{fig:voting-robust} further shows the value of $\hat{J}$ for varying $\sigma$ and $c_\lambda$.
As $\hat{J}^{(3)}$ performs similar to $\hat{J}^{(2)}$, we omit $\hat{J}^{(3)}$ and display $\hat{J}^{(1)}$ ($\hat{J}^{(2)}$) in Figure~\ref{fig:voting-robust}.
The indicators $\hat{J}^{(1)}$ ($\hat{J}^{(2)}$) for variable $x_1$ ($x_2$) are displayed in solid (dashed) curves with filled (hollow) markers.
The top row of Figure~\ref{fig:voting-robust} shows that $\hat{J}^{(1)}$ and $\hat{J}^{(2)}$ are not sensitive to $\sigma$ as long as it is in a reasonable range.
The bottom row of Figure~\ref{fig:voting-robust} shows the impact of $c_\lambda$.

\textbf{The Regret.}
Next we compare the regret of our Algorithm~\ref{alg:CB_Non_LASSO} with other learning algorithms.
Our first benchmark is the Uniform algorithm \citep{kleinberg2005nearly,lu2009showing}, which does not learn the sparse structure of the reward function.
It incurs regret $O(T^{(d_x+2)/(d_x+3)} \log(T))$ or $\tilde O(T^{5/6})$ for functions~\eqref{equ:numerical-f1} and \eqref{equ:numerical-f2}.
Our second benchmark is to first apply the standard LASSO algorithm to select the variables, and then use the Uniform algorithm on the selected variables.
It is expected to incur regret $\tilde{O}(T^{3/4})$ for linear function \eqref{equ:numerical-f2} and linear regret for nonlinear function \eqref{equ:numerical-f1} due to model misspecification.


\begin{figure}
	\centering
	\pgfplotsset{
		every axis/.append style={
			line width=1pt,
			tick style={line width=0.8pt}},
	}
	\begin{tikzpicture}
	\begin{loglogaxis}[width=0.48\textwidth,xlabel=$T$,
	xmin = 10000, xmax = 1000000, ymin = 1000, ymax = 700000,
	cycle list name=mylist,
	legend columns=3, legend style={at={(0.5,1)},anchor=north}]
	\addplot table[x = T, y = UAm, col sep=comma] {data/Nonlinear_regret_log.csv};
	\addlegendentry{UA}
	\addplot table[x = T, y = BVm, col sep=comma] {data/Nonlinear_regret_log.csv};
	\addlegendentry{BV}
	\addplot table[x = T, y = STm, col sep=comma] {data/Nonlinear_regret_log.csv};
	\addlegendentry{ST}
	\addplot[dashed] table[x = T, y = 34, col sep=comma] {data/Nonlinear_regret_log.csv};
	\addlegendentry{$T^{3/4}$}
	\addplot[dashdotted] table[x = T, y = 56, col sep=comma] {data/Nonlinear_regret_log.csv};
	\addlegendentry{$T^{5/6}$}
	\addplot[dotted] table[x = T, y = 11, col sep=comma] {data/Nonlinear_regret_log.csv};
	\addlegendentry{$T$}
	\addplot [name path=upper,draw=none] table[x=T,y=UAl, col sep=comma] {data/Nonlinear_regret_log.csv};
	\addplot [name path=lower,draw=none] table[x=T,y=UAh, col sep=comma] {data/Nonlinear_regret_log.csv};
	\addplot[fill=blue!80!black!20] fill between[of=upper and lower];
	\addplot [name path=upper,draw=none] table[x=T,y=BVl, col sep=comma] {data/Nonlinear_regret_log.csv};
	\addplot [name path=lower,draw=none] table[x=T,y=BVh, col sep=comma] {data/Nonlinear_regret_log.csv};
	\addplot[fill=red!80!black!20] fill between[of=upper and lower];
	\addplot [name path=upper,draw=none] table[x=T,y=STl, col sep=comma] {data/Nonlinear_regret_log.csv};
	\addplot [name path=lower,draw=none] table[x=T,y=STh, col sep=comma] {data/Nonlinear_regret_log.csv};
	\addplot[fill=brown!80!black!20] fill between[of=upper and lower];
	\end{loglogaxis}
	\end{tikzpicture}
	\hspace{5pt}
	\begin{tikzpicture}
	\begin{loglogaxis}[width=0.48\textwidth,xlabel=$T$,
	xmin = 10000, xmax = 1000000, ymin = 1000, ymax = 700000,
	cycle list name=mylist,
	legend columns=3, legend style={at={(0.5,1)},anchor=north}]
	\addplot table[x = T, y = UAm, col sep=comma] {data/Linear_regret_log.csv};
	\addlegendentry{UA}
	\addplot table[x = T, y = BVm, col sep=comma] {data/Linear_regret_log.csv};
	\addlegendentry{BV}
	\addplot table[x = T, y = STm, col sep=comma] {data/Linear_regret_log.csv};
	\addlegendentry{ST}
	\addplot[dashed] table[x = T, y = 34, col sep=comma] {data/Linear_regret_log.csv};
	\addlegendentry{$T^{3/4}$}
	\addplot[dashdotted] table[x = T, y = 56, col sep=comma] {data/Linear_regret_log.csv};
	\addlegendentry{$T^{5/6}$}
	\addplot[dotted] table[x = T, y = 11, col sep=comma] {data/Linear_regret_log.csv};
	\addlegendentry{$T$}
	\addplot [name path=upper,draw=none] table[x=T,y=UAl, col sep=comma] {data/Linear_regret_log.csv};
	\addplot [name path=lower,draw=none] table[x=T,y=UAh, col sep=comma] {data/Linear_regret_log.csv};
	\addplot[fill=blue!80!black!20] fill between[of=upper and lower];
	\addplot [name path=upper,draw=none] table[x=T,y=BVl, col sep=comma] {data/Linear_regret_log.csv};
	\addplot [name path=lower,draw=none] table[x=T,y=BVh, col sep=comma] {data/Linear_regret_log.csv};
	\addplot[fill=red!80!black!20] fill between[of=upper and lower];
	\addplot [name path=upper,draw=none] table[x=T,y=STl, col sep=comma] {data/Linear_regret_log.csv};
	\addplot [name path=lower,draw=none] table[x=T,y=STh, col sep=comma] {data/Linear_regret_log.csv};
	\addplot[fill=brown!80!black!20] fill between[of=upper and lower];
	\end{loglogaxis}
	\end{tikzpicture}
	\caption{Comparison of regret. The abbreviation UA/BV/ST represents Uniform Algorithm/BV-LASSO/Standard LASSO.}
	\label{fig:regret}
\end{figure}
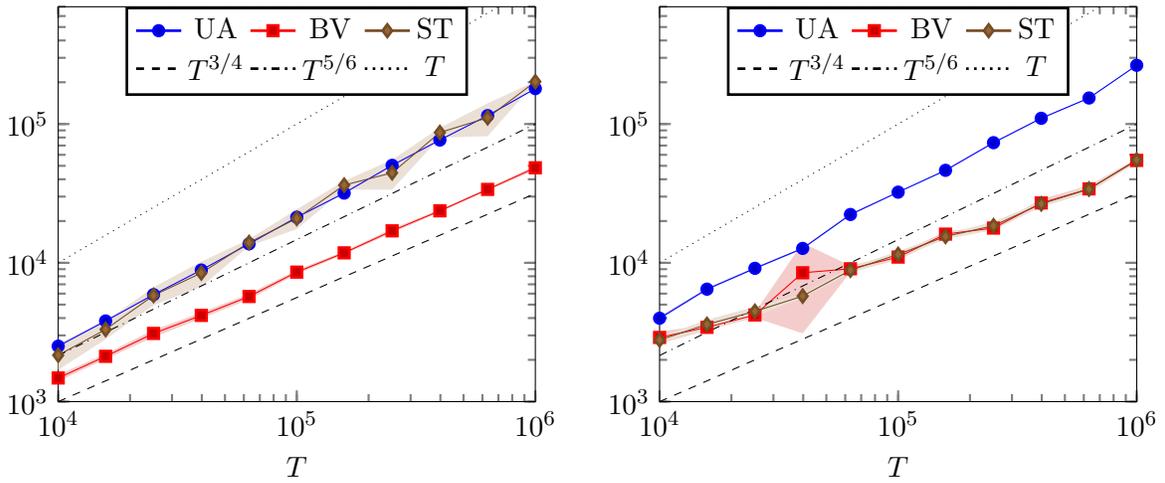

Figure \ref{fig:regret} shows the regret of the three algorithms for a range of $T$ values. We note that each point on the curve displays the average regret of $10$ independent trials and the shaded region around each curve is the $95\%$ confidence interval.
The left (right) panel corresponds to $f_1(x,y)$ ($f_2(x,y)$).
In both panels, BV-LASSO and Learning outperforms the other benchmarks.
As predicted by the theory, the regret of the Uniform algorithm always grows at rate $\tilde{O}(T^{5/6})$ while the regret of BV-LASSO grows at $\tilde{O}(T^{3/4})$.
When the function is nonlinear, the standard LASSO may fail to identify the relevant variable $x_1$ and incur large regret.
When $f$ is linear, the regret of BV-LASSO and Learning almost coincides with that of the standard LASSO, which has been proved to be one of the most effective variable selection methods for the linear setting.

\section{{\RV Extensions and Discussions}}\label{sec:discussion}

\subsection{\RV Extension to Randomized Policies and A Nested Algorithm}\label{sec:rule-stagewise}
So far, we propose the BV-LASSO and Learning algorithm to address the dimensionality issue in the nonparametric contextual online learning problem. However, the algorithm has two potential issues which may limit its practical applicability.
First, the decisions in the variable selection phase are required to be fixed.
Second, the total periods $T$ needs to be known in advance.
We address the first problem by extending the fixed decision to a randomized policy and address the second problem by a nested algorithm that integrates variable selection and online learning.

\textbf{Extension to Randomized Policies.}
We generalize the fixed decision in \eqref{equ:fix-decision} to a randomized policy, where the decision $y$ is sampled from a probability distribution $P:\mathcal{Y} \rightarrow [0,1]$.
Hence the decision maker has some flexibility in making decisions in the variable selection phase.
The observed reward is generated by
\begin{equation}\label{equ:Z-random-decision}
Z_t=f(\bm{X}_t, Y)+\epsilon_t, \  Y \sim P.
\end{equation}
We define the following mean reward function
\begin{equation}\label{equ:random-decision}
f_P(\bm{x})=\EX_{Y \sim P}[f(\bm{x},Y)].
\end{equation}
If $f_P$ maintains the good properties of $f$, then we can directly apply BV-LASSO to $f_P$ and the performance guarantees in Section \ref{sec:monotone_algorithm} still hold.

\begin{proposition}[Randomized Policy in the Variable Selection Phase]\label{prop:decision-rule}
	Applying BV-LASSO to $f_P$ defined in \eqref{equ:random-decision}, where the distribution $P$ is independent of the distribution generating the covariate $\bm{x}$, and under Assumptions~\ref{ass:sub_gau}, \ref{ass:continuous}, \ref{ass:spa_cov}, \ref{ass:pos_gra}, \ref{ass:smooth_cov} and \ref{ass:cov_des2}, we have Proposition \ref{prop:subset}, Proposition \ref{prop:global} and Theorem \ref{theo:cov_reg}.
\end{proposition}

\textbf{Extension to a Nested Algorithm.}
We propose a nested algorithm that divides the $T$ periods into stages with increasing lengths.
In each stage, we implement an independent BV-LASSO and Learning algorithm.
We make two adjustments to improve the performance of the nested algorithm.
First, at every stage $m$, we choose the side length of the partition to be $h_m=1/m$.
That is, the partition in the variable selection phase is refined with the number of stages.
In the first few stages, $h_m$ is relatively large and a bin can easily collect a large number of observations.
Although the approximation error may be large (Lemma~\ref{lem:proj}),
some significantly relevant variables can already be detected.
This adjustment allows the algorithm to perform reasonably well when $T$ is small and an overly refined partition for variable selection cannot be afforded.
Second, when selecting variables at each stage, we repeatedly use the observations in the variable selection phases of previous stages.
This adjustment improves the data utilization and reduces the exploration cost.
We are able to show that the dependence introduced by data reuse doesn't impact the regret.
Algorithm~\ref{alg:Stagewise} demonstrates the nested algorithm, which we refer to as ``Nested BV-LASSO and Learning''.
Note that this algorithm doesn't require the knowledge of $T$ in advance.
\begin{algorithm}
	\caption{Nested BV-LASSO and Learning}
	\label{alg:Stagewise}
	\begin{algorithmic}[1]
		\STATE \textbf{Input parameters}: $T, d_x, \mu_m, \mu_M, L, \sigma, l_1,\ldots, l_m, n_1,\ldots, n_m$.
		\FOR{$m=1,2,\ldots,$}
		\FOR{$t=1,2,\ldots,l_m$}
		\STATE Observe covariate $\bm X_t$
		\STATE Apply a fixed decision $y$ or randomized policy $P$ in \eqref{equ:Z-random-decision}
		\STATE Observe $Z_t^{(m)}$ \COMMENT{collecting observations in stage $m$, $\bm Z^{(m)}=\{Z_1^{(m)},\ldots,Z_{l_m}^{(m)}\}$}
		\ENDFOR
		\STATE Partition the covariate space into  $\mathcal{B}_{h_m}  (h_m=1/m)$
		\STATE Apply BV-LASSO to the previous observations $\bm Z^{(1)}, \ldots, \bm Z^{(m)}$ and obtain $\hat{J}^{(m)}$.
		\FOR{$t=1,2,\ldots,n_m$}
		\STATE Apply the contextual bandit algorithm for the variables in $\hat{J}^{(m)}$.
		\ENDFOR
		\ENDFOR
	\end{algorithmic}
\end{algorithm}

The next theorem shows that Algorithm \ref{alg:Stagewise} achieves the same order of regret as its base version, i.e.,  Algorithm~\ref{alg:CB_Non_LASSO}.
\begin{theorem}\label{theo:stagewise-LASSO}
    Suppose $T \geq \left(b_4 2^{1/b_3}\right)^{3/2}$, $(\log T)^{d_x+4} \leq T^{2/3}$ and Assumptions~\ref{ass:sub_gau}, \ref{ass:continuous}, \ref{ass:spa_cov}, \ref{ass:pos_gra}, \ref{ass:smooth_cov} and \ref{ass:cov_des2} hold.
    Taking $h_m=1/m$, $l_m=(d_x+4) b_4 m^{d_x+3}$, $n_m=b_4 2^m$ and $\lambda_m=b_2 h_m^2$ in Algorithm \ref{alg:Stagewise}, we have
\begin{equation*}
R(T)=O\left(T^{1-1/(d_x^*+3)} \log(T) \right),
\end{equation*}
where $b_4=(1+2 \log 2 + \log b_0 +2 \log d_x)/b_1$.
\end{theorem}

Note that the nested algorithm is more efficient and practical than a simple doubling trick of Algorithm \ref{alg:CB_Non_LASSO}. 
We choose $h_m=1/m$ in each stage such that the bins don't diminish rapidly and the data can be used more efficiently.
Moreover, the choice of $h_m=1/m$ removes the need of rounding to obtain an integer number of bins.
On the contrary, the doubling trick may encounter redundant stages when $\frac{1}{2} < h <1$, $\frac{1}{3} < h <\frac{1}{2}$, etc.
In Algorithm \ref{alg:Stagewise}, we address this issue by decoupling the dependence of $h$ on $T$ and making sure $h$ strictly decreasing as the stage proceeds.

The regret analysis of Algorithm \ref{alg:Stagewise} is much more involved than Algorithm \ref{alg:CB_Non_LASSO}.
Because we not only have to balance the length of the variable selection phase and probability of correct selection, but also have to balance the regret incurred across different stages.
In the proof, we show that with carefully designed hyperparameters $l_m, n_m$, the regret incurred in each stage is optimal, so is the overall regret.

\subsection{\RV Alternative to Localized LASSO}\label{sec:OLS-threshold}
\Copy{cop:ols1}{In this subsection, we formally show Remark \ref{rmk:local-alternative} by introducing the
ordinary least square (OLS) method for the algorithm.
The OLS estimator in bin $B_j$ is}
\begin{equation}\label{equ:standard_OLS}
(\theta_0,\bm{\theta}^{OLS})=\argmin_{\theta_0,\bm\theta}\left\{\frac{1}{n_j} \sum_{t=1}^{n_j} \left( Z_t- \theta_0-\bm{U}_t^T\bm{\theta} \right)^2 \right\},
\end{equation}
\Copy{cop:ols2}{where $\bm U_t$ is defined in \eqref{eq:normalize}.
Note that OLS has a property similar to the LASSO: if the linear approximation error is small enough, then the estimators $\|\bm{\theta}_i^{OLS}\|$ of the redundant variables $x_i$ will be less than the estimators of relevant variables with high probability. So we choose a threshold $r$, and classify the variable $x_i$ as ``relevant'' if $\big|\bm{\theta}_i^{OLS}\big| \geq r$, otherwise, we classify it as ``redundant''.
Proposition~\ref{prop:subset-OLS} shows the theoretical guarantee for the OLS estimator in a single bin.}

\begin{proposition}[Variable Selection of OLS]\label{prop:subset-OLS}
	 Suppose $h\leq b_3$ and Assumptions~\ref{ass:sub_gau}, \ref{ass:continuous}, \ref{ass:spa_cov}, \ref{ass:pos_gra}, \ref{ass:smooth_cov}, and part one of Assumption \ref{ass:cov_des2} hold.
         For a given bin $B_j$ of side length $h$, choosing the threshold $r=b_2 h^2$ and selecting $\hat{J}_j=\{i: \big|\bm{\theta}_i^{OLS}\big| \geq r\}$ for $\bm{\theta}^{OLS}$ in \eqref{equ:standard_OLS}, we have
	\begin{equation}\label{equ:prop_subset_OLS}
	\PR \left(\hat{J}_j=J\right) \geq 1- p_j,
	\end{equation}
	where $p_j \coloneqq b_0 \exp(-b_1 n_j h^{4})$ and
	\begin{align*}
	b_0=2(d_x+1), \, b_1=\min\left\{\frac{1}{\sigma^2},\frac{\mu_m}{6(4+d_x)}\right\}, \,
	b_2=\frac{4 \sqrt{3} (4 \sqrt{2} L d_x +1)}{\sqrt{\mu_m}}, \, b_3=\dfrac{\sqrt{\mu_m} C}{8 \sqrt{3} (4 \sqrt{2} L d_x+1)}.
	\end{align*}
\end{proposition}

The OLS estimator has two theoretical advantages over LASSO.
First, part two of Assumption \ref{ass:cov_des2}, which is imposed to make sure the variable selection consistency of LASSO, is not required for the OLS estimator.
Second, if the number of relevant variables, namely $d_x^*$, is known in advance, there's no need to choose the threshold $r$. Instead, we can just rank the coefficients $\big|\bm{\theta}_i^{OLS}\big|$ and select the largest $d_x^*$ variables as $\hat{J}_j$. The intuition shows in Corollary
\ref{cor:sparse-known}.

\begin{corollary}[Known $d_x^*$]\label{cor:sparse-known}
	If $d_x^*$ is known, we rank $\big|\bm{\theta}_i^{OLS}\big|$ and seelct the largest $d_x^*$ variables as $\hat{J}_j$. Under Assumptions~\ref{ass:sub_gau}, \ref{ass:continuous}, \ref{ass:spa_cov}, \ref{ass:pos_gra}, \ref{ass:smooth_cov}, part one of Assumption~\ref{ass:cov_des2}, and $h\leq b_3$,
	we have
	\begin{equation}\label{equ:prop_subset_OLS-h}
	\PR \left(\hat{J}_j=J\right) \geq 1- p_j,
	\end{equation}
	where $p_j \coloneqq b_0 \exp(-b_1 n_j h^{2})$ and
	\begin{align*}
	b_0=2(d_x+1), \, b_1=\min\left\{\frac{\mu_m C^2}{768 \sigma^2},\frac{\mu_m}{6(4+d_x)} \right\}, \,
	b_3=\dfrac{C \sqrt{\mu_m}}{64 \sqrt{6} L d_x}.
	\end{align*}
\end{corollary}
Note that the $p_j$ in \eqref{equ:prop_subset_OLS-h} has a milder dependence on $h$ ($\exp(-h^2)$) than that in Proposition \ref{prop:subset-OLS} ($\exp(-h^4)$).
Recall that the linear coefficients in \eqref{eq:lr-form} are scaled by $h$: $\bm{\theta}_i^*=O(Ch)$ for $i \in J$ and $\bm{\theta}_i^*=0$ for $i \in J^c$. When $d_x^*$ is known, the method in Corollay \ref{cor:sparse-known} can be viewed as choosing the threshold $r=O(Ch)$ in Proposition \ref{prop:subset-OLS}, which will give $p_j=O(\exp(-h^2))$ . But in the setting of Proposition \ref{prop:subset-OLS}, we don't know the value of $C$. So we choose the threshold $r=O(h^2)$ to make sure it shrinks faster than $\bm{\theta}_i^*$ for $i \in J$.

Despite the theoretical property, OLS has a numerical issue: when $h$ is small, some bins may include a very small number of observations, even less than $d_x$.
For numerical stability, we use LASSO in the numerical experiments.

\begin{remark}[Ridge Regression]
	Ridge regression can also be used for variable selection in a local area.
        It replaces the $\ell_1$-norm term in \eqref{equ:standard_lasso} by its $\ell_2$-norm. But as ridge regression does not create sparsity in estimates, directly applying it will not give the selected variables.
        A threshold has to be chosen to separate the variables as in OLS.
        Thus, to apply it, the decision maker has to carefully choose two hyper-parameters (the threshold and the $\ell_2$ penalty). For this reason, we do not recommend using ridge regression.
\end{remark}


\subsection{\RV Extension to Linear Bandits}\label{sec:linear-bandits}
In this section, we apply BV-LASSO to a parametric setting where the reward is a linear function of the covariate.
This is referred to as ``linear bandits'' in the literature.
We demonstrate how to combine the variable selection with algorithms designed for linear bandits such as \citet{goldenshluger2013linear,bastani2020online} to attain near-optimal regret.

Following the work of sparse linear bandits \citep{bastani2020online, wang2018minimax}, we consider discrete arms (decisions) $k \in [K]$ instead of continuous decision.
The mean reward function has a linear form:\footnote{For simplicity, we don't consider the intercept, which could easily be incorporated using a constant term in $\bm{x}$.}
\begin{equation}\label{equ:linear-bandit-model}
f(\bm{x},k)= \bm{x}^T \bm{\theta}_{k},
\end{equation}
where $\bm{\theta}_{k}$ is the unknown parameter associated with arm $k$. After assuming $f$ is linear in $\bm{x}$, the Assumption \ref{ass:continuous}, \ref{ass:pos_gra}, \ref{ass:gra-point}, \ref{ass:smooth_cov} are automatically satisfied.
Assumption \ref{ass:spa_cov} becomes:
\begin{assumption}[Sparse Parameter] \label{ass:spa_par}
	There exists $d_x^*\le d_x$ such that $\|\bm{\theta}_{k}\|_0 \le d_x^*$ for all $k \in [K]$.
\end{assumption}
Note that the sparsity structure of $\bm{\theta}_{k}$ may depend on the arm $k$. Assumption \ref{ass:spa_par} assumes a uniform upper bound for the number of relevant variables for all $k$.
The next assumption is imposed on the distribution of covariates, which is a milder version of Assumption \ref{ass:cov_des}.

\begin{assumption}[Regular Covariates for Linear Bandits] \label{ass:cov_des_lin}
	The covariance matrix $\Psi\coloneqq\EX [\bm X \bm X^T]$ satisfies
	\begin{enumerate}
		\item there exist constants $\underline{\lambda},\overline{\lambda} >0$, such that
		\begin{equation*}
		\underline{\lambda} \leq \lambda_{\min} \left( \Psi\right) \leq \lambda_{\max} \left( \Psi \right) \leq \overline{\lambda},
		\end{equation*}
		where $\lambda_{\max}$ and $\lambda_{\min}$ represent the maximum and minimum eigenvalue of a matrix.
		\item For any $i \in J$ and $j \in J^c$, there exists a constant $\gamma \in [0,1)$, such that
		\begin{equation*}
		(\Psi)_{ij} \leq \gamma \underline{\lambda}/ d^*_x.
		\end{equation*}
	\end{enumerate}
\end{assumption}

The first condition of Assumption \ref{ass:cov_des_lin} prevents singular covariate distributions. Similar conditions are imposed in linear bandits \citep{goldenshluger2013linear,bastani2020online}. The second condition of Assumption~\ref{ass:cov_des_lin} states that the pairwise correlation between relevant and redundant variables cannot be too high. See more discussion of the assumption in Appendix \ref{app:irrepresentable}.

\textbf{Algorithm.}
The algorithm for the linear setting also adopts a two-phase approach. In the variable selection phase, we try each arm for $n$ periods, then use BV-LASSO to select relevant variables.
Note that there's no need to partition the covariate space into bins since function $f$ is linear.
In this case, BV-LASSO degenerates to the standard LASSO for variable selection.
In the online learning phase, we apply the existing algorithm \citep{goldenshluger2013linear,bastani2020online} for linear bandits on the selected variables. Algorithm \ref{alg:Linear_LASSO} demonstrates the complete algorithm combing the two phases, which we refer to as ``Linear BV-LASSO and Learning''.

\begin{algorithm}
	\caption{Linear BV-LASSO and Learning}
	\label{alg:Linear_LASSO}
	\begin{algorithmic}[1]
		\STATE \textbf{Input}:
		$T,d_x,\overline{\lambda},\underline{\lambda},\gamma,C',\sigma$
		\STATE \textbf{Tunable parameters}:
		$n,\lambda$
		\FOR{$k=1,2,\ldots,K$}
		\FOR{$t=(k-1)n+1,(k-1)n+2,\ldots, (k-1)n+n$}
		\STATE Observe covariate $\bm{X}_t$
		\STATE Pull arm $k$
		\STATE Observe $Z_t$ \COMMENT{colleting observations for arm $k$}
		\ENDFOR
		\STATE $\hat{J}_k=\supp\left\{ \argmin_{\theta_0,\bm{\theta}}\left\{\frac{1}{n} \sum_{t=(k-1)n+1}^{(k-1)n+n} \left( Z_t- \theta_0-\bm{X}_t^T \bm{\theta} \right)^2 +2\lambda \|\bm{\theta}\|_{1}\right\} \right\}$ \COMMENT{applying LASSO to $\bm{\theta}_k$}
		\ENDFOR

		\FOR{$t=nK+1,nK+2,\ldots,T$}
		\STATE{Apply linear bandits algorithm to the variables in $\hat{J}_1,\hat{J}_2,\ldots,\hat{J}_K$}
		\ENDFOR
	\end{algorithmic}
\end{algorithm}

\textbf{Regret.}
The regret analysis is similar to Theorem~\ref{theo:cov_reg}. The variable selection phase incurs regret $O(Kn)$. After that, the sparsity of $\{\bm{\theta}_k\}_{k=1}^K$ can be correctly identified with a high probability.
On the correctly identified $\{\bm{\theta}_k\}_{k=1}^K$, the OLS bandit algorithm in \citet{goldenshluger2013linear,bastani2020online} can achieve regret $O((d_x^*)^3 \log T)$ under the so-called margin condition and the arm optimality condition.
For the exposition, we defer the definition of the two conditions in Appendix~\ref{proof:linear_sparse}. For properly chosen $n$ and $\lambda$, we have the following result for the regret bound.
\begin{theorem}[Regret for Linear Bandits]\label{theo:linear_sparse}
            Suppose Assumption \ref{ass:sub_gau}, \ref{ass:spa_par}, \ref{ass:cov_des_lin}, the margin condition and the arm optimality condition (Assumptions~\ref{ass:margin-condition} and \ref{ass:arm-opt} in Appendix \ref{proof:linear_sparse}) hold.
	If $T \geq (K \log T)/b_1'$, then taking $n=(\log T)/b_1'$ and $\lambda=\frac{C'(1+3\gamma) \underline{\lambda}}{4(1+\gamma)\sqrt{d_x}}$, we have
	\begin{equation*}
	R_{\pi}(T)=O\left(K\left(d_x+(d_x^*)^3\right) \log T+ K d_x^2\right),
	\end{equation*}
	where the constants $b_1', C'$ are presented in Appendix \ref{proof:linear_sparse}.
\end{theorem}

Comparing with the regret bound $O(K d_x^3 \log T)$ of linear bandits without sparsity \citep{goldenshluger2013linear}, Algorithm \ref{alg:Linear_LASSO} improves it by reducing the growth rate on $d_x$. However, the regret bound in Theorem \ref{theo:linear_sparse} is worse than the bound $O\left(K (d_x^*)^2 (\log T +\log d_x)^2\right)$ or $O(K (d_x^*)^2 (d_x^*+\log d_x) \log T)$ showed in sparse linear bandits \citep{bastani2020online,wang2018minimax}.
That's because we only use LASSO to select variables instead of fitting the reward function directly as in \citet{bastani2020online,wang2018minimax}, which turns out to be less efficient.
However, a benefit of our approach is that $d_x^*$ doesn't have to be known.
In summary, although designed for the nonparametric setting, BV-LASSO can be applied to the linear setting and achieves a regret bound that is slightly worse than the specifically designed algorithms.

\begin{remark}[Another Setting of Linear Bandits]
We note that there's another popular setting for linear bandits, where the $K$ arms are represented by $K$ different contexts $\bm{x}^{(1)},\bm{x}^{(2)},\ldots,\bm{x}^{(K)}$, and the reward functions share a single parameter $\bm \theta$ \citep{auer2002using,chu2011contextual,kim2019doubly,ren2020dynamic}.
Algorithm~\ref{alg:Linear_LASSO} can be adjusted for this setting.
In the variable selection phase, only pulling one arm is enough to select relevant variables, as the parameter $\bm{\theta}$ is shared by all the arms. In the online learning phase, we apply LinUCB \citep{chu2011contextual} to the selected variables.
We can prove that the adjusted algorithm achieves regret $O\left(\sqrt{d_x^* T}+d_x \log T\right)$, which improves the regret bound $O(\sqrt{d_x T})$ in the non-sparse setting \citep{auer2002using,chu2011contextual}. Again, our algorithm may not outperform the approaches designed for the sparse setting, i.e., $O(d_x^* \log d_x \sqrt{T})$ \citep{kim2019doubly} or $O(\sqrt{d_x^* T})$ \citep{ren2020dynamic}.
\end{remark}

\begin{remark}[Relation to Batch Bandits]\label{rmk:batch-learning}
The Algorithm \ref{alg:Linear_LASSO} adopts LASSO as a subroutine to select variables. Within the subroutine, an arm is pulled for enough periods to collect data.
It's comparable to the batch-learning setting, where the decision rule in a batch is fixed.
Our subroutine can be directly applied to the existing batch-learning algorithm \citep{ren2020dynamic}, where the first batch contains more periods than required. After the first batch, the relevant variables can be successfully selected with a high probability.
The algorithm can then be run on the selected variables in the remaining batches.
\end{remark}

\section{Conclusions}\label{sec:conclusion}
In this paper, we study the online learning problem with a high-dimensional covariate.
In particular, we address the curse of dimensionality under sparsity as the regret in existing algorithms scales exponentially in the covariate dimension $d_x$.
To our knowledge, it is the first study that proposes a nonparametric variable selection algorithm which takes advantage of the sparsity structure of the covariate.
For online learning problems, our algorithm achieves the same order of regret as if the sparsity structure is known in advance.
The BV-LASSO algorithm and its two recipes, localized LASSO and weighted voting, may be of independent interest in other nonparametric settings.

We conclude by discussing a few limitations of our algorithm and potential future directions.
First, we assume the sparsity structure remains identical for all decisions.
One may consider a setting where different decisions lead to different sparsity structures.
{\RV \Copy{Cop:conclusion-subspace}{Second, the sparsity structure in this paper is assumed to be on the variables.
It would be an interesting direction to extend the algorithm to the setting where the reward function can be represented by a low-dimensional linear subspace.}}
Finally, our algorithm requires the knowledge of several model parameters that are typically unknown and would ideally be optimized for specific applications.
An interesting research question is to develop data-driven methods to select these parameters.

\bibliography{nonpar_sparse_JMLR_r1} 

\newpage
\appendix

\setcounter{equation}{0}
\renewcommand{\theequation}{\thesection.\arabic{equation}}
\setcounter{figure}{0}
\renewcommand{\thefigure}{\thesection.\arabic{figure}}

\newtheorem{thm}{Theorem}[section]
\renewcommand{\thethm}{\thesection.\arabic{thm}}
\newtheorem{prop}{Proposition}[section]
\renewcommand{\theprop}{\thesection.\arabic{prop}}
\newtheorem{assum}{Assumption}[section]
\renewcommand{\theassum}{\thesection.\arabic{assum}}

\section{Proofs and Discussions in Section \ref{sec:problem}}
\subsection{Discussion about Assumption \ref{ass:cov_des2}}\label{app:irrepresentable}
In this section, we discuss a weaker version of Assumption \ref{ass:cov_des2} and use the weaker version in the proofs of Lemma \ref{lem:sign-events}, Lemma \ref{lem:prob-bound-good} and Proposition \ref{prop:subset}.
Recall that $\bm U$ is $\bm X$ being normalized with regard to a bin, introduced in \eqref{eq:normalize}.
\begin{lastassumption}{ass:cov_des2}[Regular Covariates] \label{ass:cov_des}
	Given any hybercube $B \subset \XX$ with side length $h$ such that $\PR (\bm{X} \in B) >0$ and the normalization \eqref{eq:normalize}, we assume that the distribution of $\bm{X}$ satisfies:
	\begin{enumerate}
		\item The conditional covariance matrix $\Psi\coloneqq\mathbb{E} [\bm U \bm U^T |\bm X \in B] $ satisfies
		\begin{equation*}
		0 < \underline{\lambda} \leq \lambda_{\min} \left( \Psi\right) \leq \lambda_{\max} \left( \Psi \right) \leq \overline{\lambda}
		\end{equation*}
		for some constants $\underline{\lambda}$ and $\overline{\lambda}$,
		where $\lambda_{\max}$ and $\lambda_{\min}$ represent the maximum and minimum eigenvalues of a matrix.
		\item For any $i \in J$ and $j \in J^c$, there exists a constant $\gamma \in [0,1)$, which may depend on $h$, such that
		\begin{equation*}
		(\Psi)_{ij} \leq \gamma \underline{\lambda}/ d^*_x.
		\end{equation*}
	\end{enumerate}
\end{lastassumption}

We first give a simple example to show the generality of Assumption~\ref{ass:cov_des}.
If $\bm X$ follows an independent uniform distribution in $\XX$, then $\Psi=\frac{1}{12} \bm{I}_{d_x}$.\footnote{This is the main reason we consider $\bm U$ instead of $\bm X$. The conditional covariance matrix of the normalized covariate is invariant when $h$ changes.}
It is easy to see that setting $\underline{\lambda}=\overline{\lambda}=\frac{1}{12}$ and $\gamma=0$ satisfies the assumption.

The first condition of Assumption~\ref{ass:cov_des} prevents singular covariate distributions.
If the covariates are linearly dependent ($\underline\lambda=0$), then the definition of relevant/redundant variables is ambiguous, as one covariate can be represented by others.
In other words, we need sufficient variations in all the dimensions of $\bm X$ in order to estimate the partial derivatives.
Similar conditions are imposed in the LASSO literature \citep{buhlmann2011statistics,goldenshluger2013linear,bastani2020online}.

The second condition of Assumption~\ref{ass:cov_des} states that the pairwise correlation between relevant and redundant variables cannot be too high. It's a sufficient condition for the well-known ``Strong Irrepresentable Condition'' for LASSO proposed in \citep{zhao2006model}.
It prevents any redundant variable to be fully linearly represented by the relevant variables.
Note that the condition two is hard to check in practice since $d_x^*$ is unknown.
One alternative is to replace $d^*_x$ by $d_x$ and make the assumption stronger: $(\Psi)_{ij} \leq \gamma \underline{\lambda}/ d_x$.

Next, we show that if the side length $h$ is small enough, Assumption \ref{ass:cov_des2} implies Assumption \ref{ass:cov_des}.

\begin{prop}\label{prop:cov_des2}
	If Assumption \ref{ass:cov_des2} is satisfied and the side length $h< \mu_m^2/(3 d_x^* L_{\mu})$,
	then Assumption~\ref{ass:cov_des} holds with $\underline{\lambda}=\mu_m/12$, $\overline{\lambda}=\mu_M$ and $\gamma=3 d_x^* L_{\mu} h/(2 \mu_m^2)$.
\end{prop}

Proposition~\ref{prop:cov_des2} states that the first condition of Assumption~\ref{ass:cov_des} can be implies by Assumption \ref{ass:cov_des2} and the second condition holds automatically when $h$ is sufficiently small, as the requirement of $\gamma$ diminishes linearly in $h$.

\begin{proof}
	Let $\bm{x}_0$ be a vector in $\mathbb{R}^{d_x}$.
	To show the first condition of Assumption~\ref{ass:cov_des} holds, by the definition of eigenvalues, we only need to provide upper and lower bounds for
	$\bm x_0^T \Psi \bm x_0/\|\bm x_0\|_2$.
	Note that
	\begin{align}\label{equ:norm_ineq}
		\int_{\bm{x} \in B} \bm{x}_0^T \bm{U(\bm{x})} \bm{U(\bm{x})}^T \bm{x}_0   \mu_m \dif \bm{x}
		& \leq \int_{\bm{x} \in B} \bm{x}_0^T \bm{U(\bm{x})} \bm{U(\bm{x})}^T \bm{x}_0  \mu({\bm{x}})\dif \bm{x} \notag\\ & \leq \int_{\bm{x} \in B} \bm{x}_0^T \bm{U(\bm{x})} \bm{U(\bm{x})}^T \bm{x}_0  \mu_M \dif \bm{x}.
	\end{align}
	Since $\bm U = (\bm X-C_{B})/h$, we have
	\begin{equation}\label{equ:norm_ineq2}
	\int_{\bm{x} \in B}  \bm{U(\bm{x})} \bm{U(\bm{x})}^T   \dif \bm{x}=
	\begin{pmatrix}
	1 & \bm{0}\\
	\bm{0} & \frac{1}{12} \bm{I}_{d_x}
	\end{pmatrix}.
	\end{equation}
	Then, plugging equation \eqref{equ:norm_ineq2} into \eqref{equ:norm_ineq}, we have
	\begin{equation*}
	\frac{\mu_{m}}{12} \|\bm{x}_0\|_2^2 \leq \bm{x}_0^T \Psi \bm{x}_0=\int_{\bm{x} \in B} \bm{x}_0^T \bm{U(\bm{x})} \bm{U(\bm{x})}^T \bm{x}_0  \mu({\bm{x}})\dif \bm{x} \leq \mu_{M} \|\bm{x}_0\|_2^2.
	\end{equation*}
	Thus, the first condition of Assumption~\ref{ass:cov_des} is satisfied by setting $\underline{\lambda}=\frac{\mu_{m}}{12}, \overline{\lambda}=\mu_{M}$.
	To prove the second condition, note that we have
	\begin{equation*}
	\mu_m h^{d_x} = \mu_m \int_{\bm{x} \in B} \dif \bm{x} \leq \PR(X \in B)=\int_{\bm{x} \in B} \mu(\bm{x}) \dif \bm{x} \leq \mu_M h^{d_x}.
	\end{equation*}
	Then, for any $i \in J$ and $j \in J^c$,
	\begin{align*}
	(\Psi)_{ij}
	&= \mathbb{E} [U_i(\bm{X}) U_j(\bm{X}) |\bm X \in B]\\
	&=\frac{1}{\PR(X \in B)}\int_{\bm{x} \in B} U_i(\bm{x}) U_j(\bm{x}) \mu (\bm{x}) \dif \bm{x}\\
	&=\frac{1}{\PR(X \in B)}\int_{\bm{x} \in B} U_i(\bm{x}) U_j(\bm{x}) \left(\mu (\bm{C_B})+\mu (\bm{x})-\mu (\bm{C_B})\right) \dif \bm{x}\\
	&\overset{(a)}=\frac{1}{\PR(X \in B)}\int_{\bm{x} \in B} U_i(\bm{x}) U_j(\bm{x}) \left(\mu (\bm{x})-\mu (\bm{C_B})\right) \dif \bm{x}\\
	& \leq \frac{1}{\PR(X \in B)}\int_{\bm{x} \in B} \left|U_i(\bm{x})\right| \left|U_j(\bm{x})\right| \left|\mu (\bm{x})-\mu (\bm{C_B})\right| \dif \bm{x}\\
	& \overset{(b)} \leq \frac{1}{\PR(X \in B)}\int_{\bm{x} \in B} \frac{1}{2} \cdot \frac{1}{2} \cdot \frac{1}{2} L_{\mu} h \dif \bm{x}\\
	&= \frac{L_{\mu} h^{d_x+1}}{8 \PR(X \in B)}\\
	& \leq \frac{L_{\mu} h}{8 \mu_m},
	\end{align*}
	where $(a)$ holds by \eqref{equ:norm_ineq2} and $(b)$ follows from $|\mu(\bm{x})-\mu(C_B) |\leq L_{\mu} \|\bm{x}-C_B\|_{\infty} \leq \frac{1}{2}L_{\mu} h$.
	Thus, the second condition of Assumption~\ref{ass:cov_des} is satisfied by choosing $\gamma=3 d_x^* L_{\mu} h/(2 \mu_m^2)$, and $\gamma <1$ if $h < 2 \mu_m^2/(3 d_x^* L_{\mu})$.
\end{proof}

\subsection{Proof of Lemma \ref{lem:existence-C}}

		We first prove the first point of Lemma \ref{lem:existence-C}. For the simplicity of notation, we denote $\partial f(\bm{x},y)/\partial x_i$ as $f_i'(\bm{x},y)$.
		By Assumption \ref{ass:continuous}, $f(\bm{x},y)$ are continuously differentiable with respect to $\bm{x} \in \XX$, $y \in \YY$. Then $f_i'(\bm{x},y)$ is a continuous function for all $i \in \{1,2,\ldots,d_x\}$.
		Applying the generalized extreme value theorem \footnote{See Theorem 4.16 on page 89 of \cite{rudin1964principles}}, we have two constants
		\begin{equation*}
		M_i=\sup_{\bm{x} \in \XX, y \in \YY} f_i'(\bm{x},y), \quad m_i=\inf_{\bm{x} \in \XX, y \in \YY} f_i'(\bm{x},y),
		\end{equation*}
		and there exist $\bm{x}_1 \in \XX, y_1 \in \YY$ such that $f_i'(\bm{x}_1,y_1)=M_i$ and $\bm{x}_2 \in \XX, y_2 \in \YY$ such that $f_i'(\bm{x}_2,y_2)=m_i$.
		By Assumption \ref{ass:pos_gra}, we know $M_i,m_i \neq 0$ for $i \in J$. Then by Theorem 4.22 on page 93 of \cite{rudin1964principles}, $f_i'(\XX,\YY)$ is a connected set. If $M_i>0>m_i$, then $o \in f_i'(\XX,\YY)$ and there exist $\bm{x}_3 \in \XX, y_3 \in \YY$ such that $f_i'(\bm{x}_3,y_3)=0$, which violates Assumption \ref{ass:pos_gra}. Thus, for $i \in J$, we have either $M_i>m_i>0$ or $0>M_i>m_i$. Let $C=\min_{i \in J}\{|m_i|,|M_i|\}$, we have $|f_i'(\bm{x},y)| \geq C$ for all $i \in J, \bm{x} \in \XX$ and $y \in \YY$. Thus, we prove the existence of $C$ satisfying \eqref{equ:global-C}.

		Next, we prove the second point in Lemma \ref{lem:existence-C}. Since we fix $\bm{x}_{(i)}$ in $\XX$, applying the generalized extreme value theorem, we have the constants
		\begin{equation*}
		M_i=\sup_{y \in \YY} f_i'(\bm{x}_{(i)},y), \quad m_i=\inf_{y \in \YY} f_i'(\bm{x}_{(i)},y).
		\end{equation*}
		Let $D=\min_{i \in J}\{|m_i|,|M_i|\}$, like the previous argument, we have $|f_i'(\bm{x}_{(i)},y)| \geq D$ for all $i \in J$ and $y \in \YY$.

		Furthermore, if $f$ is twice-differentiable with respect to $\bm{x}$, we will prove that the hypercube $\mathcal{H}_i$ with side length $\overline{h}=D/(2 L)$ and centred at $\bm{x}_{(i)}$ satisfies \eqref{equ:hypercube-H-C}.
		We omit the argument $y$ when writing $f(\bm{x},y)$ as we prove the result for any fixed $y$.
		For any $\bm{x} \in \mathcal{H}_i$, we write $\bm{x}=\bm{x}_{(i)}+\bm{l}$ where $\bm{l} \in \mathcal{R}^{d_x}$ and $\|\bm{l}\|_{\infty} \leq \overline{h}/2$. Define a function $\psi(t)=\nabla f(\bm{x}_{(i)}+ t\bm{l})$. As assumed in Assumption \ref{ass:smooth_cov}, $f$ is twice-differentiable, thus $\psi(t)$ is continuous differentiable. We have $\psi(t)=\psi(0)+\int_0^1 \psi'(t) \dif t$, i.e.,
		\begin{equation*}
		\nabla f(\bm{x}_{(i)}+\bm{l})= \nabla f(\bm{x}_{(i)})+\int_0^1 \nabla^2 f(\bm{x}_{(i)}+t \bm{l}) \bm{l} \dif t.
		\end{equation*}
		Then moving $\nabla f(\bm{x}_{(i)})$ to the left-hand-side, and taking infinity norm, we have
		\begin{equation}\label{equ:inf-gra}
		\left\|\nabla f(\bm{x}_{(i)}+\bm{l})-\nabla f(\bm{x}_{(i)})\right\|_{\infty}=\left\|\int_0^1 \nabla^2 f(\bm{x}_{(i)}+t \bm{l}) \bm{l} \dif t\right\|_{\infty}.
		\end{equation}
		According to the definition of infinity norm, for a matrix $A$, we have
		\begin{equation*}
		\|A\|_{\infty}=\max_{\bm{x} \neq \bm{0}} \frac{\|A \bm{x}\|_{\infty}}{\|\bm{x}\|_{\infty}}, \  \text{and} \ \|A \bm{x}\|_{\infty} \leq \|A\|_{\infty} \|\bm{x}\|_{\infty}.
		\end{equation*}
		Thus, we have
		\begin{equation}\label{equ:inf-norm-ine}
		\left\| \nabla^2 f(\bm{x}_{(i)}+t \bm{l}) \bm{l} \right\|_{\infty} \leq \left\| \nabla^2 f(\bm{x}_{(i)}+t \bm{l}) \right\|_{\infty} \left\|  \bm{l} \right\|_{\infty} \leq L \overline{h},
		\end{equation}
		where the last inequality holds by $\left\| \nabla^2 f(\bm{x}_{(i)}+t \bm{l}) \right\|_{\infty} \leq 2L$ (Assumption \ref{ass:smooth_cov}) and $\|\bm{l}\|_{\infty} \leq \overline{h}/2$.
		By \eqref{equ:inf-gra} and \eqref{equ:inf-norm-ine}, we have
		\begin{equation}\label{equ:gra-diff-norm}
		\left\|\nabla f(\bm{x}_{(i)}+\bm{l})-\nabla f(\bm{x}_{(i)})\right\|_{\infty}
		\leq \int_0^1\left\| \nabla^2 f(\bm{x}_{(i)}+t \bm{l}) \bm{l} \right\|_{\infty}\dif t
		\leq \int_0^1 L \overline{h} \dif t=L \overline{h}=D/2,
		\end{equation}
		where the last equality follows from $\overline{h}=D/(2L)$.
		We rewrite \eqref{equ:gra-diff-norm} in the form of partial derivatives:
		\begin{equation}\label{equ:max-diff-par}
		\Big|\frac{\partial f(\bm{x})}{\partial x_i}- \frac{\partial f(\bm{x}_{(i)})}{\partial x_i}\Big| \leq D/2, \quad \forall i \in J, \bm{x} \in \mathcal{H}_i.
		\end{equation}
		By the previous argument, we have
		\begin{equation}\label{equ:min-par-C}
		\Big| \frac{\partial f(\bm{x}_{(i)})}{\partial x_i}\Big| >D, \quad \forall i \in J.
		\end{equation}
		Combining \eqref{equ:max-diff-par} and \eqref{equ:min-par-C}, we have
		\begin{equation*}
		\Big| \frac{\partial f(\bm{x})}{\partial x_i}\Big| > D/2, \quad \forall i \in J, \bm{x} \in \mathcal{H}_i.
		\end{equation*}
		Finally, let $C=D/2$, we prove the existence of $C$ and $\mathcal{H}_i$ satisfying \eqref{equ:hypercube-H-C}.

\section{Proofs for Localized LASSO}\label{app:proof-local}
\subsection{Proof of Lemma~\ref{lem:proj}}\label{proof:proj}
		We first prove the first point of Lemma \ref{lem:proj}. Without loss of generality, we set the geometric centre in bin B as zero. Moreover, we omit the argument $y$ when writing $f(\bm{x},y)$ as $y$ is fixed in the proof.  In other words, we consider $\bm{x} \in B= [-\frac{h}{2},\frac{h}{2}]^{d_x}$ and we have
		\begin{equation}\label{equ:lemma_integral}
		\int_{\bm{x} \in B} x_i \dif \bm{x}=0 \quad \forall i \in \{1,2,\ldots,d_x\}.
		\end{equation}
		For $i \in J$ and $C>0$, if $\frac{\partial f(\bm{x})}{\partial x_i} \geq C$, then
		\begin{align}
		f(x_1,\ldots,x_{d_x})
		&\!=\!\int_{-h/2}^{x_i}  \frac{\partial f(x_1,\ldots,x_{i-1},z,x_{i+1},\ldots,x_{d_x})}{\partial z} \dif z \!+\!f(x_1,\ldots,x_{i-1},-\frac{h}{2},x_{i+1},\ldots,x_{d_x}) \notag\\
		& \geq C \cdot (x_i+\frac{h}{2})+f(x_1,\ldots,x_{i-1},-\frac{h}{2},x_{i+1},\ldots,x_{d_x}), \label{equ:partial-C}
		\end{align}
		where the first equality holds by the differentiability of $f$ (Assumption \ref{ass:continuous}).
		By the definition of $\theta_i$, we have
		\begin{align*}
		\theta_i
		&=\dfrac{\int_{\bm{x}\in B} [f(x_1,\dots,x_{d_x})-\theta_0] x_i \dif \bm{x}}{\int_{\bm{x}\in B} x_i^2  \dif \bm{x}}\\
		&\overset{(a)} \geq \dfrac{\int_{\bm{x}\in B} (C x_i^2 +h C x_i/2 +f(x_1,\ldots,x_{i-1},-h/2,x_{i+1},\ldots,x_{d_x})x_i-\theta_0 x_i) \ \dif \bm{x}}{\int_{\bm{x}\in B} x_i^2 \ \dif \bm{x}}\\
		&\overset{(b)} = \dfrac{\int_{\bm{x}\in B} C x_i^2  \ \dif \bm{x}}{\int_{\bm{x}\in B} x_i^2 \ \dif \bm{x}}\\
		&=C.
		\end{align*}
		where $(a)$ follows from \eqref{equ:partial-C} and $(b)$ holds by \eqref{equ:lemma_integral}.
		Following the previous argument, we have $\theta_i \leq -C$ if $\frac{\partial f(\bm{x})}{\partial x_i} \leq -C$.

		If $i \notin J$, according to Assumption~\ref{ass:spa_cov}, we have
		\begin{align*}
		\theta_i
		&=\dfrac{\int_{\bm{x}\in B} [f(x_1,\dots,x_{d_x})-\theta_0] x_i \dif \bm{x}}{\int_{\bm{x}\in B} x_i^2 \dif \bm{x}}\\
		&=\dfrac{\int_{[-h/2,h/2]^{d^*_x}} (g(x_1,\dots,x_{d^*_x})-\theta_0) \dif x_1\ldots dx_{d_x^*} \int_{[-h/2,h/2]^{d_x-d^*_x}}x_i \dif x_{d^*_x+1}\ldots dx_{d_x}}{\int_{\bm{x}\in B} x_i^2 \dif x_1 \ldots dx_{d_x}}\\
		&=0.
		\end{align*}
		In the second equation, we put relevant variables in the first $d_x^*$ dimensions and redundant variables in the remaining $d_x-d_x^*$ dimensions. The last equality follows from\\ $\int_{[-h/2,h/2]^{d_x-d^*_x}}x_i \dif x_{d^*_x+1}\ldots dx_{d_x}=0$ for $i \notin J$.

		Next, we prove the second point of Lemma \ref{lem:proj}.
		Let $P(\bm{x})=\theta_0+\sum_{i=1}^{d_x} \theta_i x_i$ and  $Q(\bm{x})=f(\bm{x})-P(\bm{x})$. We will prove $|Q(\bm{x})| \leq (2\sqrt{3}+1)  L d_x h^2$ in the following three steps.

		First, we claim that there must be a point $\bm{x_0} \in B$ such that $f(\bm{x_0})=P(\bm{x_0})$. From the definition of $\theta_0$ \eqref{equ:L2}, we know $\int_{\bm{x} \in B} Q(\bm{x}) \dif \bm{x}=\int_{\bm{x} \in B} f(\bm{x}) \dif \bm{x}-\theta_0=0$. Also, because $Q(\bm{x})$ is a continuous function from $B$ to $\mathbb{R}$, there must exist a point $\bm{x}_0 \in B$ such that $Q(\bm{x}_0)=0$.

		Second, we approximate $Q(\bm{x})$ by the Taylor series expansion 
		at point $\bm{x_0}$,
		\begin{equation*}
		|Q(\bm{x})-Q(\bm{x_0})-\nabla Q(\bm{x_0})^T (\bm{x}-\bm{x_0})| \leq \frac{1}{2}\|\nabla^2 Q(\bm{x_0})\|_{\infty} \|\bm{x}-\bm{x_0}\|_{\infty}^2.
		\end{equation*}
		By $Q(\bm{x_0})=0, \nabla Q(\bm{x_0})=\nabla f(\bm{x_0})-\bm{\theta}$, $\nabla^2 Q(\bm{x})=\nabla^2 f(\bm{x})$ and Assumption~\ref{ass:smooth_cov}, we have
		\begin{equation} \label{equ:tay_inq}
		|f(\bm{x})-P(\bm{x})-(\nabla f(\bm{x_0}) - \bm{\theta})^T (\bm{x}-\bm{x_0})| \leq L \|\bm{x}-\bm{x_0}\|^2_{\infty}.
		\end{equation}

		Third, we provide an upper bound for  $\|\nabla f(\bm{x_0})-\bm{\theta}\|_{\infty}$. Recalling the definition of $\theta_i$ \eqref{equ:L2}, we have
		\begin{align}
		&\frac{\partial f(\bm{x})}{\partial x_i} \Big|_{\bm{x}=\bm{x_0}}-\theta_i \notag\\
		&=\left(\int_{\bm{x} \in B} x_i^2 \dif \bm{x}\right)^{-1} \displaystyle{\int_{\bm{x} \in B} \frac{\partial f(\bm{x})}{\partial x_i} \bigg|_{\bm{x}=\bm{x}_0} x_i^2  \dif \bm{x}}-\left(\int_{\bm{x} \in B} x_i^2 \dif \bm{x}\right)^{-1} \int_{\bm{x} \in B} (f(\bm{x})-\theta_0) x_i \dif \bm{x} \notag\\
		&=\left(\int_{\bm{x} \in B} x_i^2 \dif \bm{x}\right)^{-1} \displaystyle{\int_{\bm{x} \in B} \frac{\partial f(\bm{x})}{\partial x_i} \bigg|_{\bm{x}=\bm{x}_0} x_i^2-(f(\bm{x})-\theta_0) x_i} \dif \bm{x} \notag\\
		&\overset{(a)}=\left(\int_{\bm{x} \in B} x_i^2 \dif \bm{x}\right)^{-1} \displaystyle{\int_{\bm{x} \in B} \frac{\partial f(\bm{x})}{\partial x_i} \bigg|_{\bm{x}=\bm{x}_0} x_i^2-f(\bm{x}) x_i} \dif \bm{x} \notag\\
		&\overset{(b)}=\left(\int_{\bm{x} \in B} x_i^2 \dif \bm{x}\right)^{-1}
		\displaystyle{\int_{\bm{x} \in B} \left( \int_{-h/2}^{x_i} \frac{\partial f(\bm{x})}{\partial x_i} \bigg|_{\bm{x}=\bm{x}_0}\!-\! \frac{\partial f(x_1,\ldots,x_{i-1},z,x_{i+1},\ldots,x_{d_x})}{\partial z}\dif z \right) x_i \dif \bm{x}}, \label{equ:lemma_Cauthy}
		\end{align}
		where $(a)$ follows from \eqref{equ:lemma_integral} and $(b)$ follows from writing $f(\bm{x})$ as the integration of $x_i$'s partial derivative and \eqref{equ:lemma_integral}. Then, by the Cauthy-Schwarz inequality, we have \eqref{equ:lemma_Cauthy}
		\begin{equation*}
		\!\leq\! \left(\int_{\bm{x} \in B} x_i^2 \dif \bm{x}\right)^{-1/2}
		\displaystyle{
			\left(\int_{\bm{x} \in B} \left( \int_{-h/2}^{x_i} \frac{\partial f(\bm{x})}{\partial x_i} \bigg|_{\bm{x}=\bm{x}_0}\!-\! \frac{\partial f(x_1,\ldots,x_{i-1},z,x_{i+1},\ldots,x_{d_x})}{\partial z}\dif z \right)^2 \dif \bm{x}\right)^{1/2}
		}.
		\end{equation*}
		According to Assumption~\ref{ass:smooth_cov} and following the same argument for \eqref{equ:gra-diff-norm}, we have $\|\nabla f(\bm{x_0})-\nabla f(\bm{x})\|_{\infty} \leq 2Lh$. Then, thus
		\begin{equation}\label{equ:norm-2Lh}
		\left|\frac{\partial f(\bm{x})}{\partial x_i} \bigg|_{\bm{x}=\bm{x}_0}- \frac{\partial f(x_1,\ldots,x_{i-1},z,x_{i+1},\ldots,x_{d_x})}{\partial z} \right|=\left|(\nabla f(\bm{x_0}))_i- (\nabla f(\bm{x}))_i\right| \leq 2Lh.
		\end{equation}
		By \eqref{equ:lemma_Cauthy}, \eqref{equ:norm-2Lh} and  $x_i \leq h/2$,  we have
		\begin{align*}
		\frac{\partial f(\bm{x})}{\partial x_i} \bigg|_{\bm{x}=\bm{x}_0}-\theta_i
		& \leq \left(\int_{\bm{x} \in B} x_i^2 \dif \bm{x}\right)^{-1/2} \left(\int_{\bm{x} \in B} \left(\int_{-h/2}^{x_i} 2L h \dif z\right)^2 \dif \bm{x} \right)^{1/2}\\
		& \leq \left(\int_{\bm{x} \in B} x_i^2 \dif \bm{x}\right)^{-1/2} \left(\int_{\bm{x} \in B} 4L^2 h^4 \dif \bm{x} \right)^{1/2}\\
		&=\left( \frac{48 L^2 h^{d_x+4}}{h^{d_x+2}}\right)^{1/2}=4\sqrt{3}Lh\\
		\end{align*}
		Taking maximum of all $i \in \{1,2,\ldots,d_x\}$, we have
		\begin{equation*}
		\|\nabla f(\bm{x_0})-\bm{\theta}\|_{\infty} \leq 4\sqrt{3}Lh.
		\end{equation*}
		Plugging it into~\eqref{equ:tay_inq}, we have
		\begin{align*}
		|f(\bm{x})-P(\bm{x})|
		& \leq  L\|\bm{x}-\bm{x_0}\|_{\infty}^2 + |(\nabla f(\bm{x_0})-\bm{\theta})^T (\bm{x}-\bm{x_0})|\\
		&\leq L \|\bm{x}-\bm{x_0}\|_{\infty}^2 + \|\nabla f(\bm{x_0})-\bm{\theta}\|_{\infty} \|\bm{x}-\bm{x_0}\|_1\\
		& \leq L  h^2 + 4\sqrt{3} Ld_x h^2\\
		& \leq (4\sqrt{3}+1) L d_x h^2.
		\end{align*}
	Hence, we complete the proof of Lemma~\ref{lem:proj}.

\subsection{Proof of Proposition~\ref{prop:subset}}\label{proof:subset}
	{\RV \Copy{cop:local-LASSO-proof}{Note that before applying localized LASSO in bin $B_j$, we observe the covariates $\{\bm{X}_t\}_{t=1}^{n_j}$ falling into bin $B_j$. So we define the event 
	\begin{equation*}
	\Omega_0=\{\bm{X}_1 \in B_j, \ldots, \bm{X}_{n_j} \in B_j\}.
	\end{equation*}
	All the probabilities in the proof is conditional on the event  $\Omega_0$. }}
	By Lemma \ref{lem:proj}, we know $\bm{\theta}^*$ in \eqref{equ:L2} maintains the sparsity structure of $f$, i.e.,
	\begin{equation*}
	J=\left\{i \in \{1,2,\ldots,d_x\}: \theta_i^* \neq 0 \right\}.
	\end{equation*}
	As in \eqref{equ:lasso}, the variable set selected by LASSO is
	\begin{equation*}
	\hat{J}_j=\{i \in \{1,2,\ldots,d_x\}: \hat{\theta}_i \neq 0 \},
	\end{equation*}
	where $\hat{\bm{\theta}}$ is the LASSO estimator.
	If $\hat{\theta}_i$ has the same sign with  $\theta_i^*$ for all $i \in \{1,2,\ldots,d_x\}$, then we have $J=\hat{J}_j$. That's to say, the event $\{J=\hat{J}_j| \Omega_0\}$ contains the event $\{\sign(\hat{\bm{\theta}})=\sign(\bm{\theta}^*)| \Omega_0\}$. Mathematically,
	\begin{equation}\label{equ:set-ge-sign}
	\PR \left(J=\hat{J}_j| \Omega_0\right) \ge
	\PR \left(\sign(\hat{\bm{\theta}})=\sign(\bm{\theta}^*)| \Omega_0\right)
	\end{equation}
	By Lemma \ref{lem:sign-events}, we know the event $\{\sign(\hat{\bm{\theta}})=\sign(\bm{\theta}^*)| \Omega_0\}$ contains the event $\{\cap_{i=1}^4 \Omega_i| \Omega_0\}$. Thus, we have
	\begin{equation}\label{equ:sign-ge-event}
	\PR \left(\sign(\hat{\bm{\theta}})=\sign(\bm{\theta}^*)| \Omega_0\right) \ge \PR \left(\cap_{i=1}^4 \Omega_i| \Omega_0\right).
	\end{equation}
	By Lemma \ref{lem:prob-bound-good}, we have
	\begin{equation}\label{equ:event-ge-prob}
	\PR \left(\cap_{i=1}^4 \Omega_i| \Omega_0\right) \ge 1-b_0 \exp(b_1 n_j h^4).
	\end{equation}
	Therefore, by \eqref{equ:set-ge-sign}, \eqref{equ:sign-ge-event} and \eqref{equ:event-ge-prob}, we have
	\begin{equation*}
	\PR \left(J=\hat{J}_j| \Omega_0\right) \ge 1-b_0 \exp(b_1 n_j h^4).
	\end{equation*}
	Hence, we complete the proof of Proposition \ref{prop:subset}.

\subsection{Proof of Lemma~\ref{lem:sign-events}}
	Using the notation in Section \ref{sec:analysis_local_LASSO}, the LASSO estimator~\eqref{equ:lasso} can be formulated as
	\begin{equation}\label{equ:lasso_obj}
	\hat{\Theta}=\argmin_{{\bm{\theta}}\in \mathbb{R}_{d_x+1}} \|\bm{Z}-A {\bm{\theta}}\|_2^2 +2 \lambda \|{\bm{\theta}}\|_1.
	\end{equation}
	Note that $\hat\Theta$ can be a set.
	By \eqref{eq:zhao-yu-lem1}, we know that $\bm \theta\in \hat\Theta$ if and only if it satisfies
	\begin{equation}\label{equ:lasso_sol}
	\left\{
	\begin{array}{l}
	A_{(1)}^T \left(\bm{Z} - A {\bm{\theta}} \right)= \lambda \sign \left({\bm{\theta}}_{(1)}\right)\\
	\left|A_{(2)}^T \left(\bm{Z}- A {\bm{\theta}}\right)\right| \preceq \lambda \bm{e} \\
	\end{array}
	\right.
	\end{equation}
	where the notation $|\cdot|$ takes the absolute value of every entry, $\preceq$ conducts entry-wise comparison and $\bm{e}$ denotes the unit vector in $\mathbb{R}^{d_x-d_x^*}$.

	We will first prove that on the event $\Omega_1$, the LASSO estimator $\hat{\Theta}$ is unique, thus denoted as $\hat{\bm\theta}$. Let $\phi({\bm{\theta}}):=\|\bm{Z}-A {\bm{\theta}}\|_2^2 +2 \lambda \|{\bm{\theta}}\|_1$ be the objective function in \eqref{equ:lasso_obj}.
	Taking the second-order derivative, we have $\phi''({\bm{\theta}})=2 A^T A=2 \hat{\Psi}$.
	Under event $\Omega_1$, $\hat{\Psi}$ is positive definite, implying that $\phi({\bm{\theta}})$ is strongly convex with respect to ${\bm{\theta}}$.
	Therefore, the solution to \eqref{equ:lasso_obj} exists and is unique.

	Next, we will prove on the event $\Omega_1 \cap \Omega_2 \cap \Omega_3 \cap \Omega_4$, there exists  $\bm{\theta} \in \mathbb{R}^{d_x+1}$ satisfying $\sign \big({\bm{\theta}}\big)=\sign \big(\bm{\theta}^*\big)$ and ${\bm{\theta}} \in \hat{\Theta}$. Thus, by the uniqueness of the LASSO estimator, we have $\sign \big({\hat{\bm{\theta}}}\big)=\sign \big(\bm{\theta}^*\big)$. The proof mainly follows the line of Proposition 1 in \citet{zhao2006model}. But the notations and details are somewhat different. So we write down the whole proof.

	Let $\bm{\theta}_{(1)} \in \mathbb{R}^{d_x^*+1}$ and $\bm{\theta}_{(2)} \in \mathbb{R}^{d_x-d_x^*}$ such that
	\begin{equation}\label{eq:def-theta1}
	\bm{\theta}_{(1)}=\bm{\theta^*}_{(1)}+\hat{\Psi}_{11}^{-1} A_{(1)}^T \bm{\rho}-\lambda \hat{\Psi}_{11}^{-1} \sign \left(\bm{\theta}^*_{(1)}\right), \quad \bm{\theta}_{(2)}=\bm{\theta^*}_{(2)}=\bm{0}
	\end{equation}
	Then, on the event $\Omega_3$, we have
	\begin{equation*}
	\left|\bm{\theta}_{(1)}-\bm{\theta^*}_{(1)}\right| = \left|\hat{\Psi}_{11}^{-1} A_{(1)}^T \bm{\rho}-\lambda \hat{\Psi}_{11}^{-1} \sign \left(\bm{\theta}^*_{(1)}\right)\right| \preceq \left|\bm{\theta^*}_{(1)}\right|
	\end{equation*}
	The above inequality implies that $\sign \big({\bm{\theta}_{(1)}}\big)=\sign \big(\bm{\theta}_{(1)}^*\big)$.
	Moreover, multiplying both sides of \eqref{eq:def-theta1} by $\hat\Psi_{11}$, we have
	\begin{equation}\label{equ:lasso_sign1}
	\hat{\Psi}_{11}\left(\bm{\theta^*}_{(1)}-\bm{\theta}_{(1)}\right)+A_{(1)}^T \bm{\rho}=\lambda \sign \big(\bm{\theta}_{(1)}^*\big)=\lambda \sign \big(\bm{\theta}_{(1)}\big).
	\end{equation}
	Similarly, multiplying both sides of \eqref{eq:def-theta1} by $\hat\Psi_{21}$ yields
	\begin{align}
	\left|\hat{\Psi}_{21} \left(\bm{\theta^*}_{(1)}-\bm{\theta}_{(1)}\right) +A_{(2)}^T \bm{\rho}\right|
	&=\left|\lambda \hat{\Psi}_{21} \hat{\Psi}_{11}^{-1} \sign\big(\bm{\theta}_{(1)}\big)-\hat{\Psi}_{21} \hat{\Psi}_{11}^{-1} A_{(1)}^T \bm{\rho}+A_{(2)}^T \bm{\rho}\right| \notag\\
	&\leq \left|\lambda \hat{\Psi}_{21} \hat{\Psi}_{11}^{-1} \sign\big(\bm{\theta}_{(1)}\big)\right|+\left|\hat{\Psi}_{21} \hat{\Psi}_{11}^{-1} A_{(1)}^T \bm{\rho}-A_{(2)}^T \bm{\rho}\right| \label{equ:lasso_sign2}
	\end{align}
	On the event $\Omega_4$, the second term in \eqref{equ:lasso_sign2},
	$\left|\hat{\Psi}_{21} \hat{\Psi}_{11}^{-1} A_{(1)}^T \bm{\rho}-A_{(2)}^T \bm{\rho}\right| \preceq \frac{1}{2}(1-\gamma) \lambda \bm{e}$. On event $\Omega_1 \cap \Omega_2$, we show an upper bound for the first term in \eqref{equ:lasso_sign2}
	\begin{align*}
	\left|\left(\hat{\Psi}_{21} \hat{\Psi}_{11}^{-1} \sign\left(\bm{\theta}_{(1)}\right)\right)_j\right|
	&= \left|\sum_{k=0}^{d_x^*} (\hat{\Psi}_{21})_{jk} \left(\hat{\Psi}_{11}^{-1} \sign (\bm{\theta}_{(1)})\right)_k\right|\\
	&\overset{(a)} \leq \left(\sum_{k=0}^{d_x^*} (\hat{\Psi}_{21})^2_{jk}\right)^{1/2} \left\|\hat{\Psi}_{11}^{-1} \sign(\bm{\theta}_{(1)})\right\|_2\\
	&\overset{(b)} \leq \sqrt{d^*_x} (1+\delta)\gamma \underline{\lambda}/d^*_x \|\hat{\Psi}_{11}^{-1}\|_2\|\sign(\bm{\theta}_{(1)})\|_2\\
	&\overset{(c)} \leq \sqrt{d^*_x} (1+\delta)\gamma \underline{\lambda}/d^*_x \cdot \frac{\sqrt{d^*_x}}{(1-\alpha) \underline{\lambda}}\\
	& = \frac{(1+\delta) \gamma}{1-\alpha} = \frac{1}{2} (1+ \gamma),
	\end{align*}
	where $(a)$ follows from the Cauthy-Schwarz inequality, $(b)$ follows from the definition of $\Omega_2$ as well as the matrix inequality $\|AB\|_2\le \|A\|_2\|B\|_2$ and $(c)$ is due to $\|A\|_2\le \lambda_{\max}(A)$ for square matrices.
	Combining the two terms in  \eqref{equ:lasso_sign2}, we have
	\begin{equation}\label{equ:lasso_sign3}
	\left|\hat{\Psi}_{21} \left(\bm{\theta^*}_{(1)}-\bm{\theta}_{(1)}\right) +A_{(2)}^T \bm{\rho}\right| \preceq \frac{1}{2} (1- \gamma) \lambda \bm{e}+\frac{1}{2} (1+ \gamma) \lambda \bm{e}= \lambda \bm{e}.
	\end{equation}
	By~\eqref{equ:lasso_sign1},~\eqref{equ:lasso_sign3} and $\bm{\theta}_{(2)}=\bm{\theta^*}_{(2)}=\bm{0}$, we have
	\begin{equation}\label{equ:lasso_plug}
	\left\{
	\begin{array}{l}
	\hat{\Psi}_{11} \left(\bm{\theta}^*_{(1)}-{\bm{\theta}}_{(1)}\right)+\hat{\Psi}_{12} \left(\bm{\theta}^*_{(2)}-{\bm{\theta}}_{(2)}\right)+A_{(1)}^T \bm{\rho}=\lambda \sign \left({\bm{\theta}}_{(1)}\right)\\
	\left|\hat{\Psi}_{21} \left(\bm{\theta}^*_{(1)}-{\bm{\theta}}_{(1)}\right)+\hat{\Psi}_{22} \left(\bm{\theta}^*_{(2)}-{\bm{\theta}}_{(2)}\right)+A_{(2)}^T \bm{\rho} \right| \preceq \lambda \bm{e}\\
	\end{array}
	\right.
	\end{equation}
	Notice that \eqref{equ:lasso_sol} is equivalent to \eqref{equ:lasso_plug} by $\bm{Z}=A \bm{\theta}^*+\bm{\rho}$.
	Therefore, we have found $\bm{\theta}$ having the same signs with $\bm{\theta^*}$ and satisfying~\eqref{equ:lasso_sol}. Further by the uniqueness of the LASSO estimator, we have $\bm{\theta}=\hat{\bm{\theta}}$ and $\sign \big({\hat{\bm{\theta}}}\big)=\sign \big(\bm{\theta}^*\big)$. Hence, we have proved that on the event $\Omega_1 \cap \Omega_2 \cap \Omega_3 \cap \Omega_4$, the LASSO estimator has the same signs with the linear approximation.
\subsection{Proof of Lemma~\ref{lem:prob-bound-good}}
	In this proof, we will show that $\Omega_1 \cap \Omega_2 \cap \Omega_3 \cap \Omega_4$ occurs with a high probability. First, we adopt matrix concentration inequalities to give a lower bound for $\PR(\Omega_1)$.
	Recalling that $\hat{\Psi}$ is the empirical estimate of the conditional covariance matrix $\Psi$ in Assumption \ref{ass:cov_des}.
For any constant $\alpha \in (0,1)$, we have
\begin{align}\label{equ:covar-low-lambda}
\PR \left(\lambda_{\min} (\hat{\Psi}) \leq (1-\alpha) \underline{\lambda}\right)
& \leq \PR \left(\lambda_{\min} (\hat{\Psi}) \leq (1-\alpha) \lambda_{\min}(\Psi)\right) \notag\\
& \leq (d_x+1) \left(\frac{e^{-\alpha}}{(1-\alpha)^{(1-\alpha)}}\right)^{n \lambda_{\min}(\Psi)/(1+d_x/4)} \notag\\
& \leq (d_x+1) \left(\frac{e^{-\alpha}}{(1-\alpha)^{(1-\alpha)}}\right)^{n \underline{\lambda}/(1+d_x/4)}.
\end{align}
The first and last inequalities follow from $\underline{\lambda} \leq \lambda_{\min}(\hat{\Psi})$.
The second inequality follows from Theorem 5.1.1 in \citet{tropp2015introduction}:

\begin{thm}[Theorem 5.1.1 in \citet{tropp2015introduction}] \label{thm:matrix-con}
	Consider a finite sequence of i.i.d. random Hermitian matrices $M_t\in \mathbb R^{(d_x+1)\times(d_x+1)}$. Assume that
	\begin{equation*}
	0 \leq \lambda_{\min} (M_t M_t^T) \quad and \quad \lambda_{\max} (M_t M_t^T) \leq \lambda_M, \quad \forall \ t \in \{1,2,\ldots,n\},
	\end{equation*}
	and
	\begin{equation*}
	\Psi =\EX [M_t M_t^T], \
	\hat{\Psi}
	= \frac{1}{n} \sum_{t=1}^n M_t M_t^T.
	\end{equation*}
	Then, we have
	\begin{align*}
	&\mathbb{P} \left(\lambda_{\min} (\hat{\Psi}) \leq (1-\alpha) \lambda_{\min}(\Psi) \right) \leq (d_x+1) \left(\frac{e^{-\alpha}}{(1-\alpha)^{(1-\alpha)}}\right)^{n \lambda_{\min}(\Psi)/\lambda_M} \ \forall \alpha \in [0,1), \\
	&\mathbb{P} \left(\lambda_{\max} (\hat{\Psi}) \geq (1+\alpha) \lambda_{\max}(\Psi)\right) \leq (d_x+1) \left(\frac{e^{\alpha}}{(1+\alpha)^{(1+\alpha)}}\right)^{n \lambda_{\max}(\Psi)/\lambda_M} \ \forall \alpha \geq 0.
	\end{align*}
\end{thm}
Here, to apply Theorem~\ref{thm:matrix-con}, we let $M_t=\bar{\bm{U}}_t$ and show an upper bound for $\lambda_{\max}(\bar{\bm{U}}_t \bar{\bm{U}}_t^T)$.
Recalling that $\bar{\bm{U}}_t$ is the normalized covariates, the absolute of each entry is less than $1/2$ except for the first entry, which is $1$. So the $\ell_2$-norm $\|\bar{\bm{U}}_t\|_2^2$ is less than $(1+d_x/4)$. By the Cauchy-Schwartz inequality, we have
\begin{equation*}
\bm{u}^T \bar{\bm{U}}_t \bar{\bm{U}}_t^T \bm{u} = \left(\bm{u}^T \bar{\bm{U}}_t\right)^2 \leq \|\bm{u}\|_2^2 \|\bar{\bm{U}}_t\|_2^2 \leq (1+d_x/4) \|\bm{u}\|_2^2
\end{equation*}
for any $\bm{u} \in \mathcal{R}^{d_x+1}$. Further, considering the characterization of eigenvalues, for a symmetric matrix A, its largest eigenvalue satisfies
\begin{equation}\label{eq:eigenvalue}
\lambda_{\max}(A)=\sup_{u} \frac{u^TAu}{\|u\|_2^2}.
\end{equation}
As a result,
\begin{equation*}
\lambda_{\max}(\bar{\bm{U}}_t \bar{\bm{U}}_t^T)=\sup_{u} \frac{u^T \bar{\bm{U}}_t \bar{\bm{U}}_t^T u}{\|u\|_2^2} \leq (1+d_x/4).
\end{equation*}
So we set the constant $\lambda_M=1+d_x/4$ in the above theorem.
In this way, we obtain the second inequality of \eqref{equ:covar-low-lambda}.
Moreover, we have
\begin{equation}\label{equ:lasso_range1}
0 <\frac{e^{-\alpha}}{(1-\alpha)^{(1-\alpha)}} \leq e^{-\alpha^2/2} <1, \quad \text{for} \ \alpha \in (0,1).
\end{equation}
Similarly, using Theorem~\ref{thm:matrix-con}, we have an probability bound for $\lambda_{\max} (\hat{\Psi})$:
\begin{align}
\PR \left(\lambda_{\max} (\hat{\Psi}) \geq (1+\alpha) \overline{\lambda}\right)
& \leq \PR \left(\lambda_{\max} (\hat{\Psi}) \geq (1+\alpha) \lambda_{\max}(\Psi)\right) \notag\\
& \leq (d_x+1) \left(\frac{e^{\alpha}}{(1+\alpha)^{(1+\alpha)}}\right)^{n \lambda_{\max}(\Psi)/(1+d_x/4)} \notag\\
& \leq (d_x+1) \left(\frac{e^{\alpha}}{(1+\alpha)^{(1+\alpha)}}\right)^{n \underline{\lambda}/(1+d_x/4)}, \label{equ:max-eigen-psi}
\end{align}
and
\begin{equation}\label{equ:lasso_range2}
0<\frac{e^{\alpha}}{(1+\alpha)^{(1+\alpha)}} <1, \quad \text{for} \ \alpha \in (0,1).
\end{equation}
Recall the definition of $\Omega_1$, choosing the constant $\alpha=\frac{1-\gamma}{2(1+\gamma)}$ and by \eqref{equ:covar-low-lambda}, \eqref{equ:max-eigen-psi}, we have
\begin{align}
\PR (\Omega_1)
&=\PR \left( \left\{(1-\alpha) \underline{\lambda} \ge \lambda_{\min}(\hat{\Psi}) \right\} \cup \left\{\lambda_{\max}(\hat{\Psi}) \leq (1+\alpha) \overline{\lambda}\right\} \right) \notag\\
&\geq 1- \PR \left(\lambda_{\min} (\hat{\Psi}) \leq (1-\alpha) \underline{\lambda}\right)-\PR \left(\lambda_{\max} (\hat{\Psi}) \geq (1+\alpha) \overline{\lambda}\right) \notag \\
&\geq 1-2 (d_x+1) \exp(-c_1 n), \label{equ:prob-omega1}
\end{align}
where
\begin{align*}
&\quad c_1(\underline{\lambda}, \gamma, d_x)\\
&=\frac{\underline{\lambda}}{(1+d_x/4)}\min\left\{-\log\left(\frac{e^{-\alpha}}{(1-\alpha)^{(1-\alpha)}}\right), -\log\left(\frac{e^{\alpha}}{(1+\alpha)^{(1+\alpha)}}\right)\right\}\\
&=\frac{\underline{\lambda}}{(1+d_x/4)} \min\left\{\alpha+(1\!-\!\alpha)\log(1-\alpha), -\alpha+(1+\alpha)\log(1+\alpha)\right\}\\
&=\frac{\underline{\lambda}}{2 (1+\gamma)(1+d_x/4)} \min\left\{1-\gamma+(3\gamma +1) \log\left( \frac{3\gamma+1}{2+2\gamma}\right), \gamma\!-\!1+(3+\gamma) \log\left(\frac{3+\gamma}{2+2\gamma}\right) \right\}.
\end{align*}
By \eqref{equ:lasso_range1} and \eqref{equ:lasso_range2}, we have $c_1>0$ as $\gamma \in [0,1)$.

Next, we show the event $\Omega_2$ happens with high probability.
Recalling
\begin{equation*}
(\hat{\Psi}_{21})_{ik}= \frac{1}{n}\sum_{j=1}^n (\bm U_j)_i(\bm U_j)_k,
\end{equation*}
Hoeffding's inequality\footnote{See Theorem 2.2.6 on page 18 of \citet{vershynin2018high}.} implies that
\begin{equation*}
\PR\left(\left|(\hat{\Psi}_{21})_{ik}- (\Psi_{21})_{ik}\right| \geq \delta \gamma \underline{\lambda}/d^*_x \right) \leq 2\exp(-2n \delta^2 \gamma^2 \underline{\lambda}^2/(d^*_x)^2).
\end{equation*}
According to Assumption~\ref{ass:cov_des}, $|(\Psi_{21})_{ik}| \leq \gamma \underline{\lambda}/d^*_x$. Thus, we have
\begin{equation*}
\PR\left(\left|(\hat{\Psi}_{21})_{ik}\right| \geq (1+\delta)\gamma \underline{\lambda}/d^*_x \right) \leq 2\exp(-2n \delta^2 \gamma^2 \underline{\lambda}^2/(d^*_x)^2).
\end{equation*}
Taking the union bound over $i\in J$ and $k\in J^c$
\begin{equation}\label{equ:prob-omega2}
\PR(\Omega_2) \geq 1-2 d^*_x (d_x-d^*_x)\exp(-c_3 n),
\end{equation}
where
\begin{equation*}
c_3(\gamma,\underline{\lambda},d^*_x)=2 \delta^2 \gamma^2 \underline{\lambda}^2/(d^*_x)^2=(1-\gamma)^2 \underline{\lambda}^2/(8 (d^*_x)^2).
\end{equation*}

Next, we show an upper bound for the approximation error of the linear projection. We define the vector
\begin{equation*}
\bm\Delta\coloneqq \frac{1}{\sqrt{n}} \begin{pmatrix} \Delta_1,\dots,\Delta_n \end{pmatrix}.
\end{equation*}
Then by Lemma~\ref{lem:proj}, we have
\begin{equation}\label{equ:lasso_Delta_norm}
\|\bm{\Delta}\|_2^2=\frac{1}{n} \sum_{t=1}^n \left(f(\bm X_t)-\bar{\bm U}_t^T \bm{\theta}^*\right)^2 \leq 64 L^2 d_x^2 h^4.
\end{equation}

So far, we have provided a lower bound for the probability of the event $\Omega_1 \cap \Omega_2 $.
It then suffices to bound the probabilities $\PR(\Omega_3^c \cap \Omega_1 \cap \Omega_2)$ and $\PR(\Omega_4^c \cap \Omega_1 \cap \Omega_2)$.
Recall the definition of event $\Omega_4$, the term $\hat{\Psi}_{21} \hat{\Psi}_{11}^{-1} A_{(1)}^T \bm{\rho}-A_{(2)}^T \bm{\rho}=\left(\hat{\Psi}_{21} \hat{\Psi}_{11}^{-1} A_{(1)}^T-A_{(2)}^T\right) \left(\bm{\Delta}+\frac{1}{\sqrt{n}}\bm{\epsilon}\right)$ is a linear combination of approximation errors $\bm{\Delta}$ and sub-Gaussian noises $\bm{\epsilon}$. Denote
\begin{equation}\label{equ:def-mat-G}
G\coloneqq\hat{\Psi}_{21} \hat{\Psi}_{11}^{-1} A^T_{(1)}- A^T_{(2)}=(g_{jk})_{d^*_x+1\leq j \leq d_x; 1\leq k \leq n},
\end{equation}
then we have
\begin{equation*}
\Omega_4=\left\{\left|G\left(\bm{\Delta}+\frac{1}{\sqrt{n}} \bm{\epsilon}\right)\right| \preceq \frac{1}{2} (1-\gamma) \lambda \bm{e}\right\}.
\end{equation*}
We want to bound the probability of $\Omega_4^c$
\begin{align}\label{equ:lasso_step3_1}
\Omega_4^c
&=\bigcup_{j=d_x^*+1}^{d_x} \left\{\left(\left|G\left(\bm{\Delta}+\frac{1}{\sqrt{n}} \bm{\epsilon}\right)\right|\right)_j \geq \frac{1}{2} (1-\gamma) \lambda\right\}  \notag \\
& \subset\bigcup_{j=d_x^*+1}^{d_x} \left\{\left(\left| \frac{1}{\sqrt{n}} G \bm{\epsilon} \right|\right)_j \geq \frac{1}{2} (1-\gamma) \lambda -\left(\left|G \bm{\Delta} \right|\right)_j\right\}.
\end{align}

Note that by~\eqref{equ:cov-mat-part} and~\eqref{equ:def-mat-G}, we have
\begin{align}\label{equ:eigen-GGT}
G G^T
&=\left(\hat{\Psi}_{21} \hat{\Psi}_{11}^{-1} A^T_{(1)}- A^T_{(2)}\right) \left(A_{(1)} \hat{\Psi}_{11}^{-1} \hat{\Psi}_{12}-A_{(2)}\right) 	& (\text{by} \  \hat{\Psi}_{12}=\hat{\Psi}_{21}^T) \notag\\
&=\hat{\Psi}_{21} \hat{\Psi}_{11}^{-1} \hat{\Psi}_{12}-\hat{\Psi}_{21} \hat{\Psi}_{11}^{-1} A^T_{(1)} A_{(2)}-A^T_{(2)} A_{(1)} \hat{\Psi}_{11}^{-1} \hat{\Psi}_{12}+A^T_{(2)} A_{(2)} & ( \text{by} \ \hat{\Psi}_{11}^{-1}=A^T_{(1)} A_{(1)}) \notag\\
&=-A^T_{(2)} A_{(1)}\hat{\Psi}_{11} A_{(1)}^T A_{(2)}+A^T_{(2)} A_{(2)} & (\text{by} \ \hat{\Psi}_{12}=A^T_{(1)} A_{(2)}) \notag\\
&=A^T_{(2)}\left(I-A_{(1)}\hat{\Psi}_{11}^{-1}A_{(1)}^T\right)A_{(2)} \notag\\
&=A^T_{(2)} B A_{(2)}
\end{align}
where $B \coloneqq I-A_{(1)}\hat{\Psi}_{11}^{-1}A_{(1)}^T$. Notice that $B$ is symmetric and
\begin{equation*}
B^2=I -2A_{(1)}\hat{\Psi}_{11}^{-1}A_{(1)}^T+A_{(1)}\hat{\Psi}_{11}^{-1}A_{(1)}^T A_{(1)}\hat{\Psi}_{11}^{-1}A_{(1)}^T
=\left(I-A_{(1)}\hat{\Psi}_{11}^{-1}A_{(1)}^T\right)
=B.
\end{equation*}
So $B$ is an idempotent matrix whose eigenvalues are either 0 or 1 \citep{horn1990norms}.
Since $G G^T$ is a symmetric matrix, using~\eqref{eq:eigenvalue}, we derive an upper bound for $\lambda_{\max}(G G^T)$,
\begin{align}\label{equ:eigen-uGGTu}
u^TGG^Tu
&= (A_{(2)}u)^TB(A_{(2)}u) & 			(\text{by}~\eqref{equ:eigen-GGT}) \notag\\
&\le \lambda_{\max}(B)\|A_{(2)}u\|_2^2 & (\text{by}~\eqref{eq:eigenvalue} ) \notag\\
&= \lambda_{\max}(B)(u^T \hat\Psi_{22} u)^2 & (\text{by} \  \hat\Psi_{22}=A_{(2)}^TA_{(2)})\notag\\
&\le \lambda_{\max}(B)\lambda_{\max}(\hat\Psi_{22})\|u\|_2^2 & (\text{by}~\eqref{eq:eigenvalue})\notag\\
&\le
\lambda_{\max}(\hat\Psi_{22})\|u\|_2^2 & (\text{by} \ \lambda_{\max}(B) \leq 1)
\end{align}

Moreover, on the event $\Omega_{1}$, the eigenvalue of $\hat{\Psi}_{22}=A^T_{(2)} A_{(2)}$ are smaller than $(1+\alpha) \overline{\lambda}$.
Therefore, \eqref{equ:eigen-uGGTu} implies the eigenvalues of $GG^T$ are less than $(1+\alpha) \overline{\lambda}$. This implies that
\begin{equation}\label{eq:sum-gjk2}
\sum_{k=1}^n g_{jk}^2=(GG^T)_{jj} =\bm e_j^T G G^T\bm e_j\le \lambda_{\max}(GG^T) \|\bm e_j\|_2^2 \leq (1+\alpha) \overline{\lambda},
\end{equation}
for all $j \in \{d_x^*+1,d_x^*+2,\ldots,d_x\}$, where $\bm e_j$ is the $j$-th standard basis.
Thus, we have
\begin{equation}\label{equ:lasso_max_eigen}
\left(|G \bm{\Delta}|\right)_j=\left|\sum_{k=1}^n g_{jk} \Delta_k\right| \leq \left(\sum_{k=1}^n g^2_{jk}\right)^{1/2} \|\bm{\Delta}\|_2 \leq \sqrt{(1+\alpha) \overline{\lambda}} \|\bm{\Delta}\|_2.
\end{equation}
By~\eqref{equ:lasso_Delta_norm}, we have $\|\bm{\Delta}\|_2 \leq 8 L d_x h^2$, and so
\begin{equation}\label{equ:lasso_max_Gdel}
\max_{\{j=d_x^*+1,\ldots,d_x\}} \left(|G \bm{\Delta}|\right)_j \leq 8 \sqrt{(1+\alpha) \overline{\lambda}} L d_x  h^2.
\end{equation}
Recalling that we choose
\begin{equation}\label{equ:lasso_lambda}
\lambda = 32 \cdot \sqrt{\frac{(3+\gamma) \overline{\lambda}}{(1+\gamma)(1-\gamma)^2}} L d_x h^2
\end{equation}
in Proposition~\ref{prop:subset}, by~\eqref{equ:lasso_max_Gdel}, we have
\begin{equation*}
\frac{1}{2} (1-\gamma) \lambda -\left(|G \bm{\Delta}|\right)_j \geq \frac{1}{4} (1-\gamma) \lambda.
\end{equation*}

Thus, plugging it into \eqref{equ:lasso_step3_1}, we have
\begin{align}
\Omega_4^c \cap \Omega_1
&\subset
\left\{ \bigcup_{j=d_x^*+1}^{d_x} \left\{\left(\left| \frac{1}{\sqrt{n}} G \bm{\epsilon} \right|\right)_j \geq \frac{1}{4} (1-\gamma) \lambda \right\} \right\} \bigcap \Omega_1 \notag\\
& = \left\{ \bigcup_{j=d_x^*+1}^{d_x} D_j \right\} \bigcap \Omega_1. \label{equ:lasso_Omega4}
\end{align}
where $D_j \coloneqq \left(\left| \frac{1}{\sqrt{n}} G \bm{\epsilon} \right|\right)_j > \frac{1}{4} (1-\gamma) \lambda$.
Define the realization of normalized covariates as $\mathcal{U}_n \coloneqq \left\{\bar{U}_1,\bar{U}_2,\ldots,\bar{U}_n\right\}$. It provides the full information for the empirical covariance matrix $\hat{\Psi}$ and whether $\Omega_1$ happens.
Note that the covariates and noise are independent, so
\begin{equation*}
\left( \frac{1}{\sqrt{n}}G \bm{\epsilon} \right)_j = \sqrt{\frac{1}{n}} \sum_{k=1}^n g_{jk} \epsilon_k,
\end{equation*}
it's a mean-zero $\sqrt{\frac{1}{n}\sum_{k=1}^n g_{jk}^2} \sigma$ sub-Gaussian random variable conditional on $\mathcal{U}_n$.
So we have
\begin{align}
\PR (\Omega_4^c \cap \Omega_1)
&=\EX \left[\EX \left[\mathbb{I} (\Omega_4^c \cap \Omega_1) \big|\mathcal{U}_n\right]\right] & (\text{by the tower rule}) \notag\\
& \leq \EX \left[\EX \left[\mathbb{I} \left(\left\{\cup_j D_j\right\}\right)  \big|\mathcal{U}_n\right]\cdot \mathbb{I} (\Omega_1)\right] & (\text{by}~\eqref{equ:lasso_Omega4})\notag\\
& \leq \EX \left[\EX \left[\sum_{j=d_x^*+1}^{d_x }\mathbb{I} \left( D_j\right)  \Big|\mathcal{U}_n\right]\cdot \mathbb{I} (\Omega_1)\right] &(\text{by the union bound})\notag\\
&\le \sum_{j=d_x^*+1}^{d_x } \EX\left[\PR \left(\left(\left| \frac{1}{\sqrt{n}}G \bm{\epsilon} \right|\right)_j > \frac{1}{4}(1- \gamma)\lambda ~\Big|~ \mathcal{U}_n\right)\mathbb{I} (\Omega_1)\right]&\notag\\
&\le \sum_{j=d_x^*+1}^{d_x } \EX\left[ 2\exp\left(-\frac{ (1- \gamma)^2\lambda^2 n}{32 \sum_{k=1}^n g_{ik}^2 \sigma^2}\right)\mathbb{I} (\Omega_1)\right]&(\text{sub-Gaussian})\notag\\
&\le \sum_{j=d_x^*+1}^{d_x } 2\exp\left(-\frac{(1- \gamma)^2\lambda^2  n }{32 (1+\alpha) \overline{\lambda} \sigma^2}\right)\PR(\Omega_1)&\label{equ:lasso_con_prob2}
\end{align}
where the last inequality follows from \eqref{eq:sum-gjk2} on the event $\Omega_1$.
Plugging in the value of $\lambda$ ~\eqref{equ:lasso_lambda}, we have that \eqref{equ:lasso_con_prob2} is upper bounded by $2(d_x-d_x^*) \exp\left(-c_5 n h^{4}\right)\PR (\Omega_1)$, where $c_5=64 L^2 d_x^2 / \sigma^2$.

Similarly, we study event $\Omega_3$.
The term $\hat{\Psi}_{11}^{-1} A_{(1)}^T \bm{\rho}=\hat{\Psi}_{11}^{-1} A_{(1)}^T \left(\bm{\Delta}+\frac{1}{\sqrt{n}} \bm{\epsilon}\right)$ is also a linear combination of approximation errors and sub-Gaussian noises. Denote
\begin{equation}\label{equ:def-H}
H\coloneqq\hat{\Psi}_{11}^{-1} A^T_{(1)} =(h_{jk})_{0 \leq j \leq d_x^*;1 \leq k \leq n}.
\end{equation}
We have
\begin{align}
\Omega_3^c
&=\bigcup_{j=0}^{d_x^*} \left\{\left|\lambda \left(\hat{\Psi}_{11}^{-1} \sign \left(\bm{\theta}^*_1\right)\right)_j -\left(H \left(\bm{\Delta}+\frac{1}{\sqrt{n}} \bm{\epsilon}\right)\right)_j \right|> \left(\left|\bm{\theta}^*_{(1)}\right| \right)_j \right\}\\
& \subset \bigcup_{j=0}^{d_x^*} \left\{\left(\left|\frac{1}{\sqrt{n}} H \bm{\epsilon}\right|\right)_j > \left(\left|\bm{\theta}^*_{(1)}\right|\right)_j-\left(\left|H \bm{\Delta}\right|\right)_j-\lambda \left(\left|\hat{\Psi}_{11}^{-1} \sign \left(\bm{\theta}^*_1\right)\right|\right)_j\right\}.\label{equ:lasso_omega3}
\end{align}

Recall that $\left(\bm{\theta}^*_{(1)}\right)_j$ is the coefficient of $j$-th relevant variable scaled by $h$.
According to Lemma~\ref{lem:proj}, its absolute value is greater than $Ch$.\footnote{Without loss of generality, we assume $|\theta_0| \ge Ch$, since it doesn't matter whether $0 \in J$ or not.}
Next we analyze the second term of the right-hand side of \eqref{equ:lasso_omega3}.

Note that by~\eqref{equ:def-H} and~\eqref{equ:cov-mat-part} we have $H H^T=\hat{\Psi}_{11}^{-1}$.
On the event $\Omega_1$, the eigenvalues of $\hat{\Psi}_{11}^{-1}$ are smaller than $1/\left((1-\alpha) \underline{\lambda}\right)$.
Similar to \eqref{eq:sum-gjk2}, we have
\begin{equation}\label{equ:lasso_max_H}
\sum_{k=1}^n h_{jk}^2\leq \lambda_{\max} (\hat{\Psi}_{11}^{-1}) \leq \frac{1}{(1-\alpha)  \underline{\lambda}}.
\end{equation}
Thus, for $j \in \{0,1,\ldots,d_x^*\}$, we have
\begin{equation*}
\left(\left|H \bm{\Delta}\right|\right)_j \leq \left(\sum_{k=1}^n h_{jk}^2\right)^{1/2} \|\bm{\Delta}\|_2 \leq \sqrt{\frac{1}{(1-\alpha) \underline{\lambda}}} \|\bm{\Delta}\|_2.
\end{equation*}
Since~\eqref{equ:lasso_Delta_norm} $\|\bm{\Delta}\|_2^2 \leq 64 L^2 d_x^2 h^4$, we have
\begin{equation*}
\max_{j=1,\ldots,d_x^*} \left(\left|H \bm{\Delta}\right|\right)_j\leq \sqrt{\frac{64}{(1-\alpha) \underline{\lambda}}} L d_x h^2.
\end{equation*}
Moreover, we have
\begin{equation*}
\left(\left|\hat{\Psi}_{11}^{-1} \sign \left(\bm{\theta}^*_{(1)}\right)\right|\right)_j  \leq \left\|\hat{\Psi}_{11}^{-1} \sign \left(\bm{\theta}^*_{(1)}\right)\right\|_2 \leq \frac{1}{(1-\alpha) \underline{\lambda}} \|\sign(\bm{\theta}^*_{(1)})\|_2 \leq \frac{\sqrt{d^*_x}}{(1-\alpha) \underline{\lambda}}.
\end{equation*}
Since we choose $\lambda$ as in~\eqref{equ:lasso_lambda}, we have
\begin{align}
& \quad\left(\left|H \bm{\Delta}\right|\right)_j+\lambda \left(\left|\hat{\Psi}_{11}^{-1} \sign \left(\bm{\theta}^*_{(1)}\right)\right|\right)_j \notag\\
&\leq \left(4 \sqrt{\frac{(3+\gamma) \overline{\lambda} d_x^*}{(1+\gamma)(1-\gamma)^2 (1-\alpha) \underline{\lambda}}}+1 \right) 8 L d_x h^2 \sqrt{\frac{1}{(1-\alpha) \underline{\lambda}}} \notag\\
&\leq \left(5 \sqrt{\frac{2(3+\gamma) \overline{\lambda} d_x^*}{(1+3 \gamma)(1-\gamma)^2  \underline{\lambda}}} \right) 8 L d_x h^2 \sqrt{\frac{2(1+\gamma)}{(1+3\gamma) \underline{\lambda}}} \notag\\
&=80 L d_x h^2 \cdot \dfrac{\sqrt{(3+\gamma)(1+\gamma)\overline{\lambda}d_x^*}}{(1+3\gamma)(1-\gamma) \underline{\lambda}}. \label{equ:lasso_theta1_mid}
\end{align}
According to Lemma~\ref{lem:proj} and definition of $\bm{\theta}^*$, $\left(\left|\bm{\theta}^*_{(1)}\right|\right)_j \geq Ch$ for $j \in \{0,1,\ldots,d_x^*\}$. So if $h$ is sufficient small such that
\begin{equation}\label{equ:lasso_def_h}
h \leq \dfrac{C (1+3\gamma)(1-\gamma) \underline{\lambda}}{160 L d_x  \sqrt{(3+\gamma)(1+\gamma)\overline{\lambda}d_x}},\footnote{Since $d_x^*$ is unknown, we replace $d_x^*$ by $d_x$ for a more conservative condition for $h$.}
\end{equation}
then we have
\begin{equation}\label{equ:lasso_theta1}
\left(\left|\bm{\theta}^*_{(1)}\right|\right)_j \geq Ch \geq 160 L d_x h^2 \cdot \dfrac{\sqrt{(3+\gamma)(1+\gamma)\overline{\lambda}d_x^*}}{(1+3\gamma)(1-\gamma) \underline{\lambda}}.
\end{equation}
Combing~\eqref{equ:lasso_theta1_mid} and~\eqref{equ:lasso_theta1},
the right-hand side of \eqref{equ:lasso_omega3} is at least half of $\left(\left|\bm{\theta}^*_{(1)}\right|\right)_j$, i.e.,
\begin{equation}\label{equ:lasso_step3_2}
\left(\left|\bm{\theta}^*_{(1)}\right|\right)_j-\left(\left|H \bm{\Delta}\right|\right)_j-\lambda \left(\left|\hat{\Psi}_{11}^{-1} \sign \left(\bm{\theta}^*_{(1)}\right)\right|\right)_j \geq \frac{1}{2} \left(\left|\bm{\theta}^*_{(1)}\right|\right)_j > \frac{1}{2} C h.
\end{equation}

Notice that all the parameters in right-hand side of~\eqref{equ:lasso_omega3} are known constants. So the validity of \eqref{equ:lasso_step3_2} is assured by choosing a small enough $h$.
As $h$ and $\lambda$ satisfy \eqref{equ:lasso_omega3} and \eqref{equ:lasso_def_h}, we have
\begin{align}
\Omega_3^c \cap \Omega_1
&\subset \left\{\bigcup_{j=0}^{d_x^*} \left\{\left(\left|\frac{1}{\sqrt{n}} H \bm{\epsilon}\right|\right)_j > \frac{1}{2} C h\right\}\right\} \bigcap \Omega_1. \notag\\
&= \left\{\bigcup_{j=0}^{d_x^*} E_j\right\} \bigcap \Omega_1 \label{eq:omega3-bound}
\end{align}
where $E_j \coloneqq \left\{\left(\left|\frac{1}{\sqrt{n}} H \bm{\epsilon}\right|\right)_j > \frac{1}{2} C h\right\}$.
Recalling the independence of covariates and noise, we have
\begin{equation*}
\left(\left|\frac{1}{\sqrt{n}} H \bm{\epsilon}\right|\right)_j = \sqrt{\frac{1}{n}} \sum_{k=1}^n h_{ik} \epsilon_k,
\end{equation*}
is a mean-zero $\sqrt{\sum_{k=1}^n h_{ik}^2/n} \sigma$ sub-Gaussian random variable conditional on $\mathcal{U}_n$.
Similar to \eqref{equ:lasso_con_prob2}, we have
\begin{align}\label{equ:bound-omega3c}
\PR (\Omega_3^c \cap \Omega_1)
& \leq \sum_{j=0}^{d_x^* }\EX \left[ \PR \left( E_j  \big|\mathcal{U}_n\right)\mathbb{I} (\Omega_1)\right] &\notag\\
& \leq \sum_{j=0}^{d_x^* }\EX \left[\PR \left(\left(\left|\frac{1}{\sqrt{n}} H \bm{\epsilon}\right|\right)_j > \frac{1}{2} C h ~\Big |~ \mathcal{U}_n \right)\mathbb{I} (\Omega_1)\right]&\notag\\
& \leq \sum_{j=0}^{d_x^* }\EX \left[ 2\exp\left(-\frac{C^2h^2n }{8 \sum_{k=1}^n h_{ik}^2\sigma^2}\right) \mathbb{I} (\Omega_1)\right] & \text{(sub-Gaussian)} \notag\\
& \leq \sum_{j=0}^{d_x^* }\EX \left[ 2\exp\left(-\frac{C^2h^2(1-\alpha)\underline{\lambda}n }{8 \sigma^2}\right) \mathbb{I} (\Omega_1)\right] &(\text{by \eqref{equ:lasso_max_H}})
\end{align}
where the last inequality follows from \eqref{equ:lasso_max_H} on the event $\Omega_1$. Plugging the lower bound of $Ch$ \eqref{equ:lasso_theta1}, we have that \eqref{equ:bound-omega3c} is upper bounded by $2(d_x+1) \exp\left(-c_6 n h^{4}\right)\PR (\Omega_1)$, where $c_6=\dfrac{3200(3+\gamma) \overline{\lambda} L^2 d_x^3}{(1-\gamma) \sigma^2}$.

Until now, we have demonstrated the probability lower bounds for event $\Omega_1$, $\Omega_2$, $\Omega_3^c \cap \Omega_1 $  and $\Omega_4^c \cap \Omega_1 $. We complete the proof by combining them together,
\begin{align*}
&\quad\PR (\Omega_1 \cap \Omega_2 \cap \Omega_3 \cap \Omega_4)\\
&=\PR (\Omega_1)- \PR(\Omega_1 \cap (\Omega_2^c \cup \Omega_3^c \cup \Omega_4^c))\\
&=\PR (\Omega_1)-  \PR((\Omega_1 \cap \Omega_2^c) \cup (\Omega_1 \cap \Omega_3^c) \cup (\Omega_1 \cap \Omega_4^c))\\
& \overset{(a)} \geq \PR (\Omega_1)- \PR(\Omega_1 \cap \Omega_2^c)- \PR(\Omega_1 \cap \Omega_3^c)- \PR(\Omega_1 \cap \Omega_4^c)\\
& \overset{(b)}\geq \left[1-2(d_x-d_x^*)\exp(-c_5 nh^{4})-2 (d_x^*+1) \exp(-c_6 nh^{4})\right] \PR (\Omega_1)-\PR(\Omega_2^c)\\
& \overset{(c)} \geq \left[1-2 (d_x+1)\exp\left(-(c_5 \wedge c_6) nh^{4}\right)\right] \cdot \left[1-2(d_x+1) \exp(-c_1 n)\right]\\
& \quad -2 d_x^*(d_x-d_x^*) \exp(-c_3 n)\\
& \geq 1-4 (d_x+1)\exp\left(-(c_1 \wedge c_5 \wedge c_6) nh^{4}\right)-2 d_x^*(d_x-d_x^*) \exp(-c_3 n)\\
& \geq 1-2 \max\{2 (d_x+1), d_x^*(d_x-d_x^*)\} \exp\left(-(c_1 \wedge c_3 \wedge c_5 \wedge c_6) nh^{4}\right)\\
& = 1- c_7 \exp(-c_8 n h^{4}),
\end{align*}
where $c_7=2 \max\{2 (d_x+1), d_x^*(d_x-d_x^*)\} \leq \max\{2 (d_x+1), d_x^2/4\}$, and $c_8=c_1 \wedge c_3 \wedge c_5 \wedge c_6$. Note that the inequality $(a)$ holds by the union bound, $(b)$ holds by \eqref{equ:lasso_con_prob2}, \eqref{equ:lasso_max_H} and $(c)$ holds by \eqref{equ:prob-omega1}, \eqref{equ:prob-omega2}.

Finally, we define new constants $b_0,b_1,b_2,b_3$ to summarize the results,
\begin{align}
&b_0(d_x):=2\max\{2 (d_x+1), d_x^2/4\}, \notag\\
&b_1(d_x,\gamma,\underline{\lambda},\overline{\lambda},L,\sigma):=c_1 \wedge c_3  \wedge c_5 \wedge c_6 \notag \\
& \quad \quad \quad \quad \quad \quad \quad \, =\left\{c_1 \wedge (1-\gamma)^2 \underline{\lambda}^2/(8 (d_x^*)^2) \wedge 64 L^2 d_x^2/(2 \sigma^2) \wedge 3200(3+\gamma) \overline{\lambda} L^2 d_x^3 /\left((1-\gamma) \sigma^2 \right)\right\}, \notag\\
&c_1(d_x, \gamma, \underline{\lambda}):=\frac{\underline{\lambda}}{2 (1\!+\!\gamma) (1\!+\!d_x/4)} \min\left\{1\!-\!\gamma\!+\!(3\gamma \!+\!1) \log\left( \frac{3\gamma\!+\!1}{2\!+\!2\gamma}\right), \gamma\!-\!1\!+\!(3\!+\!\gamma) \log\left(\frac{3\!+\!\gamma}{2\!+\!2\gamma}\right) \right\}, \notag\\
&b_2(\overline{\lambda},\gamma,d_x):=32 \sqrt{\frac{(3+\gamma) \overline{\lambda}}{(1+\gamma)(1-\gamma)^2}} L d_x, \notag \\
&b_3(C,\overline{\lambda},\underline{\lambda}, \gamma, L, d_x)  :=\dfrac{C (1+3\gamma)(1-\gamma) \underline{\lambda}}{160 L d_x  \sqrt{(3+\gamma)(1+\gamma)\overline{\lambda}d_x}}. \label{equ:lasso-local-constants}
\end{align}
\label{page:b3-min}
{\RV \Copy{cop:b3-min}{Under Proposition \ref{prop:cov_des2}, to guarantee Assumption \ref{ass:cov_des},
we further assume $h < \mu_m^2/(3d_x^*L_{\mu})$.
Thus, $b_3$ is required to be less than $\mu_m^2/(3d_x^*L_{\mu})$.}}
We can replace $\underline{\lambda}$, $\overline{\lambda}, \gamma$ by $\mu_m/12$, $\mu_M, 0.5$. Given $\lambda=b_2 h^2$, we have
\begin{equation*}
\PR (\cap_{i=1}^4\Omega_i) \geq 1-b_0 \exp(b_1 n_j h^4),
\end{equation*}
where the constants
\begin{align*}
&b_0(d_x)=2\max\{2 (d_x+1), d_x^2/4\},\\
&b_1(d_x,\mu_m, \mu_M,L,\sigma)=\frac{11 \mu_m}{10^4 (1+d_x/4)} \wedge \mu_m^2/(4608 (d_x^*)^2) \wedge 64 L^2 d_x^2/(2 \sigma^2) \wedge 22400 \mu_M L^2 d_x^3/\sigma^2,\\
&b_2(d_x,\mu_M)=64 \sqrt{7 \mu_M/3} L d_x,\\
&b_3(d_x,\mu_m,\mu_M,L_{\mu},C) =\min\left\{C \mu_m/(768 \sqrt{21 \mu_m d_x}), \mu_m^2/(3 d_x^* L_{\mu})\right\}.
\end{align*}

\section{Proofs for Weighted Voting}\label{app:weighted-voting}
\subsection{Proof of Proposition~\ref{prop:global}}\label{app:weighted-voting-main}
{\RV \Copy{cop:weighted-voting-appendix}{
	Before proceeding the proof, we first define a random variable $\Delta$ which represents the bins which the covariates $\bm{X}_1,\ldots, \bm{X}_n$ fall in. More precisely, $\Delta$ has $(h^{-d_x})^n$ possible values:
	\begin{align*}
	&\Delta_1=\{\bm{X}_1 \in B_1, \bm{X}_2 \in B_1, \ldots, \bm{X}_n \in B_1\}, \ \Delta_2=\{\bm{X}_1 \in B_1, \bm{X}_2 \in B_1, \ldots, \bm{X}_n \in B_2\}, \ \ldots,\\
	&\Delta_{(h^{-d_x})^n}=\{\bm{X}_1 \in B_{h^{-d_x}}, \bm{X}_2 \in B_{h^{-d_x}}, \ldots, \bm{X}_n \in B_{h^{-d_x}}\}.
	\end{align*}
	The probability mass function of $\Delta$ depends on the distribution of $\bm{X}$. The number of observations $\{n_j\}_{j=1}^{h^{-d_x}}$ can be inferred by $\Delta$. For example, $\PR(\Delta=\Delta_1)=\left(\PR(\bm{X} \in B_1)\right)^n$, and when $\Delta=\Delta_1$, we have $n_1=n,n_2=0,\ldots,n_{h^{-d_x}}=0$. So conditional on $\Delta$, we know the values of  $\{n_j\}_{j=1}^{h^{-d_x}}$ and $\{p_j\}_{j=1}^{h^{-d_x}}$. 
	
	Next, we show the proof of Proposition \ref{prop:global}. Conditional on $\Delta$, we rewrite the probability in \eqref{equ:lasso_global_prob}:
	\begin{align}\label{equ:lasso_global_rewrite}
	\PR(\hat{J}^{(i)} \geq \xi | J^{(i)}=0, \Delta)
	&=\PR (e^{\eta \hat{J}^{(i)}} \geq e^{\eta \xi} | J^{(i)}=0, \Delta)\notag\\
	&\le \exp(-\eta\xi)\EX\left[\prod_{j=1}^{h^{-d_x}} \exp(\eta w_j J^{(i)}_j) \big| J^{(i)}=0,\Delta\right].
	\end{align}
	We claim that for two different bins $B_{j_1} \neq B_{j_2}$, the random variables $J^{(i)}_{j_1}$ and $J^{(i)}_{j_2}$ are independent conditional on $\Delta$. That's because the covariates in $B_{j_1}$ are independent of the covariates in $B_{j_2}$ if we know $\Delta$. Thus, we have
	\begin{equation}\label{equ:lasso_global_prod}
	\eqref{equ:lasso_global_rewrite}=\exp(-\eta\xi)\prod_{j=1}^{h^{-d_x}} \EX\left[ \exp(\eta w_j J^{(i)}_j) \big| J^{(i)}=0,\Delta\right].
	\end{equation}
	By Proposition \ref{prop:subset}, we have the probability bound for the binary variable $J^{(i)}_{j}$:
	\begin{equation*}
	\PR\left(J^{(i)}_{j}=1 \big| J^{(i)}=0, \Delta\right) \le p_j.
	\end{equation*}
	Thus, 
	\begin{align}
	\eqref{equ:lasso_global_prod} 
	&\le \exp(-\eta\xi) \prod_{j=1}^{h^{-d_x}} \EX\left[e^{\eta w_j} p_j + 1-p_j\right]\notag\\
	&= \exp(-\eta\xi) \prod_{j=1}^{h^{-d_x}} \EX\left[1+p_j (e^{\eta w_j}-1)\right]\notag\\
	& \le \exp(-\eta\xi) \prod_{j=1}^{h^{-d_x}} \EX\left[\exp\left(p_j (e^{\eta w_j}-1)\right)\right], \label{equ:lasso_global_ex}
	\end{align}
	which gives the objective in \eqref{equ:global_obj}. The last inequality follows by $1+x \leq e^x$ for $x \ge 0$. 
	By Lemma \ref{lem:global_min}, \eqref{equ:lasso_global_rewrite}, \eqref{equ:lasso_global_prod}, \eqref{equ:lasso_global_ex}, we have
	\begin{equation*}
	\PR(\hat{J}^{(i)} \geq \xi | J^{(i)}=0, \Delta) \leq V(\eta^*,\bm{w}^*),
	\end{equation*}
	which is a function of $\{p_j\}_{j=1}^{h^{-d_x}}$.
	Then, Lemma \ref{lem:global_max} shows a uniform upper bound for $V(\eta^*,\bm{w}^*)$ for any possible $\{p_j\}_{j=1}^{h^{-d_x}}$. 
	By Lemma \ref{lem:global_max}, we have 
	\begin{equation*}
	\PR(\hat{J}^{(i)} \geq \xi | J^{(i)}=0, \Delta)\leq \exp\left\{\xi\left(h^{-d_x} (1+\log b_0-\log \xi)-b_1 n h^{4}\right)\right\}, \, \forall \Delta. 
	\end{equation*}
	Thus, we obtain an upper bound for the tail probability
	\begin{align*}
	\PR(\hat{J}^{(i)} \geq \xi | J^{(i)}=0)
	&=\sum_{i=1}^{(h^{-d_x})^n} \PR\left(J^{(i)}_{j}=1 \big| J^{(i)}=0, \Delta\right) \PR(\Delta=\Delta_i)\\
	&\leq \exp\left\{\xi\left(h^{-d_x} (1+\log b_0-\log \xi)-b_1 n h^{4}\right)\right\}.
	\end{align*}
}
}

	By the same argument, if $J^{(i)}=1$, we have
	\begin{equation*}
	\PR(\hat{J}^{(i)} \leq 1-\xi | J^{(i)}=1) \leq \exp\left\{\xi\left(h^{-d_x} (1+\log b_0-\log \xi)-b_1 n h^{4}\right)\right\}.
	\end{equation*}
	 Combining them together, we have
	\begin{equation}\label{equ:concentration-xi}
	\PR(\big|\hat{J}^{(i)} - J^{(i)}\big| \leq \xi) \leq \exp\left\{\xi\left(h^{-d_x} (1+\log b_0-\log \xi)-b_1 n h^{4}\right)\right\}.
	\end{equation}
	We choose $\xi=0.5$ to make the two tail probability equivalent. So the variable $x_i$ is classified as relevant if and only if $J^{(i)} \geq 1/2$. And the misidentification probability for $x_i$ has the upper bound
	\begin{equation*}
	\PR \left(\left|\hat{J}^{(i)} - J^{(i)}\right| \geq \frac{1}{2}\right) \leq \exp\left\{\frac{1}{2}\left(h^{-d_x} (1+\log b_0+\log 2)-b_1 n h^{4}\right)\right\}.
	\end{equation*}
	Moreover, by the union bound of all variables, we have the probability lower bound for successful variable selection
	\begin{equation*}
	\PR(\hat{J}=J) \geq 1-d_x\exp\left\{\frac{1}{2}\left(h^{-d_x} (1+\log b_0+\log 2)-b_1 n h^{4}\right)\right\}.
	\end{equation*}
	Hence, we complete the proof of Proposition \ref{prop:global}.

\subsection{\RV Proof of Lemma~\ref{lem:global_min}}\label{proof:lem:global_min}

Since the problem \eqref{equ:global_obj} involves minimizing a continuous function over a compact set \footnote{It's obvious that $V \rightarrow +\infty$ as $\eta \rightarrow +\infty$, so the minimum is obtained when $\eta$ is finite.}, it has an global minimal solution.
In the proof of Lemma \ref{lem:global_min}, we will prove the KKT condition admits a unique solution. {\RV Since the KKT condition is a necessary condition for all the local minimums and local maximums, then the unique solution must be the global minimum for problem \eqref{equ:global_obj}.} 
Considering the optimal $\eta^*$, if $\eta^*=0$, then $V(0,\bm{w})=0$ for any $\bm{w}$. Next we study the local optima with $\eta^*>0$. Finally, we compare the optimal $V$ in the two cases.

Supposing $\eta^*>0$, by the first-order optimality condition, we have
\begin{equation*}
0=\frac{\partial V(\eta,\bm{w})}{\partial \eta}=-\xi+\sum_{j=1}^{h^{-d_x}} p_j w_j e^{\eta w_j}.
\end{equation*}
Since $V(\eta,\bm{w})>0$, we have
\begin{equation}\label{equ:lasso_global_eta}
\sum_{j=1}^{h^{-d_x}} p_j w_j e^{\eta w_j} = \xi.
\end{equation}
Next, we write down the KKT condition for $w_j$. Let $v_j,u$ be the Lagrangian multipliers for constraints $w_j \geq 0$ and $\sum_{j=1}^{h^{-d_x}} w_j -1=0$, we have
\begin{align}
\frac{\partial V(\eta,\bm{w})}{\partial w_j}-v_j+u&=0, \label{equ:lasso_global_Lag2}\\
v_j w_j&=0, \label{equ:lasso_global_Lag}\\
v_j &\geq 0, \quad \forall j \in \{1,2,\ldots,h^{-d_x}\} \\
\sum_{j=1}^{h^{-d_x}} w_j&=1.
\end{align}
From \eqref{equ:lasso_global_Lag}, we know either $v_j=0$ or $w_j=0$ for all $j$.
Define a set $O$ including all the subscript $j$ satisfying $w_j>0$ and define its cardinality as $m$,
\begin{equation}
O\coloneqq\{j:v_j=0,w_j > 0\}, \ \text{and} \ m\coloneqq|O|.
\end{equation}
For $j \in O$, plugging $v_j$ into \eqref{equ:lasso_global_Lag2}, we have that
\begin{equation}\label{equ:first-order-eta}
-u=\frac{\partial V(\eta,\bm{w})}{\partial w_j}=\eta e^{\eta w_j} p_j .
\end{equation}
It is easy to see that
\begin{equation}\label{equ:lasso_global_1}
e^{\eta w_1} p_1=e^{\eta w_2} p_2=\ldots=e^{\eta w_{m}} p_{m}=-\frac{u}{\eta }.
\end{equation}
For $j \notin O$, we have $w_j=0$ so
\begin{equation}\label{equ:lasso_global_2}
\sum_{j \in O} w_j=\sum_{j=1}^{h^{-d_x}} w_j= 1.
\end{equation}
Therefore, plugging \eqref{equ:lasso_global_1}, \eqref{equ:lasso_global_2} into \eqref{equ:lasso_global_eta}, we obtain
\begin{equation}\label{equ:lasso_global_3}
e^{\eta w_j} p_j=\xi, \quad \forall j \in O.
\end{equation}
Because $\eta> 0$ and $w_j> 0$, we have $e^{\eta w_j} >1$ and thus
\begin{equation*}
p_j <\xi, \quad \forall j \in O.
\end{equation*}
Then, taking natural logarithm of both sides of \eqref{equ:lasso_global_3}, we have
\begin{equation}\label{equ:opt-wj}
w_j=\left(\log \xi- \log p_j\right)/\eta, \quad \forall j \in O.
\end{equation}
Plugging it into \eqref{equ:lasso_global_2} and \eqref{equ:lasso_global_1}, we get
\begin{equation}\label{equ:opt-eta}
\eta=\sum_{j \in O} \left(\log \xi- \log p_j\right), \quad \forall j \in O,
\end{equation}
and
\begin{equation}\label{equ:lasso_global_4}
u=-\eta \xi.
\end{equation}
For $j \notin O$, we have $v_j \ge 0, w_j=0$.
Thus, plugging \eqref{equ:lasso_global_4} and \eqref{equ:first-order-eta} into \eqref{equ:lasso_global_Lag2}, we have
\begin{equation*}
v_j=\frac{\partial V(\eta,\bm{w})}{\partial w_j}+u=\eta  p_j+u= \eta  (p_j-\xi).
\end{equation*}
As $v_j \ge 0, V(\eta,\bm{w})>0$ and $\eta>0$, we have
\begin{equation*}
p_j \ge  \xi, \quad \forall j \notin O.
\end{equation*}
Plugging~\eqref{equ:opt-wj} and~\eqref{equ:opt-eta}  into~\eqref{equ:global_obj}, we get a closed-form solution for $V(\eta^*,\bm{w}^*)$ if $\eta^*>0$:
\begin{equation}\label{equ:opt-value-V}
V(\eta^*,\bm{w}^*)=\sum_{j=1}^{h^{-d_x}}\left(\xi-\xi \log \xi-p_j+\xi \log p_j\right) \mathbb{I}\left(p_j < \xi\right)
\end{equation}
Therefore, we have prove the KKT condition admits a unique solution. 

We define a function
\begin{equation}\label{equ:def-concave-H}
H(p)\coloneqq\xi \log p-p,
\end{equation}
and
\begin{equation*}
V(\eta^*,\bm{w}^*)=\sum_{j=1}^{h^{-d_x}}\left(H(p_j)-H(\xi)\right)\mathbb{I} (p_j < \xi).
\end{equation*}
Note that $H(\cdot)$ is a concave function, attaining its maximum at $\xi$. Thus, we have $H(p_j)-H(\xi) \leq 0$ and $V(\eta^*,\bm{w}^*) \leq 0=V(0,\bm{w})$. Also, we have $V(\infty,\bm{w})=\infty$ since there exists at least a $w_j \geq h^{d_x}$ such that $e^{\eta w_j}=\infty$.

{\RV Since we have prove that there exists a global minimum for problem \eqref{equ:global_obj} when $\eta$ is finite. Then the unique solution by KKT must be a local minimum instead of a local maximum. Otherwise, if it's a local maximum, then there's no local minimum or global minimum for problem \eqref{equ:global_obj}. Also, since the local minimum solved by KKT is unique, it must be the global minimum for problem \eqref{equ:global_obj}.
}

Combining the above arguments, we have proved that $V(\eta^*,\bm{w}^*)$ is the global optimum for problem \eqref{equ:global_obj}.

Finally, we give a summary for the optimal solution $\eta^*, \bm{w}^*$ of the optimization problem \eqref{equ:global_obj}:
\begin{enumerate}
	\item $\eta^*=\sum_{j=1}^{h^{-d_x}} \left(\log \xi- \log p_j\right) \mathbb{I}\left(p_j < \xi\right)$.
	\item If $p_j < \xi$, then $w_i^*=(\log \xi- \log p_j)/\eta^*$.
	\item If $p_j \ge \xi$, then $w_i^*=0$.
	\item The optimal value $V(\eta^*,\bm{w}^*)=\sum_{j=1}^{h^{-d_x}}\left(\xi-\xi \log \xi-p_j+\xi \log p_j\right) \mathbb{I}\left(p_j < \xi\right)$.
\end{enumerate}

\subsection{Proof of Lemma~\ref{lem:global_max}}\label{proof:lem:global_max}

	Recalling the definition of $H(\cdot)$ in~\eqref{equ:def-concave-H},
	the objective function of \eqref{equ:lasso_global_max} can be rewritten as
	\begin{equation}\label{equ:lasso_global_new}
	V(\bm{n})=\sum_{j=1}^{h^{-d_x}}\left(H(p_j)-H(\xi)\right)\mathbb{I} (p_j < \xi).
	\end{equation}
	Note that $H(\cdot)$ is a negative and concave function, attaining its maximum at $\xi$. Moreover, $H(p_j)$ increases with $p_j$ when $p_j < \xi$. Since $p_j$ is a monotone decreasing function of $n_j$, there exists a threshold
	\begin{equation}\label{equ:lasso_global_thres}
	\underline{n}\coloneqq \max\{n:b_0 \exp(-b_1n h^4) \ge \xi\},
	\end{equation}
	such that $H\left(p_j(n_j)\right)$ (denoted as $H(n_j)$ for simplicity) decreases with $n_j$ when $n_j > \underline{n}$.
	In particular, we have $n$ budgets and $h^{-d_x}$ bins. We divide all the bins into two groups: active bins $A\coloneqq\{j: \mathbb{I} (p_j < \xi)\}$ and non-active bins $A^c\coloneqq\{j: \mathbb{I}(p_j \ge \xi)\}$. For active bins, $H(n_j)$ decreases as more budgets allocated to the bin. For non-active bins, they only consume budgets but have no contribution to the objective function \eqref{equ:lasso_global_new}. To maximize $V(\bm{n})$, the non-active bins should consume as much budgets as possible. So their optimal budgets should equal to the threshold that $n_j^*=\underline{n}$. Thus, if $n \leq \underline{n} h^{-{d_x}}$ (equivalent to $n \leq \log(2 b_0)/(b_1 h^{d_x+4})$), then all the bins are non-active bins and $n_j^*=n h^{d_x}$, thus we have $V(\bm{n})=1$. If $n > \underline{n} h^{-d_x}$, then there must exist active bins.
	We assume the cardinality for active bins is $m\coloneqq|A|$ and their indices are from $1$ to $m$.
	Then, we can fully separate the budgets for active and non-active bins, and reformulate \eqref{equ:lasso_global_max} as
	\begin{equation}\label{equ:lasso_global_max2}
	\begin{aligned}
	\max_{\bm{n}} \quad & V(\bm{n})=\sum_{j=1}^{m}  H(p_j)-H(\xi) \\
	\mbox{s.t.}\quad &  p_j = b_0 \exp\left(-b_1 n_j h^{4}\right) \\
	& n_j \geq \underline{n}\\
	& \sum_{j=1}^{m} n_j  =n-\underline{n}(h^{-d_x}-m)\\
	& n_j \in N^{+}, \quad \forall j \in \{1,2,\ldots,m\}.
	\end{aligned}
	\end{equation}
	Relaxing $n_j$ to $\mathbb{R^+}$, it's a concave and continuous optimization problem.
	By the KKT condition, let $v_j,u$ be the Lagrangian multipliers for $\underline{n}-n_j \leq 0$ and $\sum_{j=1}^{m} n_j  =n-\underline{n}(h^{-d_x}-m)$, we have
	\begin{align}
	\dfrac{\partial V(\bm{n})}{\partial n_j}+v_j+u&=0 \label{equ:first-order-V-n}\\
	v_j (\underline{n}-n_j)&=0 \label{equ:n-complementary} \\
	v_j &\geq 0 \notag\\
	\sum_{j=1}^{m} n_j +\underline{n}(h^{-d_x}-m)  &=n. \label{equ:lasso_global_8}
	\end{align}
	From \eqref{equ:n-complementary}, either $v_j=0$ or $n_j=\underline{n}$ for all $j$.
	Define a set
	\begin{equation*}
	O\coloneqq\{j:v_j=0,\ n_j > \underline{n} \},
	\end{equation*}
	for $j \in O$.
	Plugging $v_j=0$ into \eqref{equ:first-order-V-n}, we have
	\begin{equation}\label{equ:lasso_global_6}
	u=-\dfrac{\partial V(\bm{n})}{\partial n_j}=b_1 h^4 (\xi-p_j).
	\end{equation}
	By the definition of $\underline{n}$ in \eqref{equ:lasso_global_thres}, $n_j > \underline{n}$ implies $p_j < \xi$. Then $u > 0$.
	For $k \in O^c$, we have $v_k \ge 0$, $n_k=\underline{n}$ and
	\begin{equation}\label{equ:lasso_global_7}
	u=-\dfrac{\partial V(\bm{n})}{\partial n_k}-v_k=b_1 h^4 (\xi-p_k)-v_k.
	\end{equation}
	In fact, \eqref{equ:lasso_global_6} and \eqref{equ:lasso_global_7} cannot hold simultaneously.
	Recalling that for $j \in O$ and $k \in O^c$, $n_j > \underline{n}=n_k$.
	Thus we have $p_j < p_k$ because $p_j$ is decreasing in $n_j$.
	By \eqref{equ:lasso_global_6} and \eqref{equ:lasso_global_7}, we have
	\begin{equation*}
	u=b_1 h^4 (\xi-p_j)>b_1 h^4 (\xi-p_k) \ge b_1 h^4 (\xi-p_k)-v_k=u.
	\end{equation*}
	That's to say, either $O$ or $O^c$ is empty. Since we have supposed $n > \underline{n} h^{-d_x}$, there's at least in one bin $n_j >\underline{n}$. So $O^c$ is empty and \eqref{equ:lasso_global_6} is satisfied for all $j \in {1,2,\ldots,m}$, further implying $p_1=p_2=\ldots=p_m$. As $p_j$ is a strictly decreasing function of $n_j$, by \eqref{equ:lasso_global_8}, we get the optimal solution
	\begin{equation}\label{equ:lasso_global_n}
	n_1^*=n_2^*=\ldots=n_m^*=\left(n-\underline{n}(h^{-d_x}-m)\right)/m.
	\end{equation}
	Moreover, the optimal value in \eqref{equ:lasso_global_max2} is
	\begin{align}
	V(\bm{n})
	&=m(\xi \log p_j-p_j-\xi \log \xi+\xi) \notag\\
	& \leq m \xi (\log p_j-\log \xi+1) \notag\\
	&= \xi \left(-b_1 h^4 n+b_1 \underline{n} h^{4-d_x} +\left(\log b_0-\log \xi+1-b_1 h^4 \underline{n}\right) m\right), \label{equ:lasso_global_max3}
	\end{align}
	where the last equality follows from \eqref{equ:lasso_global_n} and $p_j=b_0 \exp(-b_1 n_j h^4)$.
	By the definition of $\underline{n}$ in  \eqref{equ:lasso_global_thres}, we have
	\begin{equation*}
	\log b_0-\log \xi+1-b_1 h^4 \underline{n} >0,
	\end{equation*}
	which implies that the term in \eqref{equ:lasso_global_max3} will increase as $m$. Therefore, when $m=h^{-d_x}$, the term in \eqref{equ:lasso_global_max3} attains its maximum, which also gives an upper bound for the optimal $V(\bm{n}^*)$ in \eqref{equ:lasso_global_max},
	\begin{equation*}
	V(\bm{n}^*)
	\leq \xi\left(h^{-d_x} (1+\log b_0-\log \xi)-b_1 n h^{4}\right).
	\end{equation*}

\section{Proofs for the Regret Bound}
\subsection{Proof of Proposition~\ref{prop:rate_selection}}\label{proof:rate_selection}
	Supposing the dimension of decision space is $d_y$, we prove the stronger version stated in Remark \ref{rem:stronger-dy}. Recall that the total regret in $T$ periods can be upper bounded by
	\begin{equation*}
	R(T) \leq 2n\max_{\bm x\in \XX,y\in \YY}|f(\bm x, y)|+ \PR (\hat{J}=J)  R_2(T-n)+2 \max_{\bm x\in \XX, y\in \YY}|f(\bm x,  y)| \PR (\hat{J} \neq J) (T-n),
	\end{equation*}
	where $n \leq T^{(d^*_x+d_y+1)/(d^*_x+d_y+2)}$ and $\PR (\hat{J} \neq J) \leq n^{-1/(d^*_x+d_y+1)}$.
	Further relaxing the right-hand side, we have
	\begin{align*}
	R(T)
	&\leq 2n\max_{\bm x\in \XX, y\in \YY}|f(\bm x, y)|\!+\!\PR (\hat{J}=J)  R_2(T-n)\!+\! 2 \max_{\bm x\in \XX, y\in \YY}|f(\bm x, y)| n^{-1/(d^*_x+d_y+1)} (T-n)  \\
	&=2n\max_{\bm x\in \XX, y\in \YY}|f(\bm x, y)|+  R_2(T)+2 \max_{\bm x\in \XX, y\in \YY}|f(\bm x,  y)| n^{-1/(d^*_x+d_y+1)} T \\
	&=O(T^{1-1/(d^*_x+d_y+2)})+R_2(T)\\
	&=O\left(T^{1-1/(d_x^*+d_y+2)} \log(T) \right).
	\end{align*}
	The first equality follows from $\PR (\hat{J}=J) \leq 1$ and $T-n \leq T$, the second equality follows from $\PR (\hat{J} \neq J) \leq n^{-1/(d^*_x+d_y+1)} \leq \left(T^{(d^*_x+d_y+1)/(d^*_x+d_y+2)}\right)^{-1/(d^*_x+d_y+1)}=O(T^{-1/(d^*_x+d_y+2)})$ and the last equality is supported by \eqref{eq:learning-regret}.

\subsection{Proof of Theorem~\ref{theo:cov_reg}}\label{proof:theorem1}
Recall that the total regret in $T$ periods can be upper bounded by
\begin{equation*}
R(T) \leq 2n\max_{\bm x \in \XX,y \in \YY}|f(\bm x, y)|+ \PR (\hat{J}=J)  R_2(T-n)+2 \max_{\bm x\in \XX, y\in \YY}|f(\bm x,  y)| \PR (\hat{J} \neq J) (T-n).
\end{equation*}
where $n=T^{2/3}$. Since $T \geq (b_3)^{-3(d_x+2)}$ and  $h=T^{-1/(3d_x+6)}$, we have $h \leq b_3$. Since $T \geq \left((3+\log 2+\log b_0)/b_1\right)^{3(1+2/d_x)}$, we have $n \geq \log(2 b_0)/(b_1 h^{d_x+4})$. Thus, applying Proposition~\ref{prop:global}, we have
\begin{align*}
\PR(\hat{J} \neq J)
&\leq d_x \exp\left\{\frac{1}{2} \left( h^{-d_x}(1+\log b_0+\log2)-b_1 n h^4\right)\right\}\\
& \overset{(a)}{=}d_x \exp\left\{\frac{1}{2} \left( n^{1/(2+4/d_x)}(1+\log b_0+\log2)-b_1 n^{1/(1+2/d_x)}\right)\right\}\\
& =d_x \exp\left\{\frac{1}{2} n^{1/(2+4/d_x)} \left( (1+\log b_0+\log2)-b_1 n^{1/(2+4/d_x)}\right) \right\}\\
& \overset{(b)}{ = } d_x \exp\left\{\frac{1}{2} T^{1/(3+6/d_x)} \left( (1+\log b_0+\log2)-b_1 T^{1/(3+6/d_x)}\right) \right\}\\
& \overset{(c)}{ \leq }d_x \exp\left\{ \frac{1}{2} \log T \left( (1+\log b_0+\log2)-b_1 T^{1/(3+6/d_x)}\right) \right\}\\
& \overset{(d)}{ \leq }d_x \exp\left\{-\frac{1}{2} \log T\right\}\\
& = d_x/\sqrt{T},
\end{align*}
where $(a)$ follows from $h=n^{-1/(2 d_x+4)}$, $(b)$ follows from $n=T^{2/3}$, $(c)$ follows from $T \geq (\log T)^{3(1+2/d_x)}$, $(d)$ follows from $T \geq \left((3+\log 2+\log b_0)/b_1\right)^{3(1+2/d_x)}$.

Further relaxing the right-hand side, we have
\begin{align*}
R(T)
&\leq 2T^{2/3}\max_{\bm x \in \XX,y \in \YY}|f(\bm x, y)|+ R_2(T-n)+ 2 \max_{\bm x\in \XX, y\in \YY}|f(\bm x,  y)| T d_x/\sqrt{T} \\
&\leq O\left(T^{2/3}\right)+ R_2(T-n)+ O(\sqrt{T})\\
&=O(R_2(T)).
\end{align*}

\subsection{Proof of Theorem~\ref{cor:cov_reg_n}}

	Recall that the total regret in $T$ periods can be upper bounded by
	\begin{equation*}
	R(T) \leq 2n\max_{\bm x \in \XX,y \in \YY}|f(\bm x, y)|+ \PR (\hat{J}=J)  R_2(T-n)+2 \max_{\bm x\in \XX, y\in \YY}|f(\bm x,  y)| \PR (\hat{J} \neq J) (T-n).
	\end{equation*}
	where $n=(\log(T))^{(2+4/d_x)}$. Since $T \geq \exp\{(b_3)^{-d_x}\}$ and  $h=(\log(T))^{-1/d_x}$, we have $h \leq b_3$. Since $T \geq \exp\{(3+\log 2+\log b_0)/b_1\}$, we have $n \geq \log(2 b_0)/(b_1 h^{d_x+4})$. Thus, applying Proposition~\ref{prop:global}, we have
	\begin{equation*}
	\PR(\hat{J} \neq J) \leq d_x \exp\left\{\frac{1}{2} \left( h^{-d_x}(1+\log b_0+\log2)-b_1 n h^4\right)\right\}.
	\end{equation*}
	Further relaxing the right-hand side, we have
	\begin{align*}
	R(T)
	&\leq 2(\log T)^{(2+4/d_x)}\max_{\bm x \in \XX,y \in \YY}|f(\bm x, y)|+ R_2(T-n) \\
	& \quad+ 2 \max_{\bm x\in \XX, y\in \YY}|f(\bm x,  y)| d_x \exp\left\{\frac{1}{2} \left( h^{-d_x}(1+\log b_0+\log2)-b_1 n h^4\right)\right\} T\\
	&\leq 2(\log T)^6\max_{\bm x \in \XX,y \in \YY}|f(\bm x, y)| + R_2(T-n)+ 2 \max_{\bm x\in \XX, y\in \YY}|f(\bm x,  y)| d_x \exp(-\log T) T\\
	&= 2(\log T)^6\max_{\bm x \in \XX,y \in \YY}|f(\bm x, y)| + R_2(T-n)+ 2 \max_{\bm x\in \XX, y\in \YY}|f(\bm x,  y)| d_x\\
	&=O(R_2(T)).
	\end{align*}
	The second inequality follows from $d_x \geq 1$ and $T \geq \exp\left\{(3+\log b_0+\log 2)/b_1\right\}$.

\section{Proofs for Local Relevance}
\subsection{Proof of Proposition~\ref{prop:inf_bins}}\label{Proof:point-prob}
	Recall that in the proof of Lemma \ref{lem:existence-C}, the hypercube $\mathcal{H}_i$ with side length $\overline{h}=C/L$ and centred at $\bm{x}_{(i)}$ satisfies \eqref{equ:hypercube-H-C}.
	We will show that $Q_i(C)$ covers at least $(1/3)^{d_x}$ proportion of $\mathcal{H}_i$, if choosing $h \leq \overline{h}/3$. Note that $\mathcal{H}_i$ is covered by $Q_i(C)$ and bins intersected with the boundary of $\mathcal{H}_i$. We consider the worst case that as more areas covered by the intersected bins as possible. When all the boundaries of the intersected bins exactly coincide with the boundary of $\mathcal{H}_i$, the intersected bins take up the most proportion of $\mathcal{H}_i$. In this case, $2/3$ proportion of each side length is covered by the intersected bins, and $(1/3)^{d_x}$ proportion of $\mathcal{H}_i$ is covered by the bins in $Q_i(C)$. Then, by Assumption \ref{ass:cov_des2}, the probability density has a lower bound $\mu_{m}$ and
	\begin{equation*}
	\PR(\bm{X} \in Q_i(C)) \geq \PR(\bm{X} \in \mathcal{H}_i) \geq (1/3)^{d_x}\mu_{m} (\overline{h}/3)^{d_x}= \mu_{m} \left(\frac{C}{3L}\right)^{d_x}.
	\end{equation*}
	Hence, we complete the proof of Proposition \ref{prop:inf_bins}.

\subsection{Proof of Proposition~\ref{prop:small-xi}}\label{proof:small-xi}

	\textbf{\textit{Step one:}} In the first step, we consider the case that the true indicator $J^{(i)}=0$, that is, $x_i$ is redundant. We show an upper bound for the misidentification probability:
	\begin{align}
	& \quad \PR(\hat{J}^{(i)} \geq \xi | J^{(i)}=0) \notag\\
	&=\PR (e^{\eta \hat{J}^{(i)}} \geq e^{\eta \xi} | J^{(i)}=0) \notag\\
	&\leq e^{-\eta \xi} \mathbb{E} \left[e^{\eta \hat{J}^{(i)}} \big| J^{(i)}=0\right] & (\text{by Markov's inequality} )\notag\\
	&=e^{-\eta \xi} \prod_{j=1}^{h^{-d_x}} \mathbb{E} \left[e^{\eta w_j \hat{J}_j^{(i)}} \big| J^{(i)}=0\right] &(\text{by the definition of } \hat{J}^{(i)} ~\eqref{eq:weighted-vote}) \notag\\
	&=e^{-\eta \xi} \prod_{j=1}^{h^{-d_x}} \left(1+(e^{\eta w_j}-1)p_j\right) & (\text{by } \hat{J}_j^{(i)}\sim~ Bernoulli(p_j)) \notag\\
	&=\exp\left\{-\eta \xi +\sum_{j=1}^{h^{-d_x}} \log \left(1+(e^{\eta w_j}-1)p_j\right)\right\} \label{equ:lasso_global2_prob}
	\end{align}
	{ \RV \Copy{cop:local-bound}{Note that the relaxation used in \eqref{equ:lasso_global_prob} under the global relevance assumption, $1+p_j \leq e^{p_j}$, is proper when $p_j$ is closed to zero.
	But under the local relevance assumption required in this proposition, some $p_j$s are not close to zero.
	Such a relaxation is not tight any more.
	In fact, we find that the analysis in Lemma~\ref{lem:global_max} doesn't go through here.
	Thus, we provide a new proof in \eqref{equ:lasso_global2_prob}: it has a $\log$ term in the exponential function and is more challenging to analyze than \eqref{equ:lasso_global_prob}.
	}}

	Since \eqref{equ:lasso_global2_prob} holds for arbitrary non-negative $\eta$ and $w_j$, we need to find $\eta, w_j$ to minimize the probability:
	\begin{equation}\label{equ:global2_obj}
	\begin{aligned}
	\min_{\eta,\bm{w}} \quad &  V(\eta,\bm{w})=-\eta \xi +\sum_{j=1}^{h^{-d_x}} \log \left(1+(e^{\eta w_j}-1)p_j\right) \\
	\mbox{s.t.}\quad & w_j \geq 0, \ \forall j \in \{1,2,\ldots,h^{-d_x}\}, \\
	& \eta \ge 0,\\
	& \sum_{j=1}^{h^{-d_x}} w_j=1.\\
	\end{aligned}
	\end{equation}

	Similar to the proof of Lemma~\ref{lem:global_min}, we apply the KKT optimality condition.
	If $\eta^*=0$, then $V(0,\bm{w})=0$ for any $\bm{w}$, which is clearly not optimal. So $\eta^*>0$ and the first-order condition holds.
	That implies the optimal $\eta$ solving
	\begin{equation}\label{equ:lasso_global2_0}
	0=\frac{\partial V(\eta,\bm{w})}{\partial \eta}=-\xi + \sum_{j=1}^{h^{-d_x}} \dfrac{w_j p_j e^{\eta w_j}}{1+(e^{\eta w_j}-1)p_j}
	\end{equation}
	Plugging it into \eqref{equ:lasso_global2_0}, we have
	\begin{equation}\label{equ:lasso_global2_4}
	\xi= -\frac{u}{\eta} \sum_{j \in O} w_j=-\frac{u}{\eta}.
	\end{equation}
	By \eqref{equ:KKT-eta-w} and \eqref{equ:lasso_global2_4}, we have
	\begin{equation}\label{equ:eta-w-equal}
	\eta w_1=\eta w_2=\ldots=\eta w_m=\log \xi +\log (1-p_j)- \log p_j- \log (1-\xi).
	\end{equation}
	Since $\eta w_1 >0$, we have $p_j < \xi$.
	Therefore, plugging \eqref{equ:eta-w-equal}, into \eqref{equ:weight-sum-to-1}, we obtain
	\begin{equation}\label{equ:eta-sol-mid}
	\eta= \sum_{j \in O} (\log \xi+\log (1-p_j)- \log p_j-\log(1-\xi)),
	\end{equation}
	and
	\begin{equation}\label{equ:eta-sol-mid_1}
	w_j=\frac{1}{\eta}\left(\log \xi+\log (1-p_j)- \log p_j-\log(1-\xi)\right), \quad \forall j \in O.
	\end{equation}
	For $j \notin O$, we have $v_j \ge 0, w_j=0$. Then, plugging \eqref{equ:eta-sol-mid}, \eqref{equ:lasso_global2_4}, \eqref{equ:first-order-eta2} into \eqref{equ:lasso_global2_Lag2}, we have
	\begin{equation*}
	v_j= \eta (p_j-\xi),
	\end{equation*}
	and
	\begin{equation*}
	p_j \ge \xi, \quad \forall j \notin O.
	\end{equation*}
	Plugging \eqref{equ:eta-sol-mid} and \eqref{equ:eta-sol-mid_1} into \eqref{equ:global2_obj}, we have
	\begin{equation}\label{equ:lasso_global2_opt}
	V(\eta^*,\bm{w}^*)=\sum_{j=1}^{h^{-d_x}}\left(\xi \log p_j+(1-\xi) \log(1-p_j) -\xi \log \xi-(1-\xi) \log(1-\xi)\right) \mathbb{I}\left(p_j < \xi\right).
	\end{equation}
	Therefore, we have prove the KKT condition admits a unique solution, which must be the global optimum for problem \eqref{equ:global2_obj}.

	We give a summary for the optimal solution $\eta^*, \bm{w}^*$ of the optimization problem \eqref{equ:global2_obj}:
	\begin{enumerate}
		\item $\eta^*=\sum_{j=1}^{h^{-d_x}} \left(\log \xi+\log(1-p_j)- \log p_j-\log(1-\xi)\right) \mathbb{I}\left(p_j < \xi\right)$.
		\item If $p_j < \xi$, then $w_j^*=(\log \xi+\log(1-p_j)- \log p_j-\log(1-\xi))/\eta^*$.
		\item If $p_j \ge \xi$, then $w_j^*=0$.
		\item The optimal value $V(\eta^*,\bm{w}^*)$ shows in \eqref{equ:lasso_global2_opt}.
	\end{enumerate}
	The optimal $V$ of \eqref{equ:lasso_global2_opt} depends on $p_j$, which is the probability bound derived in Proposition~\ref{prop:subset}. The condition of Proposition~\ref{prop:subset} holds since $h \leq b_3/2 \leq b_3$ and $\lambda=b_2 h^2$.  Plugging $p_j$ into \eqref{equ:lasso_global2_opt} and because of $\log (1-p_j)<0$, we have
	\begin{align}
	V(\eta^*,\bm{w}^*)
	&\leq \sum_{j=1}^{h^{-d_x}}\left(\xi \log p_j-\xi \log \xi-(1-\xi) \log(1-\xi)\right) \mathbb{I}\left(p_j < \xi\right) \notag\\
	&\leq \sum_{j=1}^{h^{-d_x}}\left(-\xi b_1 n_j h^4+\xi \log b_0-\xi \log \xi-(1-\xi) \log(1-\xi)\right) \mathbb{I}\left(p_j < \xi\right) \notag \\
	&\leq \left(\xi \log b_0-\xi \log \xi-(1-\xi) \log(1-\xi)\right) h^{-d_x} -\xi b_1 h^4 \cdot \left( \sum_{j=1}^{h^{-d_x}} n_j \mathbb{I}\left(p_j < \xi\right) \right). \label{equ:lasso_global2_last}
	\end{align}
	The first inequality follows from $\log (1-p_j)<0$; the second follows from $p_j=b_0 \exp(-b_1 n_j h^4)$.
	Since $p_j$ is a monotone decreasing function of $n_j$, there exists a threshold
	\begin{equation}\label{equ:lasso_global2_thres}
	\underline{n}\coloneqq \max\{n:b_0 \exp(-b_1 n h^4) \ge \xi\},
	\end{equation}
	such that $p_j < \xi$ for $n_j > \underline{n}$. By \eqref{equ:lasso_global2_thres}, we have
	\begin{equation}\label{equ:underline-n-ineq}
	b_0 \exp(-b_1 \underline{n} h^4) > \xi \Longrightarrow b_1 \underline{n} h^4 \le \log b_0 - \log \xi
	\end{equation}
	So we get a lower bound for the last term in \eqref{equ:lasso_global2_last},
	\begin{equation}\label{equ:lasso_global2_O}
	\sum_{j=1}^{h^{-d_x}} n_j \mathbb{I}\left(p_j < \xi\right) \geq n-h^{-d_x} \underline{n} \geq n- \frac{h^{-d_x}(\log b_0 - \log \xi)}{b_1 h^4}.
	\end{equation}
	Therefore, plugging \eqref{equ:lasso_global2_O} into \eqref{equ:lasso_global2_last}, we have
	\begin{align*}
	V(\eta^*,\bm{w}^*)
	&\leq \left(\xi \log b_0-\xi \log \xi-(1-\xi) \log(1-\xi)\right) h^{-d_x} -\xi b_1 h^4 \left( n-h^{-d_x} \underline{n}\right)\\
	&\leq \left(\xi \log b_0-\xi \log \xi-(1-\xi) \log(1-\xi)\right) h^{-d_x} +\xi (\log b_0- \log \xi)h^{-d_x}-\xi b_1 h^4 n \\
	&=\left(2\xi \log b_0-2\xi \log \xi-(1-\xi) \log(1-\xi)\right) h^{-d_x}-\xi b_1 h^4 n.
	\end{align*}
	The second inequality holds by \eqref{equ:underline-n-ineq}.
	Recalling the tail probability in \eqref{equ:lasso_global2_prob}, we have
	\begin{equation}\label{equ:tail-true-0}
	\PR(\hat{J}^{(i)} \geq \xi | J^{(i)}=0) \leq \exp\left\{\left(2\xi \log b_0-2\xi \log \xi-(1-\xi) \log(1-\xi)\right) h^{-d_x}-\xi b_1 h^4 n \right\}.
	\end{equation}
	So far, we show a tail probability upper bound for the variable $x_i$ satisfying $J^{(i)}=0$.
	Note that the upper bound \eqref{equ:tail-true-0} is looser than the upper bound \eqref{equ:global_max} in Lemma \ref{lem:global_max}.

	\textit{\textbf{Step two:}}
	Next we consider the case when $J^{(i)}=1$.
	We start with the following bound
	\begin{align}
	\PR(\hat{J}^{(i)} \leq \xi | J^{(i)}=1)
	&=\PR(1-\hat{J}^{(i)} \geq 1-\xi | J^{(i)}=1) \notag\\
	&\leq e^{-\eta (1-\xi)} \mathbb{E} \left[\exp\left(\eta \left(1-\hat{J}^{(i)}\right)\right) \big| J^{(i)}=1\right] \notag\\
	&=e^{-\eta (1-\xi)} \prod_{j=1}^{h^{-d_x}} \mathbb{E} \left[\exp \left(\eta w_j \left(1-\hat{J}_j^{(i)}\right)\right) \big| J^{(i)}=1\right] \notag\\
	&=\exp\left\{-\eta (1-\xi) +\sum_{j=1}^{h^{-d_x}} \log \left(1+(e^{\eta w_j}-1)p_j\right)\right\}. \label{equ:lasso_global2_prob2}
	\end{align}
	Notice that \eqref{equ:lasso_global2_prob2} is the same as \eqref{equ:lasso_global2_prob} except that $\xi$ is replaced by $1-\xi$.
	Thus, replacing $\xi$ by $1-\xi$ in \eqref{equ:eta-sol-mid} and \eqref{equ:eta-sol-mid_1}, we get the optimal solution for \eqref{equ:lasso_global2_prob2},
	\begin{equation}\label{equ:opt-eta-flip}
	\eta^*=\sum_{j=1}^{h^{-d_x}} \left(\log (1-\xi)+\log(1-p_j)- \log p_j-\log \xi \right) \mathbb{I}\left(p_j < 1-\xi\right),
	\end{equation}
	and
	\begin{align}
	w_j^*
	&=\left(\log (1-\xi)+\log(1-p_j)- \log p_j-\log \xi\right)/\eta^* & \text{for} \ p_j < 1-\xi, \label{equ:opt-w1-flip}\\
	w_j^*
	&=0 & \text{for} \ p_j \ge 1-\xi. \label{equ:opt-w0-flip}
	\end{align}
	Plugging them into \eqref{equ:lasso_global2_prob2}, we have
	\begin{equation}\label{equ:opt-V-star}
	V(\eta^*,\bm{w}^*)=
	\exp\left\{-\eta^* (1-\xi) +\sum_{j=1}^{h^{-d_x}} \log \left(1+(e^{\eta^* w_j^*}-1)p_j\right)\mathbb{I}\left(p_j < 1-\xi\right)\right\}
	\end{equation}
	Note that on the event $p_j < 1-\xi$,
	\begin{align*}
	\log \left(1+(e^{\eta^* w_j^*}-1)p_j\right)
	&= \log\left(1+\left(\exp\left(\log (1-\xi)+\log(1-p_j)- \log p_j-\log(\xi)\right)-1\right)p_j\right)\\
	&=\log \left(1+\left(\frac{(1-\xi)(1-p_j)}{p_j \xi}-1\right)p_j\right)\\
	&=\log(1-p_j)-\log \xi.
	\end{align*}
	Moreover, for the uninformative bins $j \in Q^c$ (with a slight abuse, we omit $C$ and $i$ in $Q_i(C)$ and use $j$ to represent $B_j$), we simply use an upper bound $p_j\le 1$ and\\
	$\log \left(1+(e^{\eta^* w_j^*}-1)p_j\right)
	\le \eta^* w_j^*$.
	Plugging them to \eqref{equ:opt-V-star}, we have
	\begin{equation*}
	\log V(\eta^*,\bm{w}^*)
	\!\le\! -\eta^*(1-\xi)\!+\!\sum_{j=1}^{h^{-d_x}} \left(\eta^* w_j^* \mathbb{I}(j \in Q^c)\!+\!\left(\log(1-p_j)\!-\!\log\xi\right) \mathbb{I}(j \in Q)\right) \mathbb{I}\left(p_j < 1\!-\!\xi\right).
	\end{equation*}
	Like the previous argument, we define
	\begin{equation*}
	O \coloneqq \left\{j: p_j < 1-\xi\right\}.
	\end{equation*}
	Plugging in the form of $\eta^*$ and $w_j^*$ from \eqref{equ:opt-eta-flip} and \eqref{equ:opt-w1-flip}, we have
	\begin{align} \label{equ:plug-uninfor-V}
	&\quad \log  V(\eta^*,\bm{w}^*)\notag\\
	&\le -\eta^*(1-\xi)+\sum_{j=1}^{h^{-d_x}} \left(\eta^* w_j^* \mathbb{I}(j \in Q^c)+\left(\log(1-p_j)-\log\xi\right) \mathbb{I}(j \in Q)\right) \mathbb{I}\left(j \in O\right) \notag\\
	&=\sum_{j=1}^{h^{-d_x}} -(1-\xi) \left(\log (1-\xi)+\log(1-p_j)- \log p_j-\log \xi \right) \mathbb{I}(j \in O)\notag\\
	& \quad \!+\! \left(\log (1\!-\!\xi)+\log(\!1-\!p_j)\!-\! \log p_j\!-\!\log\xi\right)\mathbb{I}(j \!\in\! O \cap Q^c) \!+\! \left(\log(1-p_j)\!-\!\log \xi\right) \mathbb{I}(j \!\in\! O \cap Q).
	\end{align}
	Recombining the terms in \eqref{equ:plug-uninfor-V} by~$\log(1-\xi),~\log \xi,~\log(1-p_j)$ and $\log p_j$, we have \eqref{equ:plug-uninfor-V}
	\begin{align*}
	&=\sum_{j=1}^{h^{-d_x}} \left(-(1-\xi) \mathbb{I} (j \in O)+\mathbb{I} (j \in O \cap Q^c)\right) \log(1-\xi)\\
	& \quad +\left((1-\xi) \mathbb{I} (j \in O)-\mathbb{I} (j \in O \cap Q^c)-\mathbb{I} (j \in O \cap Q)\right) \log \xi\\
	& \quad + \left(-(1-\xi) \mathbb{I} (j \in O)+\mathbb{I} (j \in O \cap Q^c)+\mathbb{I} (j \in O \cap Q)\right) \log(1-p_j)\\
	& \quad + \left((1-\xi) \mathbb{I} (j \in  O)-\mathbb{I} (j \in O \cap Q^c)\right) \log p_j.
	\end{align*}
	Further by $\mathbb{I} (j \in O)= \mathbb{I} (j \in  O \cap Q)+\mathbb{I} (j \in  O \cap Q^c)$, \eqref{equ:plug-uninfor-V} is simplified to
	\begin{align}\label{equ:logV-simple}
	&=\sum_{j=1}^{h^{-d_x}} \left(-(1-\xi) \mathbb{I} (j \in O)+\mathbb{I} (j \in O \cap Q^c)\right) \log(1-\xi)- \xi \mathbb{I} (j \in O \cap Q^c) \log \xi \notag\\
	& \quad +\xi \mathbb{I} (j \in O) \log(1-p_j)
	+\left((1-\xi) \mathbb{I} (j \in  O \cap Q)-\xi \mathbb{I} (j \in  O \cap Q^c)\right) \log p_j.
	\end{align}
	Note that $\log \xi \le 0,~\log(1-\xi) \le 0,~\log(1-p_j) \le 0$ and $\mathbb{I} (j \in O \cap Q^c) \leq \mathbb{I} (j \in O)$, we have \eqref{equ:logV-simple}
	\begin{align}\label{label:equ-V-bound}
	& \leq \sum_{j=1}^{h^{-d_x}} \left(-(1-\xi)\log(1-\xi)\mathbb{I} (j \in O)\right)- \xi \mathbb{I} (j \in O \cap Q^c) \log \xi \notag\\
	& \quad +\left((1-\xi) \mathbb{I} (j \in  O \cap Q)-\xi \mathbb{I} (j \in  O \cap Q^c)\right) \log p_j \notag\\
	& \leq \sum_{j=1}^{h^{-d_x}} \mathbb{I}(j \in O)   \left(-\xi \log \xi\!-\!(1\!-\!\xi)\log(\!1-\!\xi)\right)
	\!+\!\left((1\!-\!\xi) \mathbb{I} (j \in  O \cap Q)-\xi \mathbb{I} (j \in  O \cap Q^c)\right) \log p_j \notag\\
	& \leq h^{-d_x} \left(-\xi \log \xi\!-\!(1\!-\!\xi)\log(1-\xi)\right) \!+\!\sum_{j=1}^{h^{-d_x}} \mathbb{I}(j \in O \cap Q)   (1\!\!-\xi)\log p_j\!-\!\mathbb{I}(j \in O \cap Q^c)   \xi \log p_j.
	\end{align}
	The last inequality follows from $\mathbb{I}(j \in O) \le 1$.

	Next, we give an upper bound for \eqref{label:equ-V-bound}.
	Recalling that $Q$ ($Q_i(C)$) is defined as the union of bins in $A_i(C)$, then the probability bound in Proposition~\ref{prop:subset} holds as long as $h \leq b_3/2$ and $\lambda=b_2 h^2$.
	Plugging $p_j=b_0 \exp\left(-b_1 n_j h^4\right)$ into the last two terms,
	we have
	\begin{align}
	& \quad \sum_{j=1}^{h^{-d_x}} \mathbb{I}(j \in O \cap Q)   (1-\xi)\log p_j-\mathbb{I}(j \in O \cap Q^c)   \xi \log p_j \notag \\
	&= \sum_{j=1}^{h^{-d_x}}  \left(\mathbb{I}(j \!\in O \!\cap\! Q) (1\!-\!\xi)\!-\!\mathbb{I}(j \!\in O\! \!\cap\! Q^c) \xi\right)\log b_0 \!-\! \left(\mathbb{I}(j \!\in\! O\! \cap\! Q) (1\!-\!\xi)n_j\!-\!\mathbb{I}(j \!\in\! O \!\cap\! Q^c) \xi n_j\right)b_1 h^4 \notag \\
	& \leq (1-\xi) h^{-d_x} \log b_0- (1-\xi)b_1 h^4 \cdot \sum_{j=1}^{h^{-d_x}} \mathbb{I}(j \in O \cap Q) n_j+ \xi b_1 h^4 \cdot \sum_{j=1}^{h^{-d_x}} \mathbb{I}(j \in Q^c) n_j. \label{equ:lasso_global2_5}
	\end{align}
	The last inequality follows from  $\mathbb{I}(j \in O \cap Q) \leq 1$ and $\log b_0 \ge 0$.

	Notice that $\sum_{j=1}^{h^{-d_x}} \mathbb{I}(j \in Q^c) n_j=\sum_{k=1}^n \mathbb{I}(\bm X_k \in Q^c)$ is the number of covariates falling in $Q^c$, which is a binomial distribution. According to Proposition \ref{prop:inf_bins}, the mean probability $\PR (\bm X \in Q^c) < 1-p_Q$. Then, applying the Hoeffding's inequality for binomial random variable, we have
	\begin{equation}\label{equ:binomial-tail}
	\PR \left(\sum_{k=1}^n \mathbb{I}(\bm X_k \in Q^c)-n(1-p_Q) \geq \frac{1}{3}p_Q n \right) \leq e^{-\frac{2}{9} p_Q^2 n}.
	\end{equation}
	Thus, with probability no less than $1-e^{-\frac{2}{9} p_Q^2 n}$, we have
	\begin{equation}\label{equ:low-Qc}
	\sum_{j=1}^{h^{-d_x}} \mathbb{I}(j \in Q^c) n_j \leq (1-\frac{2}{3} p_Q)n.
	\end{equation}
	Similar to \eqref{equ:lasso_global2_thres} and \eqref{equ:lasso_global2_O}
	we define the threshold
	\begin{equation*}
	\underline{n}\coloneqq \max\{n:b_0 \exp(-b_1 n h^4) \ge 1-\xi\},
	\end{equation*}
	and we have
	\begin{equation}\label{equ:samples-in-O}
	\sum_{j=1}^{h^{-d_x}}  \mathbb{I}(j \in O)n_j \geq n-h^{-d_x} \underline{n} \geq n- \frac{h^{-d_x}}{b_1 h^4} (\log b_0 - \log (1-\xi)).
	\end{equation}
	By \eqref{equ:low-Qc} and \eqref{equ:samples-in-O}, we have
	\begin{align}\label{equ:low-O-cap-Q}
	\sum_{j=1}^{h^{-d_x}} \mathbb{I}(j \in O \cap Q) n_j
	&\geq \sum_{j=1}^{h^{-d_x}} \mathbb{I}(j \in O) n_j - \sum_{j=1}^{h^{-d_x}} \mathbb{I}(j \in Q^c)n_j \notag\\
	&\geq n- h^{-d_x}\underline{n}-(1-\frac{2}{3} p_Q)n\notag\\
	&=\frac{2}{3} p_Q n-h^{-d_x} \underline{n}\notag\\
	&\geq \frac{2}{3} p_Q n-\frac{h^{-d_x}}{b_1 h^4} (\log b_0 - \log (1-\xi)).
	\end{align}
	Plugging \eqref{equ:low-Qc} and \eqref{equ:low-O-cap-Q} into \eqref{equ:lasso_global2_5} also \eqref{equ:plug-uninfor-V}, we have
	\begin{align*}
	&\quad\log V(\eta^*,\bm{w}^*)\\
	& \leq h^{-d_x} \left((1\!-\!\xi) \log b_0\!-\!\xi \log \xi\!-\!(1\!-\!\xi)\log(1\!-\!\xi)\right)\!+\!(1\!-\!\xi) \underline{n} b_1 h^4 h^{-d_x}\!-\!\left(\frac{2}{3} p_Q\!-\!\xi\right)b_1 h^4 n \\
	& \leq h^{-d_x} \left(2(1-\xi) \log b_0- \xi \log \xi-2(1-\xi) \log(1-\xi)\right)-\left(\frac{2}{3} p_Q-\xi\right)b_1 h^4 n.
	\end{align*}
	Plugging it and \eqref{equ:binomial-tail} into \eqref{equ:lasso_global2_prob2}, we have
	\begin{align*}
	& \quad \PR(\hat{J}^{(i)} \leq \xi | J^{(i)}=1) \\
	&\leq \exp\left\{h^{-d_x} \left(2(1-\xi) \log b_0\!-\! \xi \log  \xi\!-\!2(1\!-\!\xi) \log(1-\xi)\right)\!-\!\left(\frac{2}{3} p_Q-\xi\right)b_1 h^4 n\right\} \!+\!\exp\left(-\frac{2}{9} p_Q^2 n\right).
	\end{align*}
	Hence, we complete the proof of Proposition~\ref{prop:small-xi}.

\subsection{\RV Proof of Theorem \ref{theo:local-cov_reg}}
Note that we choose $n=T^{2/3}$ and
\begin{equation}\label{equ:xi-n-T}
\xi=0.5 n^{-\frac{d_x}{3 d_x+4}}=0.5 T^{-\frac{2d_x}{9 d_x+12}}.
\end{equation}
Since $T$ satisfies
$T \geq \left(\frac{3}{2 p_Q} \right)^{\frac{9 d_x +12}{2 d_x}}$,
we have
\begin{equation}\label{equ:xi-pQ}
\xi \leq \frac{p_Q}{3}.
\end{equation}
Recall that the total regret in $T$ periods can be upper bounded by
\begin{equation}\label{equ:regret-lcoal}
R(T) \leq 2n\max_{\bm x \in \XX,y \in \YY}|f(\bm x, y)|+ \PR (\hat{J}=J)  R_2(T-n)+2 \max_{\bm x\in \XX, y\in \YY}|f(\bm x,  y)| \PR (\hat{J} \neq J) (T-n).
\end{equation}
where $n=T^{2/3}$. Since $T \geq (b_3)^{-(4.5 d_x+6)}$ and  $h=T^{-2/(9d_x+12)}$, we have $h \leq b_3$. Since $T \geq \left((1+6 e^{-1} +4\log b_0)/b_1\right)^{4.5+6/d_x}$, we have $n \geq \log(2 b_0)/(b_1 h^{d_x+4})$. Thus, applying Proposition~\ref{prop:small-xi}, for $\xi < 1/2$, we have
\begin{align*}
\PR(\hat{J} \neq J)
& \leq d_x\exp\left\{\left(2(1\!-\!\xi) \log b_0\!-\!\xi \log  \xi\!+\!2(1\!-\!\xi)\log(1\!-\!\xi)\right) h^{-d_x} \!-\!\min\left\{\xi,\frac{2p_Q}{3}\!-\!\xi\right\}b_1 h^4 n\right\}\\
&+ d_x\exp\left(-\frac{2}{9} p_Q^2 n\right)\\
& \overset{(a)}{\leq} d_x\exp\left\{\left(2 \log b_0-3\xi \log  \xi\right) h^{-d_x} -\min\left\{\xi,\frac{2p_Q}{3}-\xi\right\}b_1 h^4 n\right\}
+ d_x\exp\left(-\frac{2}{9} p_Q^2 n\right)\\
& \overset{(b)}{\leq} d_x\exp\left\{\left(2 \log b_0-3\xi \log  \xi\right) n^{d_x/(3d_x+4)} -\xi b_1 n^{3d_x/(3d_x+4)}\right\}
+ d_x\exp\left(-\frac{2}{9} p_Q^2 n\right)\\
& \overset{(c)}{\leq} d_x\exp\left\{\left(2 \log b_0-3\xi \log  \xi\right) T^{2d_x/(9d_x+12)} - b_1 T^{4d_x/(9d_x+12)}\right\}
+ d_x/\sqrt{T}\\
& \overset{(d)}{=} d_x\exp\left\{T^{2d_x/(9d_x+12)} \left(2 \log b_0-3\xi \log  \xi  - b_1 T^{2d_x/(9d_x+12)} \right)\right\}
+ d_x/\sqrt{T}\\
& \overset{(e)}{\leq} d_x\exp\left\{ \log T \left(2 \log b_0+3\xi \log (1/\xi)  - b_1 T^{2d_x/(9d_x+12)} \right)\right\}
+ d_x/\sqrt{T}\\
& \overset{(f)}{ \leq }d_x \exp\left\{-\frac{1}{2} \log T\right\}+d_x/\sqrt{T}\\
& = 2d_x/\sqrt{T},
\end{align*}
where $(a)$ follows from $0 \leq \xi \leq 1/2$ and $1-\xi \geq \xi$; $(b)$ follows from $h=n^{-1/(3 d_x+4)}$, $\xi \leq \frac{p_Q}{3}$ and \eqref{equ:xi-pQ}; $(c)$ follows from $n=T^{2/3}$, \eqref{equ:xi-n-T} and $T \geq \left(3/(2 p_Q)\right)^3 (\log T)^{3/2}$; $(e)$ follows from $T \geq \left(\log T\right)^{4.5+6/d_x}$; $(f)$ follows from $x \log (1/x) \leq e^{-1}$ for any $x \in (0,1]$ and $T \geq \left((1+6 e^{-1}+4 \log b_0)/b_1\right)^{4.5+6/d_x}$.

Further relaxing the right-hand side in \eqref{equ:regret-lcoal}, we have
\begin{align*}
R(T)
&\leq 2T^{2/3}\max_{\bm x \in \XX,y \in \YY}|f(\bm x, y)|+ R_2(T-n)+ 4 \max_{\bm x\in \XX, y\in \YY}|f(\bm x,  y)| T d_x/\sqrt{T} \\
&\leq O\left(T^{2/3}\right)+ R_2(T-n)+ O(\sqrt{T})\\
&=O(R_2(T)).
\end{align*}

\section{More Numerical Experiments}\label{app:numerical}


We conduct more numerical experiments where $d_x$ is reasonably large ($d_x=5,10$). Additionally, we focus on the results of variable selection phase since the algorithms used in the online learning phase are pretty standard. We omit $y$ in the reward function $f$ for the exposition and consider more complicated relationships between $\bm{x}$ and $f$.

The first function is designed to compare the performances of BV-LASSO and LASSO for an additive function. It's an additive combination of linear, discontinuous and polynomial functions, i.e.,
\begin{equation}\label{equ:numerical-f3}
f_3(x_1,x_2,x_3)=x_1+x_2 \cdot \mathbb{I} (x_2 > 0.2)+\sqrt{x_3}.
\end{equation}
The second function incorporates compound operators and non-trivial interactions between covariates, i.e.,
\begin{equation}\label{equ:numerical-f4}
f_4(x_1,x_2,x_3,x_4)=\exp\left\{2 x_1 -3 (x_2+x_3-1)^2 -\frac{1}{0.5+3 x_4}\right\}.
\end{equation}
Note that $f_4$ is increasing in $X_1$ and $X_4$, but non-monotone (even seemingly symmetric) in $X_2, X_3$. It's predictable that applying the standard LASSO is likely to miss $X_2$ and $X_3$.
The third function considers the periodic fluctuant covariates which is a more difficult task for the standard LASSO, i.e.,
\begin{equation}\label{equ:numerical-f5}
f_5(x_1,x_2,x_3,x_4)=\sin(4 x_1 x_2)+\sin(2 \pi x_1) \sin(\pi x_4)+\sin(2 \pi x_2) \cos(\pi x_3).
\end{equation}
{\RV Note that $f_3$, $f_4$, and $f_5$ all satisfy local relevance (Assumption~\ref{ass:gra-point}) but not global relevance (Assumption~\ref{ass:pos_gra}).}
The effective dimensions $d_x^*$ of $f_3,f_4,f_5$ are $3,3,4$. We test the algorithm in the settings where $d_x=5, 10$ and $n=10^2, 10^3, 10^5, 10^6$.

The side length $h$ is a critical hyper-parameter which balances the approximation error and statistical error. It should be small enough to guarantee the theoretical performance of BV-LASSO (see Proposition \ref{prop:global} and \ref{prop:small-xi}). However, considering the implementation, when $d_x$ is large but $n$ is limited, $h$ should not be too small. Otherwise, there will be few observations in each bin, which leads to a poor performance of the localized LASSO.
To make sure enough observations in each bin, we set $h=1/\lfloor{ n^{1/(d_x +4)}\rfloor}$, which also satisfies the theoretical guarantee in Proposition \ref{prop:rate_selection} and Theorem \ref{theo:cov_reg}.



When generating covariates, we consider a non-trivial distribution, where the relevant and redundant variables are not independent. Specifically, the relevant variables are independently sampled from a uniform distribution in $[0,1]^{d_x^*}$, while a redundant variable is generated by a linear combination of two relevant variables and an independently sampled external variable. For example, considering $d_x^*=3$, we first generate relevant $X_1, X_2, X_3 \sim U[0,1]^{3}$. Then we randomly select two of them, such as $X_1, X_3$, and let $X_4=(X_1+X_3)/8+3 X_4'/4$, where $X_4' \sim U[0,1]$ is independently sampled.

Table~\ref{table:BV-LASSO-dx5} and \ref{table:BV-LASSO-dx10} shows the $95\%$ confidence intervals of $\hat{J}$ according to \eqref{eq:weighted-vote} based on the average of 20 trials. The tables also show $\hat{J}$ for various $h$, ranging from $1, 1/2, 1/3$ to $1/\lfloor{ n^{1/(d_x +4)}\rfloor}$. Note that when $h=1$, BV-LASSO degrades to the standard LASSO.

In the following, we will carefully explain the numerical results in Table~\ref{table:BV-LASSO-dx5}, and the results in Table~\ref{table:BV-LASSO-dx10} follow the same except for higher $d_x$.

For the function $f_3$, we find that $\hat{J}^{(1)}, \hat{J}^{(2)}, \hat{J}^{(3)}$ are statistical significantly greater than $\hat{J}^{(4)}, \hat{J}^{(5)}$. So $X_1, X_2, X_3$ can be easier distinguished by choosing a suitable $\xi$. Additionally, when $h=1/2$ or $1/3$, we find $\hat{J}^{(1)}, \hat{J}^{(2)}, \hat{J}^{(3)}$ are less (not greater) than those when $h=1$. That's because the standard LASSO outperforms BV-LASSO when $f$ is (approximately) linear (see Remark \ref{rmk:rate-BV-LASSO}).

For the function $f_4$, $\hat{J}^{(1)}, \hat{J}^{(4)}$ are significantly greater than $\hat{J}^{(5)}$ for all $h$. But when $h=1$, the confidence intervals of $\hat{J}^{(2)}, \hat{J}^{(3)}$ have overlaps with $\hat{J}^{(5)}$. So if just applying the standard LASSO ($h=1$), it's impossible to distinguish $X_2, X_3$ from $X_5$. But when $h$ becomes smaller ($h=1/2$ or $1/3$), $\hat{J}^{(2)}, \hat{J}^{(3)}$ becomes larger and  $\hat{J}^{(5)}$ becomes smaller, also the lengths of confidence intervals diminish. Thus, when $h$ is small enough, $\hat{J}^{(2)}, \hat{J}^{(3)}$ are significantly greater than $\hat{J}^{(5)}$. So $X_2, X_3$ can be
successfully screened out by choosing a constant $\xi$ (such as $\xi=0.4$), or choosing $\xi$ diminishing as $n$.

The result of $f_5$ is similar to that of $f_4$. It's easy to screen out $X_1,X_2$ even if $h=1$. But the relevance of $X_3,X_4$ can only be detected when $h<1$.

In summary, the numerical results of $f_4, f_5$ show the advantage of BV-LASSO, namely, the relevant variables can be successfully selected even if the function is highly non-linear. The results also support the theoretical guarantee of BV-LASSO showed in Proposition \ref{prop:small-xi} and Corollary \ref{cor:local-relevance}.

\textbf{Computation complexity of BV-LASSO.} The overall computation complexity for implement BV-LASSO variable selection algorithm is $O\left(n^{1+8/(d_x+4)}\right)$. To see this, we set the bin size $h=1/ (n^{1/(d_x +4)})$, then we have $h^{-d_x}=n^{d_x/(d_x+4)}$ bins and averagely there are $n h^{-d_x}=n^{4/(d_x+4)}$ observations in each bin. Implementing the localized LASSO in all bins incurs the computing times $O\left(h^{-d_x} \left(n^{4/(d_x+4)}\right)^3 \right)=O\left(n^{1+8/(d_x+4)}\right)$. Then applying the
weighted voting incurs the computing times $O(h^{-d_x})=O\left(n^{d_x/(d_x+4)}\right)$.
So the bottleneck of implementing BV-LASSO is the localized LASSO. Fortunately, we can reduce the running time by using a multi-core computer, as the computing tasks in bins are independent. We perform BV-LASSO using a PC with 16 GB RAM and Inter Core i7-3770, 8 cores and 3.40 GHz. The last two rows in Table~\ref{table:BV-LASSO-dx5} and \ref{table:BV-LASSO-dx10} show the running times of parallel computing (denoted by P) and non-parallel computing (denoted by N) in seconds. The parallel computing requires more (less) running time than the non-parallel counterpart when $n$ is small (large). We also test the setting where $n=10^7$, and observe the running time of parallel (non-parallel) is 356 (5037) seconds. So when $n$ is large, the BV-LASSO algorithm can be implemented efficiently by using a parallel computer.

\begin{table}[htbp]
	\centering
	\caption{Numerical results for functions $f_3,f_4,f_5$ when $d_x=5$. For each function, the first $d_x^*$ variables are relevant, and the remaining are redundant. The values show the $95\%$ confidence intervals (CI) of $X_i$, e.g., $0.73 \pm 0.04$ represents the CI=$(0.69, 0.77)$. The last two rows show the running times of parallel (P) and non-parallel (N) computing in seconds.}
	\label{table:BV-LASSO-dx5}
	\begin{tabular}{cccccccc}
		\hline \\[-2.0ex]
		\multicolumn{2}{c}{\multirow{2}{*}{$d_x=5$}} & \multicolumn{1}{c}{$n=10^2$} &  \multicolumn{2}{c}{$n=10^3$} & \multicolumn{3}{c}{$n=10^5$}\\
		\cline{3-3} \cline{4-5} \cline{6-8}\\[-2.0ex]
		& & $h=1$ & $h=1$ & $h=1/2$ & $h=1$ & $h=1/2$ & $h=1/3$ \\
		\hline \\[-2.0ex]
		\multirow{5}{*}{$f_{3}$} & $X_{1}$ & $ 1.00 \pm 0.00$ & $ 1.00 \pm 0.00$ & $ 0.73 \pm 0.04$  & $ 1.00 \pm 0.00$ & $ 1.00 \pm 0.00$ & $ 1.00 \pm 0.00$ \\
		& $X_{2}$ & $ 1.00 \pm 0.00$  & $ 1.00 \pm 0.00$ & $ 0.77 \pm 0.03$  & $ 1.00 \pm 0.00$ & $ 1.00 \pm 0.00$ & $ 1.00 \pm 0.00$ \\
		& $X_{3}$ & $ 1.00 \pm 0.00$  & $ 1.00 \pm 0.00$ & $ 0.58 \pm 0.04$  & $ 1.00 \pm 0.00$ & $ 1.00 \pm 0.00$ & $ 0.96 \pm 0.01$ \\
		& $X_{4}$ & $ 0.05 \pm 0.10$  & $ 0.05 \pm 0.10$ & $ 0.05 \pm 0.02$  & $ 0.00 \pm 0.00$ & $ 0.01 \pm 0.01$ & $ 0.02 \pm 0.00$ \\
		& $X_{5}$ & $ 0.00 \pm 0.00$  & $ 0.10 \pm 0.14$ & $ 0.05 \pm 0.02$  & $ 0.05 \pm 0.10$ & $ 0.01 \pm 0.01$ & $ 0.02 \pm 0.01$ \\
		\hline \\[-2.0ex]
		\multirow{5}{*}{$f_{4}$} & $X_{1}$ & $ 1.00 \pm 0.00$  & $ 1.00 \pm 0.00$ & $ 0.71 \pm 0.03$  & $ 1.00 \pm 0.00$ & $ 1.00 \pm 0.00$ & $ 0.88 \pm 0.01$ \\
		& $X_{2}$ & $ 0.65 \pm 0.23$  & $ 0.45 \pm 0.24$ & $ 0.55 \pm 0.04$  & $ 0.30 \pm 0.22$ & $ 0.69 \pm 0.04$ & $ 0.68 \pm 0.01$ \\
		& $X_{3}$ & $ 0.45 \pm 0.24$  & $ 0.45 \pm 0.24$ & $ 0.58 \pm 0.05$  & $ 0.40 \pm 0.24$ & $ 0.65 \pm 0.02$ & $ 0.68 \pm 0.01$ \\
		& $X_{4}$ & $ 1.00 \pm 0.00$  & $ 1.00 \pm 0.00$ & $ 0.52 \pm 0.04$  & $ 1.00 \pm 0.00$ & $ 0.98 \pm 0.01$ & $ 0.69 \pm 0.01$ \\
		& $X_{5}$ & $ 0.45 \pm 0.24$  & $ 0.40 \pm 0.24$ & $ 0.25 \pm 0.04$  & $ 0.20 \pm 0.19$ & $ 0.13 \pm 0.03$ & $ 0.16 \pm 0.01$ \\
		\hline \\[-2.0ex]
		\multirow{5}{*}{$f_{5}$} & $X_{1}$ & $ 0.90 \pm 0.14$  & $ 1.00 \pm 0.00$ & $ 0.59 \pm 0.04$  & $ 1.00 \pm 0.00$ & $ 1.00 \pm 0.00$ & $ 0.96 \pm 0.00$ \\
		& $X_{2}$ & $ 0.65 \pm 0.23$  & $ 0.95 \pm 0.10$ & $ 0.60 \pm 0.05$  & $ 1.00 \pm 0.00$ & $ 1.00 \pm 0.00$ & $ 0.82 \pm 0.01$ \\
		& $X_{3}$ & $ 0.50 \pm 0.24$  & $ 0.50 \pm 0.24$ & $ 0.57 \pm 0.04$  & $ 0.40 \pm 0.24$ & $ 1.00 \pm 0.00$ & $ 0.68 \pm 0.01$ \\
		& $X_{4}$ & $ 0.60 \pm 0.24$  & $ 0.50 \pm 0.24$ & $ 0.52 \pm 0.05$ & $ 0.25 \pm 0.21$ & $ 1.00 \pm 0.00$ & $ 0.58 \pm 0.01$ \\
		& $X_{5}$ & $ 0.45 \pm 0.24$  & $ 0.35 \pm 0.23$ & $ 0.27 \pm 0.03$  & $ 0.30 \pm 0.22$ & $ 0.15 \pm 0.04$ & $ 0.16 \pm 0.01$ \\
		\hline \\[-2.0ex]
		\multirow{2}{*}{Time}& P & $2.34$  & $2.34$ & $2.32$  & $2.00$ & $2.04$ & $2.51$ \\
		& U & $0.00$  & $0.00$ & $0.02$  & $0.06$ & $0.08$ & $0.27$ \\
		\hline \\[-2.0ex]
	\end{tabular}
\end{table}

\begin{table}[htbp]
	\centering
	\caption{Numerical results for functions $f_3,f_4,f_5$ when $d_x=10$.}
	\label{table:BV-LASSO-dx10}
	\begin{tabular}{cccccccccc}
		\hline \\[-2.0ex]
		\multicolumn{2}{c}{\multirow{2}{*}{$d_x=10$}} & & \multicolumn{1}{c}{$n=10^3$} & &  \multicolumn{2}{c}{$n=10^5$} & & \multicolumn{2}{c}{$n=10^6$}\\
		\cline{4-4} \cline{6-7} \cline{9-10}\\[-2.0ex]
		& & & $h=1$ & & $h=1$ & $h=1/2$ & & $h=1$ & $h=1/2$ \\
		\hline \\[-2.0ex]
		\multirow{10}{*}{$f_{3}$} & $X_{1}$ & & $ 1.00 \pm 0.00$ &  & $ 1.00 \pm 0.00$ & $ 0.94 \pm 0.00$ &  & $ 1.00 \pm 0.00$ & $ 1.00 \pm 0.00$ \\
		& $X_{2}$ & & $ 1.00 \pm 0.00$ &  & $ 1.00 \pm 0.00$ & $ 0.96 \pm 0.00$ &  & $ 1.00 \pm 0.00$ & $ 1.00 \pm 0.00$ \\
		& $X_{3}$ & & $ 1.00 \pm 0.00$ &  & $ 1.00 \pm 0.00$ & $ 0.72 \pm 0.01$ &  & $ 1.00 \pm 0.00$ & $ 1.00 \pm 0.00$ \\
		& $X_{4}$ & & $ 0.00 \pm 0.00$ &  & $ 0.10 \pm 0.14$ & $ 0.01 \pm 0.00$ &  & $ 0.00 \pm 0.00$ & $ 0.00 \pm 0.00$ \\
		& $X_{5}$ & & $ 0.00 \pm 0.00$ &  & $ 0.00 \pm 0.00$ & $ 0.01 \pm 0.00$ &  & $ 0.00 \pm 0.00$ & $ 0.00 \pm 0.00$ \\
		& $X_{6}$ & & $ 0.00 \pm 0.00$ &  & $ 0.00 \pm 0.00$ & $ 0.01 \pm 0.00$ &  & $ 0.00 \pm 0.00$ & $ 0.00 \pm 0.00$ \\
		& $X_{7}$ & & $ 0.10 \pm 0.14$ &  & $ 0.00 \pm 0.00$ & $ 0.01 \pm 0.00$ &  & $ 0.00 \pm 0.00$ & $ 0.00 \pm 0.00$ \\
		& $X_{8}$ & & $ 0.05 \pm 0.10$ &  & $ 0.00 \pm 0.00$ & $ 0.01 \pm 0.00$ &  & $ 0.00 \pm 0.00$ & $ 0.00 \pm 0.00$ \\
		& $X_{9}$ & & $ 0.00 \pm 0.00$ &  & $ 0.00 \pm 0.00$ & $ 0.01 \pm 0.00$ &  & $ 0.00 \pm 0.00$ & $ 0.00 \pm 0.00$ \\
		& $X_{10}$ & & $ 0.05 \pm 0.10$ &  & $ 0.00 \pm 0.00$ & $ 0.01 \pm 0.00$ &  & $ 0.00 \pm 0.00$ & $ 0.00 \pm 0.00$ \\
		\hline \\[-2.0ex]
		\multirow{10}{*}{$f_{4}$} & $X_{1}$ & & $ 1.00 \pm 0.00$ &  & $ 1.00 \pm 0.00$ & $ 0.80 \pm 0.01$ &  & $ 1.00 \pm 0.00$ & $ 0.99 \pm 0.00$ \\
		& $X_{2}$ & & $ 0.40 \pm 0.24$ &  & $ 0.50 \pm 0.24$ & $ 0.49 \pm 0.01$ &  & $ 0.25 \pm 0.21$ & $ 0.61 \pm 0.01$ \\
		& $X_{3}$ & & $ 0.60 \pm 0.24$ &  & $ 0.25 \pm 0.21$ & $ 0.49 \pm 0.01$ &  & $ 0.20 \pm 0.19$ & $ 0.60 \pm 0.01$ \\
		& $X_{4}$ & & $ 1.00 \pm 0.00$ &  & $ 1.00 \pm 0.00$ & $ 0.54 \pm 0.01$ &  & $ 1.00 \pm 0.00$ & $ 0.86 \pm 0.00$ \\
		& $X_{5}$ & & $ 0.30 \pm 0.22$ &  & $ 0.15 \pm 0.17$ & $ 0.08 \pm 0.00$ &  & $ 0.10 \pm 0.14$ & $ 0.06 \pm 0.00$ \\
		& $X_{6}$ & & $ 0.35 \pm 0.23$ &  & $ 0.15 \pm 0.17$ & $ 0.08 \pm 0.00$ &  & $ 0.10 \pm 0.14$ & $ 0.06 \pm 0.00$ \\
		& $X_{7}$ & & $ 0.30 \pm 0.22$ &  & $ 0.20 \pm 0.19$ & $ 0.08 \pm 0.00$ &  & $ 0.10 \pm 0.14$ & $ 0.06 \pm 0.00$ \\
		& $X_{8}$ & & $ 0.45 \pm 0.24$ &  & $ 0.15 \pm 0.17$ & $ 0.08 \pm 0.00$ &  & $ 0.05 \pm 0.10$ & $ 0.06 \pm 0.00$ \\
		& $X_{9}$ & & $ 0.45 \pm 0.24$ &  & $ 0.15 \pm 0.17$ & $ 0.08 \pm 0.00$ &  & $ 0.15 \pm 0.17$ & $ 0.07 \pm 0.00$ \\
		& $X_{10}$ & & $ 0.30 \pm 0.22$ &  & $ 0.15 \pm 0.17$ & $ 0.08 \pm 0.00$ &  & $ 0.25 \pm 0.21$ & $ 0.06 \pm 0.00$ \\
		\hline \\[-2.0ex]
		\multirow{10}{*}{$f_{5}$} & $X_{1}$ & & $ 1.00 \pm 0.00$ &  & $ 1.00 \pm 0.00$ & $ 0.62 \pm 0.00$ &  & $ 1.00 \pm 0.00$ & $ 0.93 \pm 0.00$ \\
		& $X_{2}$ & & $ 1.00 \pm 0.00$ &  & $ 1.00 \pm 0.00$ & $ 0.62 \pm 0.01$ &  & $ 1.00 \pm 0.00$ & $ 0.93 \pm 0.00$ \\
		& $X_{3}$ & & $ 0.30 \pm 0.22$ &  & $ 0.30 \pm 0.22$ & $ 0.67 \pm 0.01$ &  & $ 0.20 \pm 0.19$ & $ 1.00 \pm 0.00$ \\
		& $X_{4}$ & & $ 0.35 \pm 0.23$ &  & $ 0.50 \pm 0.24$ & $ 0.68 \pm 0.01$ &  & $ 0.25 \pm 0.21$ & $ 1.00 \pm 0.00$ \\
		& $X_{5}$ & & $ 0.35 \pm 0.23$ &  & $ 0.20 \pm 0.19$ & $ 0.09 \pm 0.00$ &  & $ 0.15 \pm 0.17$ & $ 0.07 \pm 0.00$ \\
		& $X_{6}$ & & $ 0.10 \pm 0.14$ &  & $ 0.10 \pm 0.14$ & $ 0.09 \pm 0.00$ &  & $ 0.05 \pm 0.10$ & $ 0.07 \pm 0.00$ \\
		& $X_{7}$ & & $ 0.10 \pm 0.14$ &  & $ 0.15 \pm 0.17$ & $ 0.09 \pm 0.01$ &  & $ 0.25 \pm 0.21$ & $ 0.07 \pm 0.00$ \\
		& $X_{8}$ & & $ 0.25 \pm 0.21$ &  & $ 0.10 \pm 0.14$ & $ 0.09 \pm 0.00$ &  & $ 0.15 \pm 0.17$ & $ 0.07 \pm 0.00$ \\
		& $X_{9}$ & & $ 0.30 \pm 0.22$ &  & $ 0.15 \pm 0.17$ & $ 0.09 \pm 0.00$ &  & $ 0.20 \pm 0.19$ & $ 0.07 \pm 0.00$ \\
		& $X_{10}$ & & $ 0.25 \pm 0.21$ &  & $ 0.25 \pm 0.21$ & $ 0.09 \pm 0.01$ &  & $ 0.20 \pm 0.19$ & $ 0.07 \pm 0.00$ \\
		\hline \\[-2.0ex]
		\multirow{2}{*}{Time}& P & & $2.32$ &  & $2.04$ & $2.64$ &  & $3.76$ & $3.78$ \\
		& U & & $0.00$ &  & $0.10$ & $1.15$ &  & $1.18$ & $9.68$ \\
		\hline \\[-2.0ex]
	\end{tabular}
\end{table}

\clearpage
\section{\RV Proofs for Section~\ref{sec:discussion}}
\subsection{\RV Proof of Proposition \ref{prop:decision-rule}}
\begin{proof}
    If $f_P$ satifies Assumptions~\ref{ass:continuous},	\ref{ass:spa_cov}, \ref{ass:smooth_cov} and Assumption \ref{ass:pos_gra}, then the proof of Proposition \ref{prop:subset} also goes through for $f_P$. So does Proposition \ref{prop:global}, and Theorem \ref{theo:cov_reg}. We will check these condition one by one.

	\textbf{Continuously differentiable.} We first prove the interchangeability of expectation and derivative. Considering the variable $x_i$,
	\begin{align}
	\frac{\partial}{\partial x_i} \EX_P[f(\bm{x}, Y)]
	&=\lim_{h \rightarrow 0} \dfrac{\EX_P[f(\bm{x}+h e_i, Y)]-\EX_P[f(\bm{x}, Y)]}{h} \notag \\
	&\overset{(a)}{=}\lim_{h \rightarrow 0} \EX_P \left[\dfrac{f(\bm{x}+h e_i, Y)-f(\bm{x}, Y)}{h} \right] \notag\\
	&\overset{(b)}{=}\EX_P \left[\lim_{h \rightarrow 0} \dfrac{f(\bm{x}+h e_i, Y)-f(\bm{x}, Y)}{h} \right] \notag\\
	&= \EX_P\left[\frac{\partial}{\partial x_i} f(\bm{x}, Y)\right], \label{equ:exp-der-change}
	\end{align}
	where $e_i$ denotes the unit vector with the $i$-th entry 1, $(a)$ follows from $P$ is independent of $\bm x$, $(b)$ follows from the dominated convergence theorem.
	So we have
	\begin{equation}\label{equ:pathwise}
	\nabla f_P(\bm{x})=\frac{\partial}{\partial \bm{x}} \EX_P[f(\bm{x}, Y)]=\EX_P\left[\frac{\partial}{\partial \bm{x}} f(\bm{x}, Y)\right].
	\end{equation}
	Next, we use \eqref{equ:pathwise} to show $f_P$ is continuously differentiable. For a sequence of $\bm{x}_n$ converges to $\bm{x}$, we have
	\begin{equation*}
	\nabla f_P(\bm{x}_n)=\EX_P\left[\frac{\partial}{\partial \bm{x}} f(\bm{x}_n, Y)\right] \overset{(c)}{\longrightarrow}  \EX_P\left[\frac{\partial}{\partial \bm{x}} f(\bm{x}, Y)\right]=\nabla f_P(\bm{x}),
	\end{equation*}
	where $(c)$ is supported by the bounded convergence theorem since the partial derivative $\frac{\partial}{\partial \bm{x}} f(\bm{x}_n, y)$ is bounded and pointwise converges to $\frac{\partial}{\partial \bm{x}} f(\bm{x}, y)$ for any $y$.

	\textbf{Sparse Reward Function.} We will show that $f_P$ depends on the same set of relevant variables as in $f$. By \eqref{equ:random-decision}, we have
	\begin{equation*}
	f_P(x_1,\ldots,x_{d_x})=\EX_P[f(x_1,\ldots,x_{d_x},Y)]=\EX_P[g(x_{i_1},\ldots,x_{i_{d_x^*}},Y)]=g_P(x_{i_1},\ldots,x_{i_{d_x^*}}).
	\end{equation*}

        \textbf{Global Relevance.} By Assumption \ref{ass:pos_gra}, we know that
        \begin{equation*}
            \frac{\partial f(\bm{x},y)}{\partial x_i} \geq C \ \text{or} \ \le -C, \quad \forall i \in J, \bm{x} \in \XX, y \in \YY.
        \end{equation*}
        So we have
        \begin{equation*}
            \left| \frac{\partial f_P(\bm{x})}{\partial x_i} \right| = \left|\EX_P \left[\frac{\partial f(\bm{x},Y)}{\partial x_i}\right]\right| \geq C.
        \end{equation*}

	\textbf{Second-order Smoothness.} For any $\bm x_1, \bm x_2 \in \XX$, we have
	\begin{align*}
	&\left|f_P(\bm x_1)-f_P(\bm x_2)- \nabla f_P(\bm x_2)^T (\bm x_1 - \bm x_2)\right|\\
	&\overset{(a)}{=}\left|\EX_P[f(\bm{x}_1, Y)]-\EX_P[f(\bm{x}_2, Y)]-\EX_P\left[\frac{\partial}{\partial \bm x} f(\bm{x}_2, Y)\right]^T (\bm x_1- \bm x_2)\right|\\
	&\overset{(b)}{=}\left|\EX_P\left[f(\bm{x}_1, y)-f(\bm{x}_1, y)-\frac{\partial}{\partial \bm x} f(\bm{x}_2, Y)^T (\bm x_1- \bm x_2) \right]\right|\\
	& \leq \EX_P\left[ \left|f(\bm{x}_1, y)-f(\bm{x}_1, y)-\frac{\partial}{\partial \bm x} f(\bm{x}_2, Y)^T (\bm x_1- \bm x_2) \right| \right]\\
	& \overset{(c)}{\le} L \|\bm x_1- \bm x_2\|^2_{\infty},
	\end{align*}
	where $(a)$ follows from \eqref{equ:exp-der-change}, $(b)$ follows from the independence of $\bm x$ and $P$, $(c)$ follows from Assumsption \ref{ass:smooth_cov}.
\end{proof}

\subsection{\RV Proof of Theorem \ref{theo:stagewise-LASSO}}
\begin{proof}
	We introduce some notations before the proof. Let $R(m)$ denote the regret incurred in stage $m$ and $R_2(m)$ denote the regret incurred in the exploitation phase of stage $m$.
        We also define $M_f=\max_{\bm{x} \in \mathcal{X}, y \in \mathcal{Y}} f(\bm{x},y)$, $m_0=\lfloor 1/b_3 \rfloor$.

	For the stage $m \leq m_0$, the conditions of Propostion \ref{prop:global} are not satisfied because of $h_m = 1/m > b_3$. So there's no probability guarantee for the BV-LASSO to correctly select the variables. Then the exploitation phase will incur a linear growing regret $R_2(m)=O(n_m)$. While for $m > m_0$, the BV-LASSO correctly selects the variables with a high probablity and the regret $R_2(m)=O(n_m) \PR(\hat{J}^m \neq J)+\PR(\hat{J}^m=J)O(n_m^{1-1/(d_x^*+3)} \log(n_m))$. Thus we decompose the total regret in $T$ periods into two parts
	\begin{align}
	R(T)
	&= \sum_{m=1}^{m_0} l_m M_f+R_2(m) + \sum_{m=m_0+1}^{M} l_m M_f+R_2(m) \notag \\
	&=\sum_{m=1}^M l_m M_f + \sum_{m=1}^{m_0} n_m M_f + \sum_{m=m_0+1}^M O\left(n_m^{(d_x^*+2)/(d_x^*+3)} \log n_m\right) \PR(\hat{J}^m=J) \\
	& \ \ + n_m M_f\PR(\hat{J}^m \neq J),\label{equ:stagewise-regret}
	\end{align}
	where $M$ is number of total stages when the algorithm proceeds to period $T$.

	Next we will give upper bound for each term in \eqref{equ:stagewise-regret}. For the ease of notation, we denote $S_m \coloneqq \sum_{i=1}^m l_i$.
	For the first term in \eqref{equ:stagewise-regret}, by the definition of $l_m$, we have
	\begin{align} \label{equ:stagewise-lm-up}
	S_m=\sum_{i=1}^m l_i
	= (d_x+4) b_4 \sum_{i=1}^m i^{d_x+3}
	\leq (d_x+4) b_4 \int_{1}^{m+1} x^{d_x+3} dx
	\leq b_4 (m+1)^{d_x+4}.
	\end{align}
	Also, we have for any $m$,
	\begin{align} \label{equ:stagewise-lm-low}
	S_m=\sum_{i=1}^m l_i \geq (d_x+4) b_4 (1+\int_{1}^{m} x^{d_x+3} dx) \geq b_4 m^{d_x+4}.
	\end{align}
	For the second term in \eqref{equ:stagewise-regret}, we have
	\begin{align} \label{stagewise-n_m-m_0}
	\sum_{m=1}^{m_0} n_m
	= b_4 \sum_{m=1}^{m_0} 2^m \leq b_4 2^{m_0+1} \leq b_4 2^{1+1/b_3},
	\end{align}
	where the last inequality follows from the definition of $m_0$.
	The third term in \eqref{equ:stagewise-regret} is easy to handle as $\PR(\hat{J}^m = J) \leq 1$. For the last term in \eqref{equ:stagewise-regret}, we use Propsition \ref{prop:global} to bound $\PR(\hat{J}^m \neq J)$. The conditions of Proposition \ref{prop:global} are satisfied by \eqref{equ:stagewise-lm-low}, $b_4 \geq \log(2 b_0)/b_1$ and $m \geq m_0$. Thus,
	we have
	\begin{align}
	\PR(\hat{J}^m \neq J)
	& \overset{(a)}{\leq} d_x \exp \left\{\frac{1}{2} \left(h_{m}^{-d_x} (1+\log 2+ \log b_0)-b_1 S_m h_m^4\right)\right\} \notag\\
	& \overset{(b)}{=} d_x \exp \left\{\frac{1}{2} \left(m^{d_x} (1+\log 2+ \log b_0)-b_1 S_m m^{-4}\right)\right\} \notag\\
	& \overset{(c)}{\leq} d_x \exp \left\{\frac{1}{2} \left(m^{d_x} (1+\log 2+ \log b_0)-(1+2 \log 2+ \log b_0 + 2 \log d_x) m^{d_x}\right)\right\} \notag\\
	& \overset{(d)}{=} d_x \exp \left\{-\frac{1}{2} m^{d_x}\left( \log 2 + 2 \log d_x\right) \right\} \notag\\
	& \overset{(e)}{\le} \exp \left\{-\frac{1}{2} m \log 2\right\} \notag\\
	&\overset{(f)}{=} (n_m/b_4)^{-1/2}
	\overset{(g)}{\leq}  (n_m/b_4)^{-1/(d_x^*+3)}, \label{equ:stagewise-false}
	\end{align}
	where $(a)$ follows from Proposition \ref{prop:global}, $(b)$ follows from $h_m=1/m$, $(c)$ follows from \eqref{equ:stagewise-lm-low} and definition of $b_4$, $(e)$ follows from $d_x \geq 1, m^{d_x} \ge m$, $(f)$ follows from the definition $n_m=b_4 2^m$ and $(g)$ follows from $n_m/b_4=2^m \geq 1$ and $d_x^* \geq 0$.
	Thus, the last term in \eqref{equ:stagewise-regret} has the order $O(n_m^{1-1/(d_x^*+3)})$.

	Next, we give an upper bound for $M$. Recall that $M$ is the minimal integer such that
	\begin{equation}\label{equ:stagewise-M}
	S_M + \sum_{m=1}^M n_m\ge T.
	\end{equation}
	We claim that $M \leq \log(T/b_4)$. To see this, substituting $M$ by $\log(T/b_4)$ into the LHS of \eqref{equ:stagewise-M}, we have
	\begin{equation}
	S_M + \sum_{m=1}^M n_m \geq \sum_{m=1}^M n_m
	\geq  b_4 2^{M+1}
	\geq 2 T \geq T.
	\end{equation}

	At the end, combining \eqref{equ:stagewise-regret}, \eqref{equ:stagewise-lm-up}, \eqref{stagewise-n_m-m_0}, \eqref{equ:stagewise-false}, we have
	\begin{align*}
	R(T)
	&\leq M_f b_4 (M+1)^{d_x+4} + 2M_f b_4 2^{1/b_3} + \sum_{m=1}^M  O\left(n_m^{(d_x^*+2)/(d_x^*+3)} \log n_m\right)\\
	&\ \ + \sum_{m=1}^M M_f n_m (n_m/b_4)^{-1/(d_x^*+3)}\\
	&\overset{(a)}{\leq} M_f b_4 (M+1)^{d_x+4} + 2M_f b_4 2^{1/b_3} + \sum_{m=1}^M  O\left(n_m^{(d_x^*+2)/(d_x^*+3)} \log n_m\right)\\
	&\overset{(b)}{\leq} M_f b_4 (M+1)^{d_x+4} + 2M_f b_4 2^{1/b_3} +  O\left(2^{(M+1)(d_x^*+2)/(d_x^*+3)} \log T\right)\\
	&\overset{(c)}{\leq} M_f b_4 (\log(T/b_4)+1)^{d_x+4} + 2M_f b_4 2^{1/b_3} + O\left(T^{(d_x^*+2)/(d_x^*+3)} \log T\right)\\
	&\overset{(d)}{=} O\left(T^{(d_x^*+2)/(d_x^*+3)} \log T\right),
	\end{align*}
	where $(a)$ follows from $n_m^{1-1/(d_x^*+3)}=n_m^{(d_x^*+2)/(d_x^*+3)}$, $(b)$ follows from $n_m =b_4 2^m\leq T$, and $\sum_{m=1}^M 2^{m(d_x^*+2)/(d_x^*+3)}= \frac{2^{(M+1)(d_x^*+2)/(d_x^*+3)}-1}{2^{(d_x^*+2)/(d_x^*+3)}-1} \leq 2^{(M+1)(d_x^*+2)/(d_x^*+3)}$, $(c)$ follows from $M \leq \log(T/b_4)$, $(d)$ follows from $T \geq \left(b_4 2^{1/b_3}\right)^{3/2}$ and $(\log T)^{d_x+4} \leq T^{2/3}$.
\end{proof}

\subsection{\RV Proof of Proposition \ref{prop:subset-OLS}}

\begin{proof}
	For simplicity, we use $\bm{\hat{\theta}}$ instead of $\bm{\theta}^{OLS}$ in \eqref{equ:standard_OLS}. Recalling the definitions for the design matrix $A$ and the sample covariance matrix $\hat{\psi}$ in \eqref{equ:cov-mat-part}, $\bm{\theta}^*$ and $\bm{\rho}$ in \eqref{eq:lr-form}, we have
	\begin{equation}\label{equ:OLS-formula}
	\bm{\hat{\theta}}=(\hat{\Psi})^{-1} A^T \bm{Z}=\bm{\theta}^*+(\hat{\Psi})^{-1} A^T \bm{\rho} \coloneqq \bm{\theta}^*+F \bm{\rho}.
	\end{equation}
	Then
	\begin{equation*}
	F F^T =(\hat{\Psi})^{-1} A^T A (\hat{\Psi})^{-1} =(\hat{\Psi})^{-1}.
	\end{equation*}
	By \eqref{equ:covar-low-lambda}, we have
	\begin{equation}\label{equ:eigenvalue-OLS}
	\PR(\lambda_{\min}(\hat{\Psi}) \leq (1-\alpha) \underline{\lambda}) \leq (d_x+1) \left(\frac{e^{-\alpha}}{(1-\alpha)^{(1-\alpha)}}\right)^{n \underline{\lambda}/(1+d_x/4)}.
	\end{equation}
	Conditional on the event $\lambda_{\min}(\hat{\Psi}) \leq (1-\alpha) \underline{\lambda}$, we have
	\begin{equation*}
	\sum_{k=1}^n f_{ik}^2 = (F F^T)_{jj} =\bm{e}_j^T F F^T \bm{e}_j \leq \lambda_{\max} (F F^T) =\lambda_{\max} ((\hat{\Psi})^{-1}) =\lambda_{\min}^{-1}(\hat{\Psi} ) \leq \frac{1}{(1-\alpha) \underline{\lambda}}.
	\end{equation*}
	Next, by equation \eqref{equ:OLS-formula}, we have
	\begin{equation}\label{equ:OLS-error}
	|\hat{\bm{\theta}}_i-\bm{\theta}_i^*|=\left(\big|F(\Delta+\frac{1}{\sqrt{n}} \epsilon)\big|\right)_i.
	\end{equation}
	We give an upper bound for the first term in \eqref{equ:OLS-error},
	\begin{equation}\label{equ:OLS-error-bound}
	(|F \Delta|)_i \leq \left(\sum_{k=1}^n f_{ik}^2\right)^{1/2} \|\Delta\|_2 \leq \sqrt{\lambda_{\min}^{-1}(\hat{\Psi} )} \|\Delta\|_2 \leq \sqrt{\frac{64}{(1-\alpha) \underline{\lambda}}} L d_x h^2,
	\end{equation}
	where the last inequality follows from equation \eqref{equ:lasso_Delta_norm}.
	The second term in \eqref{equ:OLS-error},
	\begin{equation*}
	\left(\frac{1}{\sqrt{n}} F \bm{\epsilon}\right)_i=\sqrt{\frac{1}{n}} \sum_{k=1}^n f_{ik} \epsilon_k
	\end{equation*}
	is a mean-zero $\sqrt{\frac{1}{n}\sum_{k=1}^n f_{ik}^2} \sigma$ sub-Gaussian random variable. Then, setting
	\begin{equation}\label{equ:OLS-r}
	r=\frac{8 L d_x +\sqrt{2}}{\sqrt{(1-\alpha) \underline{\lambda}}} h^2,
	\end{equation}
	and conditional on $\lambda_{\min}(\hat{\Psi}) \leq (1-\alpha) \underline{\lambda}$, we have
	\begin{align}
	\PR(|\hat{\bm{\theta}}_i-\bm{\theta}_i^*| \geq r)
	&=\PR\left(\left(|F(\Delta+\frac{1}{\sqrt{n}} \epsilon)|\right)_i \geq r\right) \notag \\
	&\leq \PR\left(\left(\frac{1}{\sqrt{n}} F \bm{\epsilon}\right)_i \geq r-(|F \Delta|)_i\right) \notag\\
	& \leq \PR\left(\left(\frac{1}{\sqrt{n}} F \bm{\epsilon}\right)_i \geq \sqrt{\frac{2}{(1-\alpha) \underline{\lambda}}} h^2\right) \notag\\
	& \leq \exp(-n h^4/\sigma^2). \label{equ:concentration-OLS}
	\end{align}
	For $i \in J$, $|\bm{\theta}_i^*| \geq Ch$; for $i \notin J$, $\bm{\theta}_i^* =0 $.
	Then, if $r \leq 0.5 Ch $, we can seperate $J, J^c$. Namely, when $r$ satisfies \eqref{equ:OLS-r} and
	\begin{equation*}
	h \leq \dfrac{\sqrt{(1-\alpha) \underline{\lambda}} C}{2(8 L d_x +\sqrt{2})},
	\end{equation*}
	and setting $\alpha=0.5, \underline{\lambda}=\frac{\mu_m}{12}$, we have
	\begin{equation}\label{equ:OLS-bound}
	\PR \left(\hat{J}_j=J\right) \geq 1- 2(d_x+1) \exp\left(-n h^4 \min\left\{\frac{1}{\sigma^2},\frac{\mu_m}{6(4+d_x)} \right\} \right),
	\end{equation}
	where \eqref{equ:OLS-bound} follows from \eqref{equ:eigenvalue-OLS} and \eqref{equ:concentration-OLS}.
\end{proof}

\subsection{\RV Proof of Corollary \ref{cor:sparse-known}}

Define a good event $G \coloneqq \{\big|\hat{\bm{\theta_i}}-\bm{\theta^*_i}\big| < 0.5 Ch, \forall i\}$. We first show that under the event $G$, $\hat{J}_j=J$. Recalling that for $i \in J$, $|\bm{\theta}_i^*| \geq Ch$; for $i \notin J$, $\bm{\theta}_i^* =0 $. So on the event $G$, we have
\begin{equation*}
\max_{j \in J^c} |\hat{\bm{\theta}}_j|< 0.5 Ch, \  \min_{j \in J} |\hat{\bm{\theta}}_j|> 0.5 Ch,
\end{equation*}
which implies
\begin{equation*}
\min_{j \in J} |\hat{\bm{\theta}}_j| > \max_{j \in J^c} |\hat{\bm{\theta}}_j|.
\end{equation*}
So the variables picked up the $d_x^*$ largest estimates is the true relevant set, namely, $\hat{J}_j=J$.

Next, we give an upper bound for the probability of $G^c$.
\begin{align}
\PR(\hat{J}_j \neq J)
&\le \PR(G^c)= \PR\left(\exists i, \big|\hat{\bm{\theta}}_i-\bm{\theta}_i^*\big| \geq 0.5 Ch\right)\notag\\
& \leq d_x \PR\left( \big|\hat{\bm{\theta}}_i-\bm{\theta}_i^*\big| \geq 0.5 Ch\right) \notag\\
&\overset{(a)}{=}d_x \PR\left(\left(\big|F(\Delta+\frac{1}{\sqrt{n}} \epsilon)\big|\right)_i \geq 0.5 Ch\right) \notag\\
&\overset{(b)}{\le} d_x \PR\left(\left(\frac{1}{\sqrt{n}} F \bm{\epsilon}\right)_i \geq 0.5 Ch-\sqrt{\frac{64}{(1-\alpha) \underline{\lambda}}} L d_x h^2 \right), \label{equ:concentration-h}
\end{align}
where $(a)$ follows from \eqref{equ:OLS-error}, $(b)$ follows from \eqref{equ:OLS-error-bound}.
Setting $\alpha=0.5$, $ \underline{\lambda}=\frac{\mu_m}{12}$, if $h$ is small enough such that
\begin{equation*}
h < \dfrac{C \sqrt{\mu_m}}{64 \sqrt{6} L d_x},
\end{equation*}
then we have
\begin{equation*}
\sqrt{\frac{64}{(1-\alpha) \underline{\lambda}}} L d_x h^2 < \frac{1}{4} Ch,
\end{equation*}
and
\begin{align*}
\eqref{equ:concentration-h}
\leq d_x\PR\left(\left(\frac{1}{\sqrt{n}} F \bm{\epsilon}\right)_i \geq \frac{1}{4} Ch\right) \leq d_x\exp\left(-\frac{\mu_m C^2}{768 \sigma^2} n h^2\right).
\end{align*}
Thus, combining with \eqref{equ:eigenvalue-OLS}, we have
\begin{equation}
\PR \left(\hat{J}_j=J\right) \geq 1- 2(d_x+1) \exp\left(-n h^2 \min\left\{\frac{\mu_m C^2}{768 \sigma^2},\frac{\mu_m}{6(4+d_x)} \right\} \right),
\end{equation}

\subsection{\RV Proof of Theorem~\ref{theo:linear_sparse}}\label{proof:linear_sparse}
We first show a probability bound for the variable selection, which is parallel to Proposition \ref{prop:subset}.
\begin{proposition}[Variable Selection of LASSO in linear model]\label{prop:linear-LASSO}
	Under Assumption \ref{ass:sub_gau}, \ref{ass:spa_par}, \ref{ass:cov_des_lin}, choosing $\lambda=\frac{C'(1+3\gamma) \underline{\lambda}}{4(1+\gamma)\sqrt{d_x}}$ in Algorithm \ref{alg:Linear_LASSO}, we have the following high probablity bound
	\begin{equation}\label{equ:linear-LASSO-bound}
	\PR \left(\hat{J}_k=\supp\{\bm{\theta}_k\}\right) \geq 1- b_0' \exp(-b_1' n), \ \text{for} \ k \in [K]
	\end{equation}
	where $C' \leq \min\{(\bm{\theta}_k)_j: (\bm{\theta}_k)_j \neq 0 \ \text{for}\ k \in [K], j \in [d_x]\}$ and the constants $b_0'$, $b_1'$ show in \eqref{equ:linear-lasso-constants}.
\end{proposition}

\begin{proof}
	The proof mainly follows the same argument as in Proposition \ref{prop:subset}. The only difference is that $h=1$ and the approximation error $\bm{\Delta}=0$ since the ``best'' linear approximation in \eqref{equ:L2} is the true parameter $\bm{\theta}_k$ in \eqref{equ:linear-bandit-model}. So there's no need to consider the approximation error in choosing $\lambda$. After choosing
	\begin{equation*}
	\lambda=\frac{C'(1+3\gamma) \underline{\lambda}}{4(1+\gamma)\sqrt{d_x}},
	\end{equation*}
	we revise the constants $c_5, c_6$ in \eqref{equ:lasso_con_prob2} and \eqref{equ:lasso_max_H}:
	\begin{equation*}
	c_5=\dfrac{C'^2 \underline{\lambda}^2 (1-\gamma)^2 (1+3 \gamma)^2 }{16 (1+\gamma)(3+\gamma) \overline{\lambda} \sigma^2 d_x}, \  c_6=\dfrac{C'^2 (1+3 \gamma) \underline{\lambda}}{16 (1+\gamma) \sigma^2},
	\end{equation*}
	and the constants $b_0$, $b_1$ in \eqref{equ:lasso-local-constants}
	\begin{align}
	&b_0':=2\max\{2 (d_x+1), d_x^2/4\}, \notag\\
	&b_1':=c_1 \wedge c_3  \wedge c_5 \wedge c_6 \notag\\
	& \quad \ =\left\{c_1 \wedge (1-\gamma)^2 \underline{\lambda}^2/(8 (d_x^*)^2) \wedge \dfrac{C'^2 \underline{\lambda}^2 (1-\gamma)^2 (1+3 \gamma)^2 }{16 (1+\gamma)(3+\gamma) \overline{\lambda} \sigma^2 d_x} \wedge \dfrac{C'^2 (1+3 \gamma) \underline{\lambda}}{16 (1+\gamma) \sigma^2} \right\}, \notag\\
	&c_1:=\frac{\underline{\lambda}}{2 (1\!+\!\gamma) (1\!+\!d_x/4)} \min\left\{1\!-\!\gamma\!+\!(3\gamma \!+\!1) \log\left( \frac{3\gamma\!+\!1}{2\!+\!2\gamma}\right), \gamma\!-\!1\!+\!(3\!+\!\gamma) \log\left(\frac{3\!+\!\gamma}{2\!+\!2\gamma}\right) \right\}. \label{equ:linear-lasso-constants}
	\end{align}
\end{proof}

Next, we state the margin condition and arm optimality condition (see \citealt{bastani2020online} for more discussion).

\begin{assumption}[Margin Condition]\label{ass:margin-condition}
	There exists a constant $C_0 >0$ such that for all $i$ and $j$ in $[K]$ where $i \neq j$, $\PR\left(0 < |\bm X^T (\bm \theta_i-\bm \theta_j)| \le \kappa\right) \le C_0 \kappa$ for all $\kappa >0$.
\end{assumption}

\begin{assumption}[Arm Optimality Condition]\label{ass:arm-opt}
	Let $\mathcal{K}_{opt}$ and $\mathcal{K}_{sub}$ be mutually exclusive sets that include all $K$ arms. Sub-optimal arms $i \in \mathcal{K}_{sub}$ satisfy $\bm X^T \bm \theta_i < \max_{j \neq i} \bm X^T \bm \theta_j-h$ for some $h >0$ and every $\bm X \in \mathcal{X}$. On the other hand, each optimal arm $i \in \mathcal{K}_{opt}$, has a corresponding set
	\begin{equation*}
	U_i \coloneqq \left\{\bm X \big | \bm X^T \bm \theta_i > \max_{j \neq i} \bm X^T \bm \theta_j +h \right\}.
	\end{equation*}
	We assume there exists $p_{*}>0$ such that $\min_{i \in \mathcal{K}_{opt}} \PR(U_i) \ge p_{*}$. Define $\Sigma_{i} \coloneqq \EX[\bm X \bm X^T | \bm X \in U_i]$ for all $i \in [K]$. Then, there exists $\phi_0 >0$ such that for all $i \in [K]$ the minimum eigenvalue $\lambda_{\min} (\Sigma_i) \geq \phi_0 >0$.
\end{assumption}

Finally, we prove the upper bound for the combined regret. Recall that the total regret in $T$ periods can be upper bounded by
\begin{align*}
R(T) &\leq 2Kn\max_{\bm x,k }|\bm{x}^t \bm{\theta}_k|+ \PR (\forall \ k, \ \hat{J}_k=\supp\{\bm{\theta}_k\})  R_2(T-Kn)\\
&+2 \max_{\bm x ,k}|\bm{x}^t \bm{\theta}_k| \PR (\exists k, \ \hat{J_k} \neq \supp\{\bm{\theta}_k\}) (T-Kn).
\end{align*}
where $n=\log T/b_1'$. Then by Proposition~\ref{prop:linear-LASSO}, we have
\begin{equation*}
\PR(\forall \ k, \ \hat{J}_k=\supp\{\bm{\theta}_k\}) \geq  1- Kb_0' \exp(-b_1' n),
\end{equation*}
where $b_0'=O(d_x^2), \  b_1'=O(1/d_x)$.
Further relaxing the right-hand side, we have
\begin{align*}
R(T)
&\leq 2K \log T \max_{\bm x ,k}|\bm{x}^t \bm{\theta}_k|/b_1'+ R_2(T-Kn)+ 2 \max_{\bm x, k}|\bm{x}^t \bm{\theta}_k| Kb_0' \exp(-b_1' n) T\\
&=O(K d_x \log T)+O(R_2(T))+ O(K d_x^2)\\
&=O\left(K\left(d_x+(d_x^*)^3\right) \log T+ K d_x^2\right),
\end{align*}
where the last equation follows from $R_2(T)=O((d_x^*)^3 \log T)$.

\end{document}